\documentclass[twoside]{article}

\usepackage[accepted]{aistats2022}
\usepackage[utf8]{inputenc}
\usepackage[T1]{fontenc}
\usepackage{url}
\usepackage{booktabs}
\usepackage{amsfonts}
\usepackage{nicefrac}
\usepackage{microtype}
\usepackage[hidelinks]{hyperref}
\hypersetup{
    colorlinks = true,
    linkcolor = blue,
    urlcolor  = blue,
    citecolor = blue,
    anchorcolor = blue}
\usepackage{listings}
\usepackage[inline]{enumitem}
\usepackage[table, dvipsnames]{xcolor}
\usepackage{graphicx}
\usepackage{amsmath}
\usepackage{float}
\usepackage[export]{adjustbox}
\usepackage{subfig}
\usepackage{caption} 
\usepackage{mathtools, nccmath}
\usepackage{tikz}
\usepackage{multicol}
\usepackage{wrapfig}
\usepackage{algorithm}
\usepackage[algo2e, ruled,vlined, boxed, linesnumbered]{algorithm2e}
\SetArgSty{textnormal}
\usepackage[colorinlistoftodos]{todonotes}
\usepackage{setspace}

\usepackage{amsmath,amsfonts,bm}

\def\1{\bm{1}}

\DeclareMathAlphabet{\mathsfit}{\encodingdefault}{\sfdefault}{m}{sl}
\SetMathAlphabet{\mathsfit}{bold}{\encodingdefault}{\sfdefault}{bx}{n}

\newcommand{\defeq}{\vcentcolon=}
\newcommand{\norm}[1]{\left\lVert#1\right\rVert}

\def\bP{\mathbf{P}}

\def\cW{\overline{W}_{\varepsilon}}
\def\We{W_{\varepsilon}}

\DeclarePairedDelimiterX{\dotp}[2]{\langle}{\rangle}{#1, #2}

\definecolor{darkblue}{HTML}{1A254B}
\definecolor{lightblue}{HTML}{A7BED3}
\definecolor{blue}{HTML}{2B50AA}
\definecolor{green}{HTML}{81B5AE}
\definecolor{pink}{HTML}{F2545B}
\definecolor{red}{HTML}{A4243B}
\definecolor{lightgray}{HTML}{E5EBEF}

\lstdefinestyle{codestyle}{
    commentstyle=\color{blue},
    keywordstyle=\color{lightblue},
    numberstyle=\tiny\color{gray},
    stringstyle=\color{pink},
    basicstyle=\ttfamily\footnotesize,
    breakatwhitespace=false,         
    breaklines=true,                 
    captionpos=b,                    
    keepspaces=true,                 
    showspaces=false,                
    showstringspaces=false,
    showtabs=true,                  
    tabsize=2,
    frame=leftline
}
\usepackage[round]{natbib}
\renewcommand{\bibname}{References}
\renewcommand{\bibsection}{\subsubsection*{\bibname}}

\newcommand*{\tabindent}{\hspace{3mm}}

\begin{document}
\setlength{\abovedisplayskip}{6pt}
\setlength{\belowdisplayskip}{6pt}
% If your paper is accepted and the title of your paper is very long,
% the style will print as headings an error message. Use the following
% command to supply a shorter title of your paper so that it can be
% used as headings.
%
%\runningtitle{I use this title instead because the last one was very long}

% If your paper is accepted and the number of authors is large, the
% style will print as headings an error message. Use the following
% command to supply a shorter version of the authors names so that
% they can be used as headings (for example, use only the surnames)
%
%\runningauthor{Surname 1, Surname 2, Surname 3, ...., Surname n}

\twocolumn[

\aistatstitle{% \textsc{JKOnet}: 
    Proximal Optimal Transport Modeling of Population Dynamics}

\aistatsauthor{Charlotte Bunne \And Laetitia Meng-Papaxanthos \And  Andreas Krause \And Marco Cuturi}

% \aistatsaddress{ETH Zurich \\ \texttt{bunnec@ethz.ch} \And Google Research \\ \texttt{lpapaxanthos@google.com} \And ETH Zurich \\ \texttt{krausea@ethz.ch} \And Google Research \\ \texttt{cuturi@google.com} } ]
\aistatsaddress{ETH Zurich \And Google Research \And ETH Zurich \And Google Research$^\ddag$} ]

\begin{abstract}
We propose a new approach to model the collective dynamics of a population of particles evolving with time. As is often the case in challenging scientific applications, notably single-cell genomics, measuring features for these particles requires destroying them. As a result, the population can only be monitored with periodic snapshots, obtained by sampling a few particles that are sacrificed in exchange for measurements. Given only access to these snapshots, can we reconstruct likely individual trajectories for all other particles? We propose to model these trajectories as collective realizations of a causal Jordan-Kinderlehrer-Otto (JKO) flow of measures: The JKO scheme posits that the new configuration taken by a population at time $t+1$ is one that trades off an improvement, in the sense that it decreases an \textit{energy}, while remaining close (in Wasserstein distance) to the previous configuration observed at $t$. %In a JKO model, the trajectory is entirely characterized by the energy function and the initial configuration. 
In order to learn such an energy using only snapshots, we propose \textsc{JKOnet}, a neural architecture that computes (in end-to-end differentiable fashion) the JKO flow given a parametric energy and initial configuration of points. % \textsc{JKOnet} outputs (iteratively) a new configuration with optimal displacements, recovered as gradients of input convex neural networks (ICNN). 
We demonstrate the good performance and robustness of the \textsc{JKOnet} fitting procedure, compared to a more direct forward method.
\end{abstract}

\everypar{\looseness=-1}

\section{Introduction}\label{sec:intro}
\paragraph{Population Dynamics ...} Many fields in science carry out experiments by monitoring complex systems composed of evolving particles. That monitoring consists in sampling, every now and then, a few representative particles in the system, and measure their features. As a result, the observer has access to a collection of time-resolved point-clouds describing partially the dynamic of that population on aggregate. Such problems arise in many fields, when for instance, observing a population of cells in biology \citep{schiebinger2019, moon2019}, densities in meteorology~\citep{fisher2009data,sigrist2015stochastic} or multi-target tracking~\citep{luo2020multiple, sheldon2007collective, sheldon2011collective, haasler2019estimating, haasler2021multimarginal, haasler2021multi}.

\paragraph{... Without Individual Paths.}
While modeling and estimating parametric dynamics using datasets of point trajectories is the core subject of time series analysis (see~\citealt{li2020scalable, krishnan2017structured} and references therein), the setting we consider makes it difficult to track the evolution of individual particles. Indeed, this would require tagging and measuring repeatedly the same particles, which can be costly or even impossible: For instance, measuring a cell's transcriptome requires splitting the cell o. With this constraint in mind, our goal is to better understand the evolution of single particles, using only the aggregate data described in point clouds.

\begin{figure}[t]
    \centering
    \includegraphics[width=.9\linewidth]{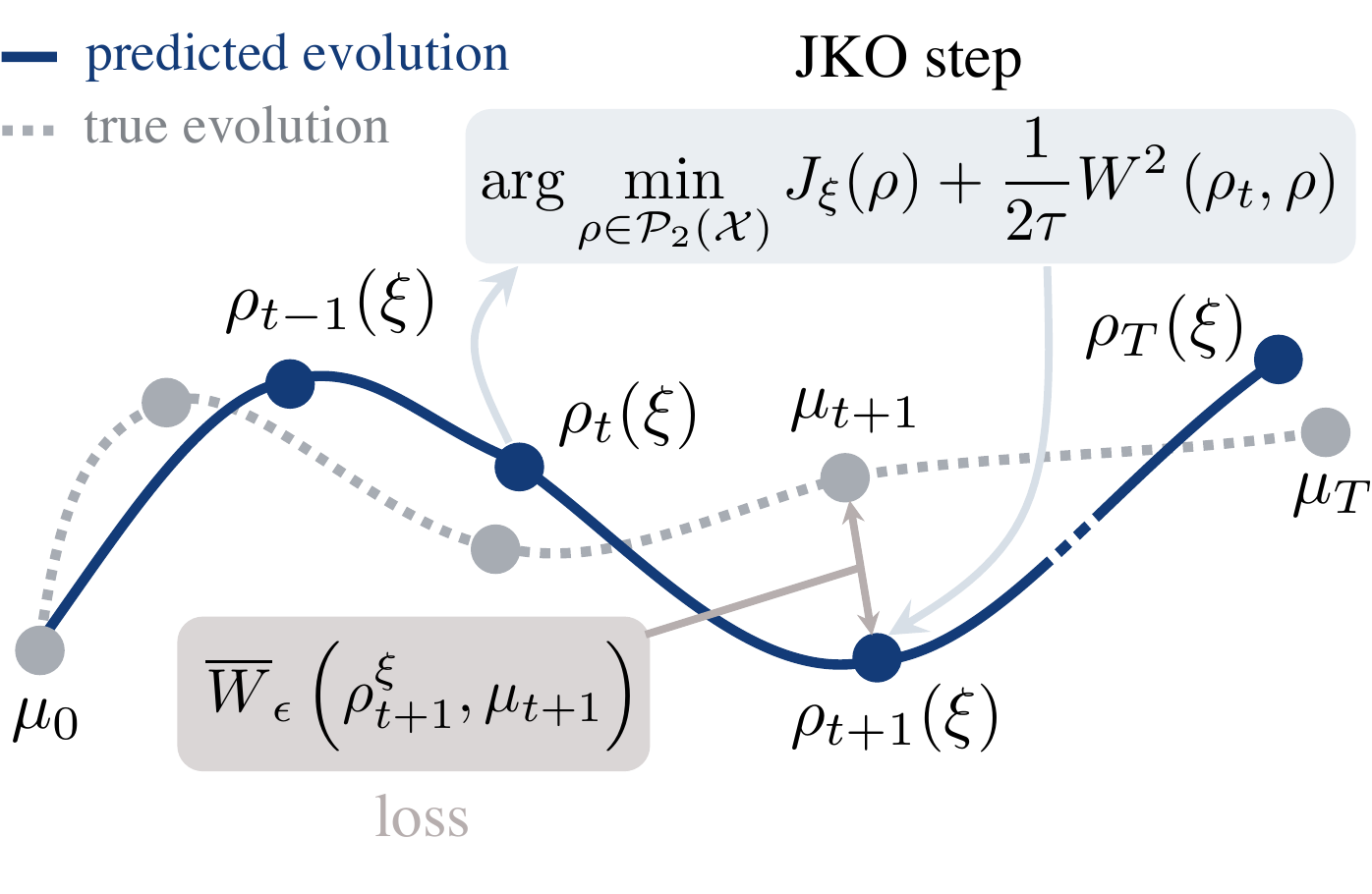}
    \caption{Given an observed trajectory $(\mu_0,\dots,\mu_T)$ of point clouds (gray), we seek parameters $\xi$ for the energy $J_\xi$ such that the predictions $\rho_1, \dots, \rho_T$ (blue) following a JKO flow from $\rho_0=\mu_0$ are close the observed trajectory (gray), by minimizing (as a function of $\xi$) the sum of Wasserstein distances between $\rho_{t+1}$, the JKO step from $\rho_{t-1}$ using $J_\xi$, and data $\mu_{t+1}$.}
    \label{fig:overview}
\end{figure}

\newpage
\paragraph{Inferring Particle Paths from Cloud Trajectories.}
When the observer only seeks to reconstruct particles' paths given starting and ending point cloud configurations, the machinery of optimal transport (OT)~\citep{schiebinger2019,yang2020predicting,yang2018scalable} or likelihood-based normalizing flows (NF)~\citep{rezende2015variational,grathwohl2018ffjord} can be used, either separately, or even combined: \citet{tong2020trajectorynet} use OT to motivate a regularizer (squared norm of displacements) in their NF estimation pipeline; ~\citet{huang2021convex} restrict their attention to flows expressed as gradients of convex functions. This choice is motivated by OT because it agrees with the \citet{Brenier1987} principle that displacements arising from convex potentials give rise to optimal flows.
When the observer seeks instead a \textit{causal model}, namely one that is able to explain/predict future configurations of the point cloud (and not only interpolate between configurations), the parameters of that model can also be fitted with OT, as proposed by \citet{hashimoto2016learning}. Their model assumes a Langevin dynamic for the particles, driven by the gradient flow of a (neural) energy function; They fit the parameters of that network by minimizing regularized OT distances~\citep{cuturi2013sinkhorn} between their model's predictions and the corresponding ground truth snapshots. %Conceptually, the approach of \citet{hashimoto2016learning} can do more than just interpolation, to instead provide a model that can be used to extrapolate / predict future configurations from initial configurations.
%Both methods restrict themselves to interpolations and require access to the start and end point of the underlying dynamics.

\textbf{Modeling Particle Dynamics as a JKO Scheme.} In this paper, we draw inspiration from both approaches above---the intuition from the recent NF literature that flows should mimic an optimal transport (OT as prior), and be able, through training, to predict future configurations (OT as a loss)---to propose a causal model for population dynamics. Our approach relies on a powerful hammer: the Jordan-Kinderlehrer-Otto (JKO) flow~\citep{jordan1998variational}, widely regarded as one of the most influential mathematical breakthroughs in recent history. While the JKO flow was initially introduced as an alternative method to solve the Fokker-Planck partial differential equation (PDE), its flexibility can be showcased to handle more complex PDEs \cite[\S4.7]{santambrogio2017euclidean}, or even describe the gradient flows of non-differentiable energies that have no PDE representation.
On a purely mechanical level, a JKO step is to measures what the proximal step~\citep{combettes2011proximal} is to vectors: In a JKO step, particles move to decrease collectively an {\em energy} (a real-valued function defined on measures), yet remain close (in Wasserstein sense) to the previous configuration. Our goal in this paper is to treat JKO steps as parameterized modules, and fit their parameter (the energy function) so that its outputs agree repeatedly over time with observed data. 
This approach presents several challenges: While numerical approaches to solve JKO steps have been proposed in low dimensional settings~\citep{burger2010, carrillo2021primal, 2015-Peyre-siims,benamou2016augmented}, scaling it to higher dimensions is an open problem. Moreover, minimizing a loss involving a JKO step w.r.t. energy requires not only solving the JKO problem, but also computing the (transpose) Jacobian of its output w.r.t. energy parameters.

% We wish to reconstruct that energy potential from observations, assuming each observations follows iteratively a JKO flow.

\textbf{Contributions.}\hspace{1em} Our contributions are two-fold. First, we propose a method, given an input configuration and an energy function, to compute JKO steps using input convex neural networks (ICNN)~\citep{amos2017input,pmlr-v119-makkuva20a} (see also concurrent works that have proposed similar approaches~\citep{alvarez2021optimizing, mokrov2021large}). Second, we view the JKO step as an inner layer, a \textsc{JKOnet} module parameterized by an energy function, which is tasked with moving the particles of an input configuration along an OT flow (the gradient of an optimal ICNN), trading off a lower energy with proximity to the previous configuration.
%This module, \textsc{JKOnet}, can be therefore seen as a parameterized \emph{mover} of particles that is constrained by design to produce locally (OT) optimal displacements. 
We propose to estimate the parameters of the energy by minimizing a fitting loss %(as a function of the energy itself) 
computed between the outputs of the \textsc{JKOnet} module (the prediction) and the ground truth displacements, as illustrated in Figure~\ref{fig:overview}.
We demonstrate \textsc{JKOnet}'s range of applications by applying in on synthetic potential- and trajectory-based population dynamics, as well as developmental trajectories of human embryonic stem cells based on single-cell genomics data.

\section{Background}
% In this section, we introduce cornerstones of optimal transport, such as the regularized optimal transport formulation and Brenier's theorem, together with the JKO flow scheme and convex neural architectures. 

\paragraph{Optimal Transport.} 
For two probability measures $\mu, \nu$ in $\mathcal{P}(\mathbb{R}^d)$, their squared 2-Wasserstein distance is
\begin{equation} \label{eq:ot}
    W_2^2(\mu, \nu) = \inf_{\gamma\in \Gamma(\mu,\nu)}\iint \|x-y\|^2_2 \gamma(dx, dy),
\end{equation}
where $\Gamma(\mu, \nu)$ is the set of couplings on $\mathbb{R}^d\times\mathbb{R}^d$ with respective marginals $\mu, \nu$. When instantiated on finite discrete measures, such as $\mu=\sum_{i=1}^n a_i\delta_{x_i}$ and $\nu=\sum_{j=1}^m b_j\delta_{y_j}$, this problem translates to a linear program, which can be regularized using an entropy term~\citep{cuturi2013sinkhorn,Peyre2019computational}. For $\varepsilon\geq0$, set 
\begin{equation}\label{eq:reg-ot}
\We(\mu,\nu) \defeq \min_{\bP\in U(a,b)} \dotp{\bP}{[\|x_i - y_j\|^2]_{ij}}  \,-\varepsilon H(\bP),
\end{equation}
where $H(\bP) \defeq -\sum_{ij} \bP_{ij} (\log \bP_{ij} - 1)$ and the polytope $U(a,b)$ is the set of $n\times m$ matrices $\{\bP\in\mathbb{R}^{n \times m}_+, \bP\mathbf{1}_m =a, \bP^\top\mathbf{1}_n=b\}$. 
Notice that the definition above reduces to the usual (squared) 2-Wasserstein distance when $\varepsilon=0$. Setting $\varepsilon>0$ yields a faster and differentiable proxy to approximate $W_{0}$, but introduces a bias, since $\We(\mu,\mu)\ne 0$ in general. % To correct that bias, we use 
In the rest of this work, we therefore use the \textit{Sinkhorn divergence}~\citep{ramdas2017wasserstein,genevay2018,salimans2018improving,feydy2018interpolating} as %to recover 
a valid non-negative discrepancy,
\begin{equation} \label{eq:sinkhorn}
\cW(\mu,\nu) \defeq \We(\mu,\nu)-\frac{1}{2}\left(\We(\mu,\mu)+\We(\nu,\nu)\right).\\
\end{equation}

\paragraph{OT and Convexity.} 
An alternative formulation for OT is given by the \citet{Monge1781} problem  
\begin{align}\label{eq:monge}
W_2^2(\mu,\nu) &= \inf_{T:T_{\#} \mu = \nu} \int_\mathcal{X} ||x - T(x)||^2d\mu(x) \,
\end{align}
where $\#$ is the push-forward operator, and the optimal solution $T^\star$ is known as the \citeauthor{Monge1781} map between $\mu$ and $\nu$. The \citeauthor{Brenier1987} theorem \citeyearpar{Brenier1987} states that if $\mu$ has a density, the Monge map $T^\star$ between $\mu$ and $\nu$ can be recovered as the gradient of a unique (up to constants) convex function $\psi$ whose gradient pushes forward $\mu$ to $\nu$. Namely, if $\psi:\mathbb{R}^d \rightarrow \mathbb{R}$ is convex and $(\nabla \psi)_{\#}\mu = \nu$, then $T^\star(x)=\nabla \psi(x)$ and
\begin{align}\label{eq:brenier}
W_2^2(\mu,\nu) &= \int_\mathcal{X} ||x - \nabla \psi(x)||^2 d\mu(x)\,.
\end{align}
%establishing an equivalence between the Kantorovich formulation of OT~\eqref{eq:ot} with optimal transport plan $\gamma$, and the Monge formulation involving map $T$. %, which consists of finding a map that associates points coming from both probability measures. 

\paragraph{JKO Flows.}
In their seminal paper, \citet{jordan1998variational} study diffusion processes under the lens of the OT metric \citep[see also][]{ambrosio2006gradient} and introduce a scheme that is now known as the JKO flow: Starting with $\rho_0$, and given a real-valued energy function $J:\mathcal{P}(\mathbb{R}^d)\rightarrow \mathbb{R}$ driving the evolution of the system, they define iteratively for $t\geq 0$, :
\begin{equation} \label{eq:jko}
    \rho_{t+1} = \arg \min_{\rho\in \mathcal{P}_2(\mathbb{R}^d)} J(\rho) + \frac{1}{2\tau} W^2(\rho, \rho_{t})\,,
\end{equation}
where $\tau$ is a time step parameter. These successive minimization problems result in a sequence of probability measures in $\mathcal{P}(\mathbb{R}^d)$. The JKO flow can thus be seen as the analogy of the usual proximal descent scheme, tailored for probability measures~\citep[p.285]{santambrogio2015optimal}. \citet{jordan1998variational} show that as step size $\tau \rightarrow 0$, and for a specific energy $J$ that is the sum of a linear term and the negentropy, the measures describing the JKO flow recover solutions to a Fokker-Planck equation. In this work, following in the footsteps of more general applications of the JKO scheme~\citep[\S4.8]{santambrogio2017euclidean}, we model dynamics without necessarily having in mind PDE solutions in mind, to interpret instead the JKO step as a more general parametric type of dynamic for probability measures, exclusively parameterized by the energy $J$ itself.

\paragraph{Convex Neural Architectures.}
Input convex neural networks are neural networks $\psi_\theta(x)$ with specific constraints on the architecture and parameters $\theta$, such that their output is a convex function of some (or all) elements of the input $x$~\citep{amos2017input}. We consider in this work \textit{fully} input convex neural networks (ICNNs), such that the output is a convex function of the entire input $x$. A typical ICNN is a $L$-layer, fully connected network such that, for $l = 0, \dots, L-1$:
\begin{equation} \label{eq:icnn}
    z_{l+1} = a_l(W^x_lx + W^z_l z_l + b_l)  \text{ and } \psi_\theta(x) = z_L,
\end{equation}
where by convention, $z_0$ and $W^z_0$ are $0$, $a_l$ are convex non-decreasing (non-linear) activation functions, $\theta=\{b_l, W^z_l, W^x_l\}_{l=0}^{L-1}$ are the weights and biases of the neural network, with weight matrices $W^z_l$ associated to latent representations $z$ that have non-negative entries. Since \citet{amos2017input}'s work, convex neural architectures have been further extended and shown to capture relevant models despite these constraints~\citep{amos2017input, pmlr-v119-makkuva20a, huang2021convex}. In particular, \citet{chen2018optimal} provide a theoretical analysis that any convex function over a convex domain can be approximated in sup norm by an ICNN.

\section{Proximal Optimal Transport Model} 

Given $T$ discrete measures $\mu_0, \dots, \mu_T$ describing the time evolution of a population, we posit that such an evolution follows a JKO flow for the free energy functional $J$, and assume that energy does not change throughout the dynamic. We parameterize the energy $J$ as a neural network with parameters $\xi$, and fit $\xi$ so that the JKO flow model matches the observed data. 

Fitting parameter $\xi$ with a reconstruction loss requires, using the chain rule, being able to differentiate the JKO step's output w.r.t. $\xi$ (see Fig.~\ref{fig:overview}), and more precisely provide a way to apply that transpose Jacobian to an arbitrary vector when using reverse-mode differentiation. To achieve this, we introduce a novel approach to numerically solve JKO flows using ICNNs (\S~\ref{sec:jko_icnn}), resulting in a bilevel optimization problem targeting the energy $J_\xi$ (\S~\ref{sec:learn_energy}).

\subsection{Reformulation of JKO Flows via ICNNs} \label{sec:jko_icnn}
Given a starting condition $\rho_t$ and energy functional $J_\xi$, the JKO step consists in producing a new measure $\rho_{t+1}$ implicitly defined as the minimizer of~\eqref{eq:jko}. Solving directly~\eqref{eq:jko} on the space of measures, involves substantial computational costs. Different numerical schemes have been developed, e.g., based notably on Eulerian discretization of measures \citep{carrillo2021primal, benamou2016}, and/or entropy-regularized optimal transport \citep{2015-Peyre-siims}. However, these methods are limited to small dimensions since the cost of discretizing such spaces grows exponentially. Except for the Eulerian approach proposed in \citep{2015-Peyre-siims}, obtained as the fixed point of a Sinkhorn type iteration, the differentiation would also prove extremely challenging as a function of the energy parameter $\xi$.

\looseness=-1 To reach scalability and differentiability, we build upon the approach outlined in \citet{benamou2016} to reformulate the JKO scheme as a problem solved over convex functions, rather than on measures $\rho$. Effectively, this is equivalent to making a change of variables in~\eqref{eq:jko}: Introduce a (variable) convex function $\psi$, and replace the variable $\rho$ by the variable $\nabla \psi_{\#}\rho_t$. Writing
\begin{equation}\label{eq:en}
\begin{split}
\mathcal{E}_J(\rho, \nu) := J(\rho) +\frac{1}{2 \tau}W_2^2(\rho, \nu),\\
\end{split}
\end{equation}
this identity states that, assuming $\mu$ and $\nu$ being absolutely continuous w.r.t. Lebesgue measure that
$$\min_{\rho}\mathcal{E}_J(\rho,\nu) = \min_{\psi \text{ convex}} \mathcal{F}_J(\psi, \nu):= \mathcal{E}_J(\nabla \psi_{\#}\nu, \nu)\,,$$
simplifying the Wasserstein term in \eqref{eq:en}, using the assumption that $\psi$ is convex and \citeauthor{Brenier1987}'s theorem (\S~\ref{sec:intro}):
\begin{equation}\mathcal{F}_J(\psi, \nu) = J(\nabla \psi_{\#}\nu) +\frac{1}{2 \tau} \!\! \int\!\! \| x - \nabla \psi(x) \|^2 d \nu(x)\label{eq:JKO_psi}
\end{equation}

We pick an ICNN architecture to optimize over a restricted family of convex functions, $\{\psi_{\theta}\}$, and define, starting from $\rho_0(\xi):=\mu_0$, the recursive sequence for $t\geq 0$,
\begin{equation} \label{eq:next_pop}
\rho_{t+1}(\xi) := \nabla \psi_{\theta^\star\!(\xi, \rho_t(\xi))\, \#}\, \rho_{t}(\xi)\,,
\end{equation}
with $\theta^\star(\xi, \rho_t)$ defined implicitly using $\xi$ and any $\nu$ as 
\begin{align} \label{eq:thetastar}
    \theta^\star(\xi, \nu):=\arg \min_{\theta} \mathcal{F}_J(\psi_{\theta},\nu)
\end{align}

\vspace{-20pt} \paragraph{Strong Convexity of $\psi_\theta$.} The strong convexity and smoothness of a potential $\psi$ impacts the regularity of the corresponding OT map $\nabla\psi$ ~\citep{caffarelli2000monotonicity,figalli2010optimal}, since one can show that for a $\ell$-strongly convex, $L$-smooth $\psi$ one has~\citep{paty2020regularity} that
$$
\ell \|x - y\| \leq \|\nabla\psi(x) -\nabla\psi(y)\|  \leq L\|x - y\|.
$$
While it is more difficult to enforce the $L$-smoothness of a neural network, and more generally its Lipschitz constants \citep{scaman2018lipschitz} it is easy to enforce its strong convexity, by simply adding a term $\ell \|x\|^2/2$ to the corresponding potential, or a residual rescaled term $\ell x$ to the output $\nabla\psi(x)$. This approach can be used to enforce that the push-forward of the gradient of an ICNN does not collapse to a single point, maintaining spatial diversity.

\subsection{Learning the Free Energy Functional}  \label{sec:learn_energy}
The energy function $J_\xi : \mathcal{P}(\mathbb{R}^d) \rightarrow \mathbb{R}$ can be any parameterized function taking a measures as an input. 
Since our model assumes that the observed dynamic is parameterized entirely by that energy (and the initial observation $\rho_0$), the more complex this dynamic, the more complex one would expect the energy $J_\xi$ to be. We focus in this first attempt on linear functions in the space of measures, that is expectations over $\rho$ of a vector-input neural network $E_\xi$
\begin{equation} \label{eq:energy}
    J_\xi(\rho) := \int E_\xi(x) d\rho(x),
\end{equation}
where $E_\xi:\mathbb{R}^d \rightarrow \mathbb{R}$ is a multi-layer perceptron (MLP).

\begin{algorithm}[t]
\KwIn{Dataset $\mathcal{D}=\{\{\mu_t^0 \}_{t=0}^T, \ldots, \{\mu_t^N \}_{t=0}^T\}$ of $N$ population trajectories, $\xi^0$ energy parameter initialization, $\theta^0$ ICNN parameter initialization, learning rates $\text{lr}_\theta$ and $\text{lr}_\xi$, step $\tau$, regularizer $\varepsilon$, tolerance $\alpha$, {\texttt{TeacherForcing}} flag.}
\KwOut{Free energy $J_{\xi}$ explaining underlying population dynamics of snapshot data.}
$\xi\leftarrow \xi^0$

\For{$\{\mu_t\}_{t=0}^T \in \mathcal{D}$} {
\For{$t \gets 0$ \textbf{to} $T-1$} {

  $\theta\leftarrow \theta^0$

  \If{\texttt{TeacherForcing}}{
    $\nu \leftarrow \mu_t$
  }
  \Else{$\nu \leftarrow \rho_t(\xi)$}
  \While{$\frac{\sum_i \norm{\nabla_{\theta_i}\mathcal{F}_{J_\xi}(\theta)}_2}{\sum_i \text{count}(\theta_i)} \ge \alpha$}{
    
    $\theta \leftarrow \theta - \text{lr}_\theta \times \nabla_\theta \mathcal{F}_{J_\xi,\nu}(\theta)$
  }
  $\rho_{t+1}(\xi) \leftarrow \nabla \psi_{\theta \#} \nu$

  $\xi \leftarrow \xi - \text{lr}_\xi \times \nabla_\xi \cW(\rho_{t+1}(\xi), \mu_{t+1})$
}
}
\Return{$J_{\xi}$}

\caption{\textsc{JKOnet} Algorithm.}
\label{algo:jkonet}
\end{algorithm}

\begin{figure}[H]
 \vspace{-10pt}
    \centering
    \includegraphics[width=.9\linewidth]{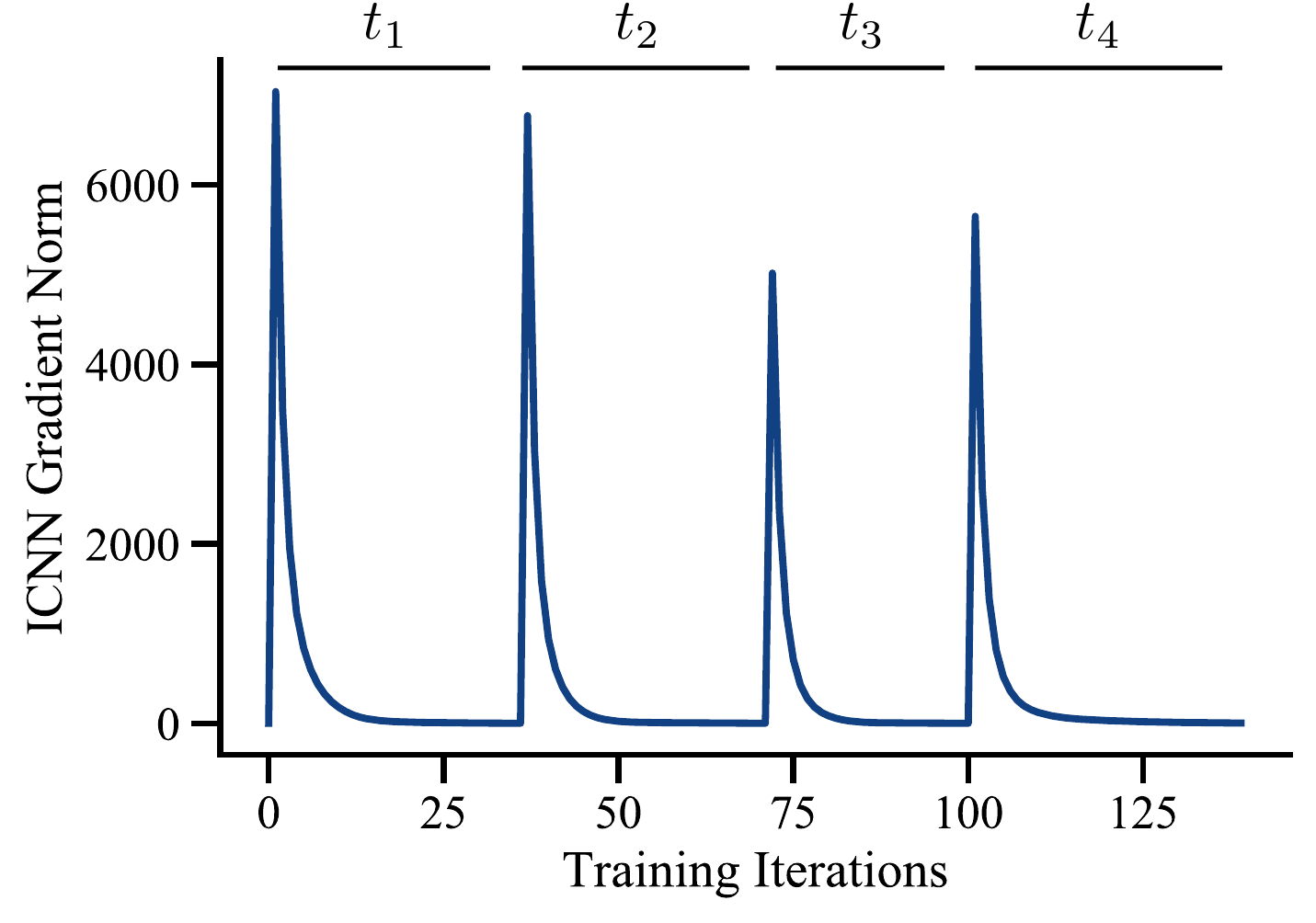}
    \caption{Optimization of the ICNN used in JKO steps. The bumps correspond to a change in the outer iteration, the smooth decrease in between correspond to a single minimization~\eqref{eq:thetastar} of a time step $t_i$. \vspace{-10pt}}
    \label{fig:training_icnn}
\end{figure}

Inferring nonlinear energies accounting for population growth and decline, as well as interactions between points, using the formalism of \citep{de2019stochastic}, transformers~\citep{vaswani2017attention} or set pooling methods \citep{edwards2016, zaheer2017}, is an exciting direction for future work.

To address slow convergence and instabilities for dynamics with many snapshots, we use teacher forcing \citep{williams1989learning} to learn $J_\xi$ through time. In those settings, during training, $J_\xi$ uses the ground truth as input instead of predictions from the previous time step. At test time, we do not use teacher forcing.

\begin{figure*}[ht]
\vspace{-10pt}
\centering
\includegraphics[width=1.0\textwidth]{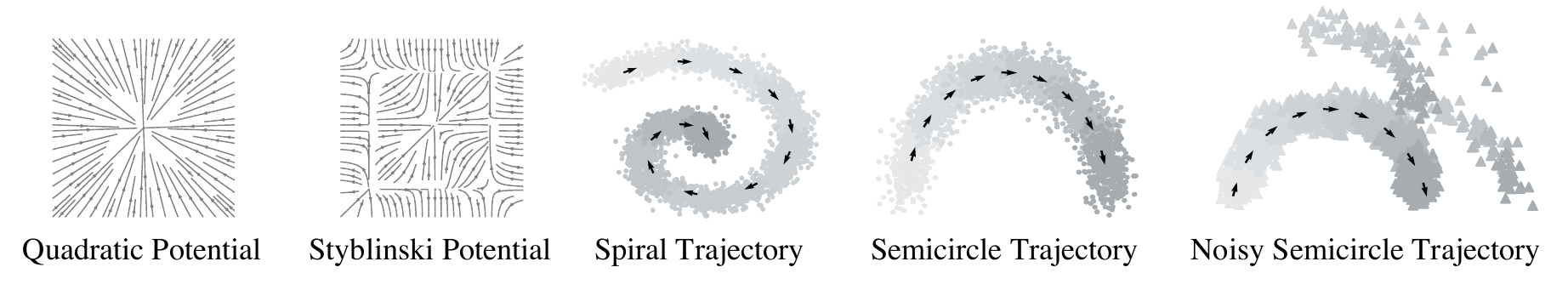}
\caption{Overview on different tasks including trajectory- and potential-based dynamics.}
\label{fig:task_overview}
\end{figure*}

\begin{figure*}[h]
\vspace{-10pt}
\subfloat[\centering Quadratic Potential.]{\includegraphics[width=.22\linewidth]{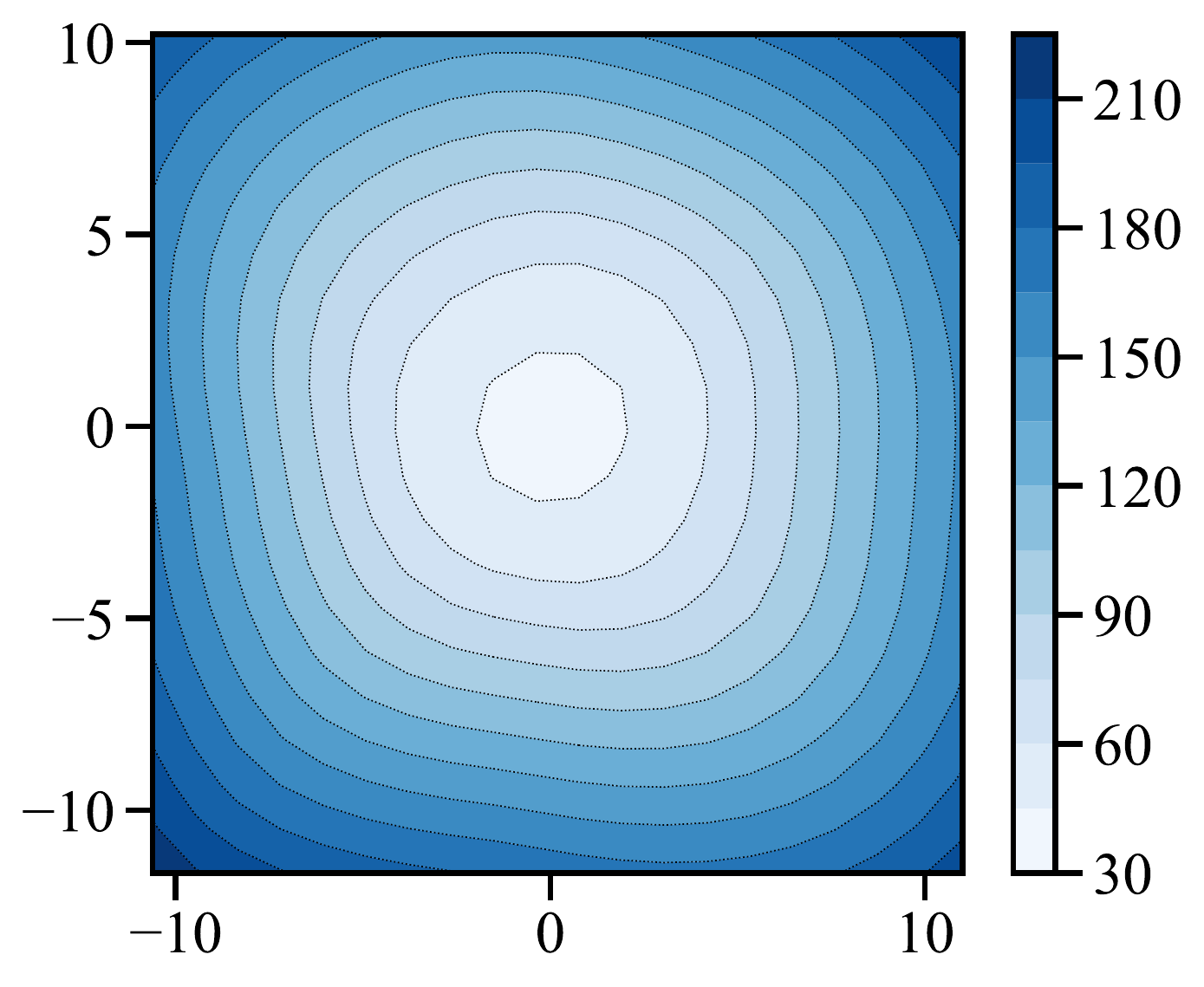}}
\subfloat[\centering Styblinski Potential.]{\includegraphics[width=.22\linewidth]{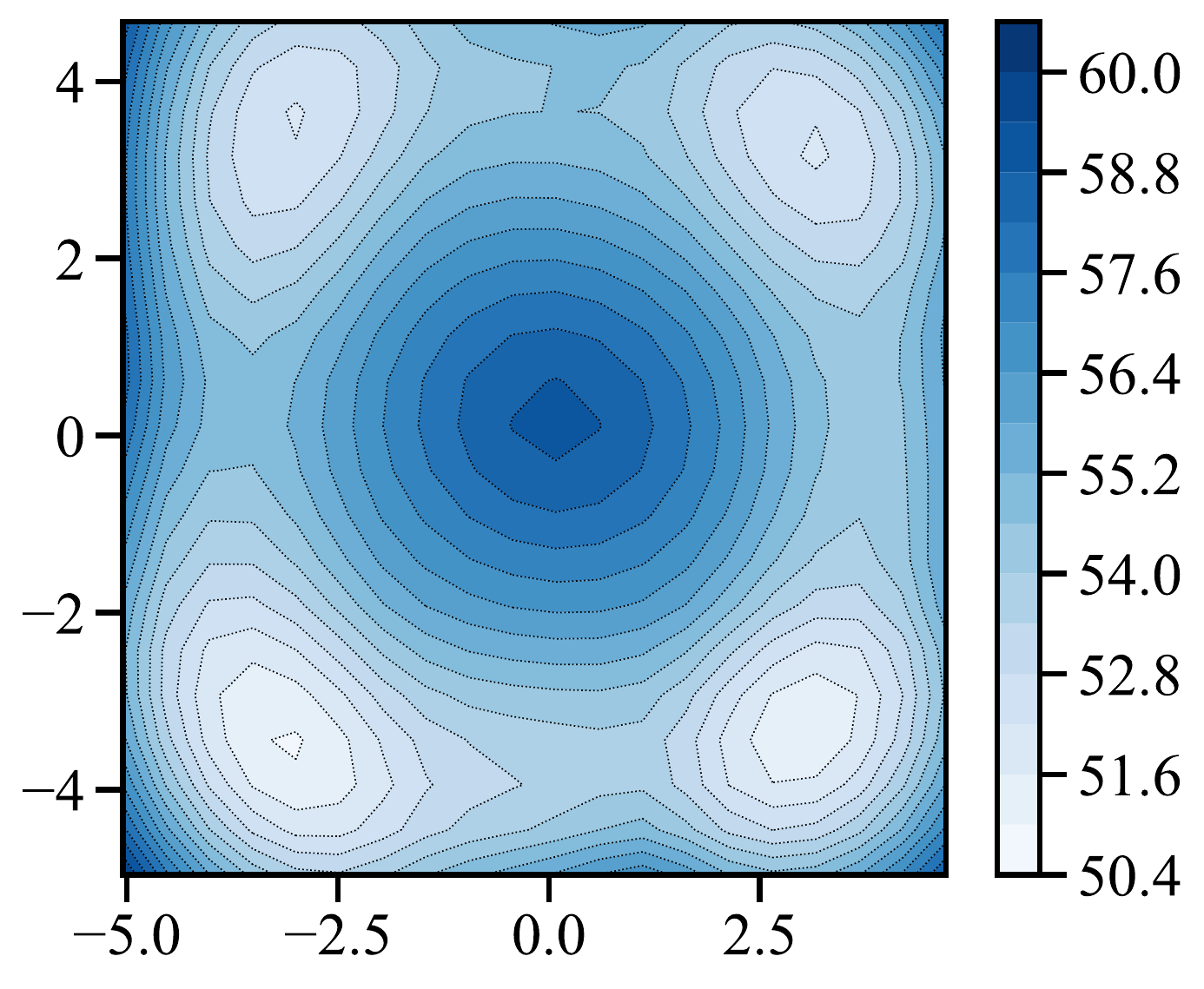}}
\subfloat[\centering Semicircle Trajectory \emph{with} teacher forcing.]{\includegraphics[width=.23\linewidth]{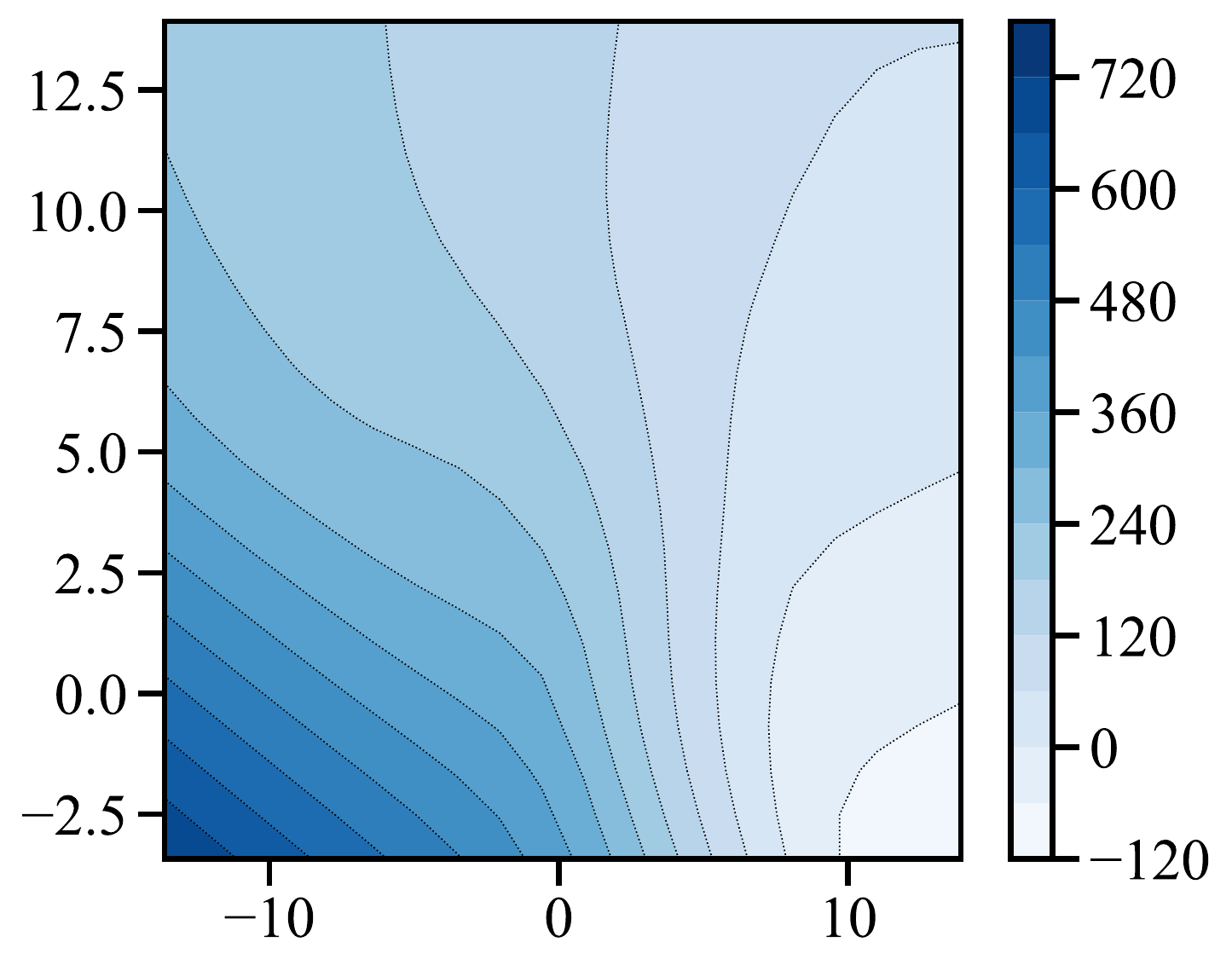}}
\subfloat[\centering Predicted Population Evolution on Semicircle Trajectory.]{\includegraphics[width=.33\linewidth]{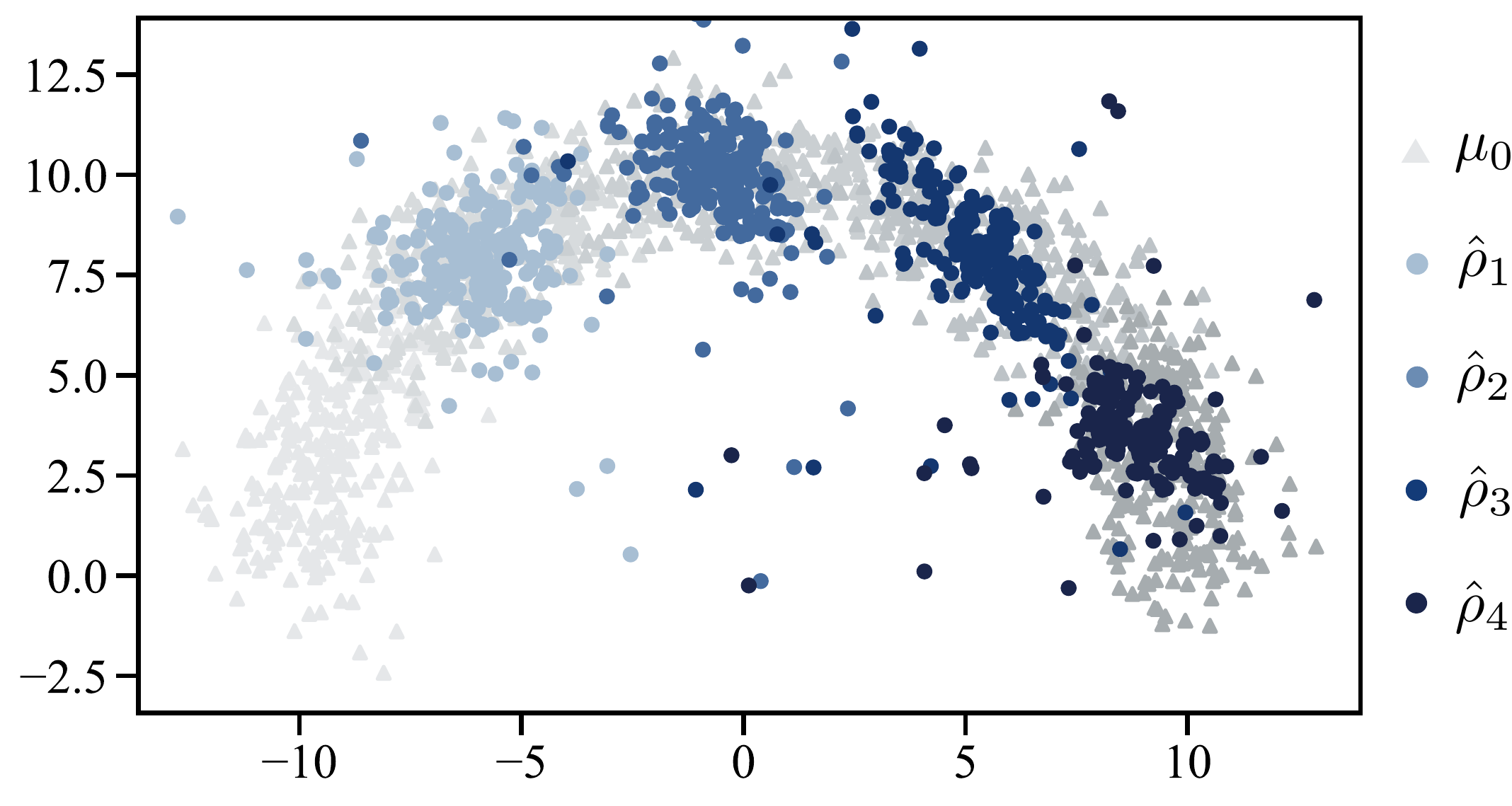}}
\caption{\textbf{Results of \text{JKOnet} on Potential- and Trajectory-based Dyanamics.} (a)-(c) Contour plots of the energy functionals $J_\xi$ of \textsc{JKOnet} on potential- and trajectory-based population dynamics in different training settings (i.e., trained with or without teacher forcing \S~ \ref{sec:learn_energy}), color gradients depict the magnitude of $J_\xi$. (d) Predicted population snapshots ($\hat{\rho}_1, \dots, \hat{\rho}_4$) (blue) and data trajectory ($\mu_0, \dots, \mu_4)$ (gray).}
\label{fig:exp_jkonet_pot_traj}
\end{figure*}

\subsection{Bilevel Formulation of \textsc{JKOnet}}
Learning the free energy functional $J_\xi$ while solving each JKO step via an ICNN results in a challenging bilevel optimization problem.
At each time step, the predicted dynamics are compared to the ground truth trajectory $(\mu_0, \mu_1, \dots, \mu_T)$ with a Sinkhorn loss \eqref{eq:sinkhorn},
\begin{align}\label{eq:fittingloss}
\begin{split}
    \min_\xi & \sum_{t=0}^{T-1} \cW(\rho_{t+1}(\xi), \mu_{t+1}), \\
    \text{s.t. } & \rho_{0}(\xi) := \mu_0, \\
      & \rho_{t+1}(\xi) := \nabla \psi_{\theta^\star\, \#}\, \rho_{t}(\xi)\,, \\
      & \theta^\star:=\arg \min_{\theta} \mathcal{F}_{J_{\xi}}(\psi_{\theta},\rho_t(\xi))
\end{split}
\end{align}
The dependence of the Sinkhorn divergence losses in \eqref{eq:fittingloss} on $\xi$ only appears in the fact that the predictions $\rho_{t+1}(\xi)$ are themselves implicitly defined as solving a JKO step parameterized with the energy $J_\xi$. 
Learning  $J_\xi$ through the exclusive supervision of data observations requires therefore to differentiate the arg-minimum of a JKO problem, down therefore through to the lower-level optimization of the ICNN. We achieve this by implementing a differentiable double loop in \texttt{JAX}, differentiating first the Sinkhorn divergence using the \texttt{OTT}\footnote{\href{https://github.com/ott-jax/ott}{github.com/ott-jax/ott}} package \citep{cuturi2022optimal}, and then backpropagating through the ICNN optimization by unrolling Adam steps \citep{kingma2014adam, metz2016unrolled, lorraine2020}.

\begin{figure*}[ht]
\subfloat[\centering \textsc{JKOnet} on 30\% corrupted data.]{\includegraphics[width=.28\linewidth]{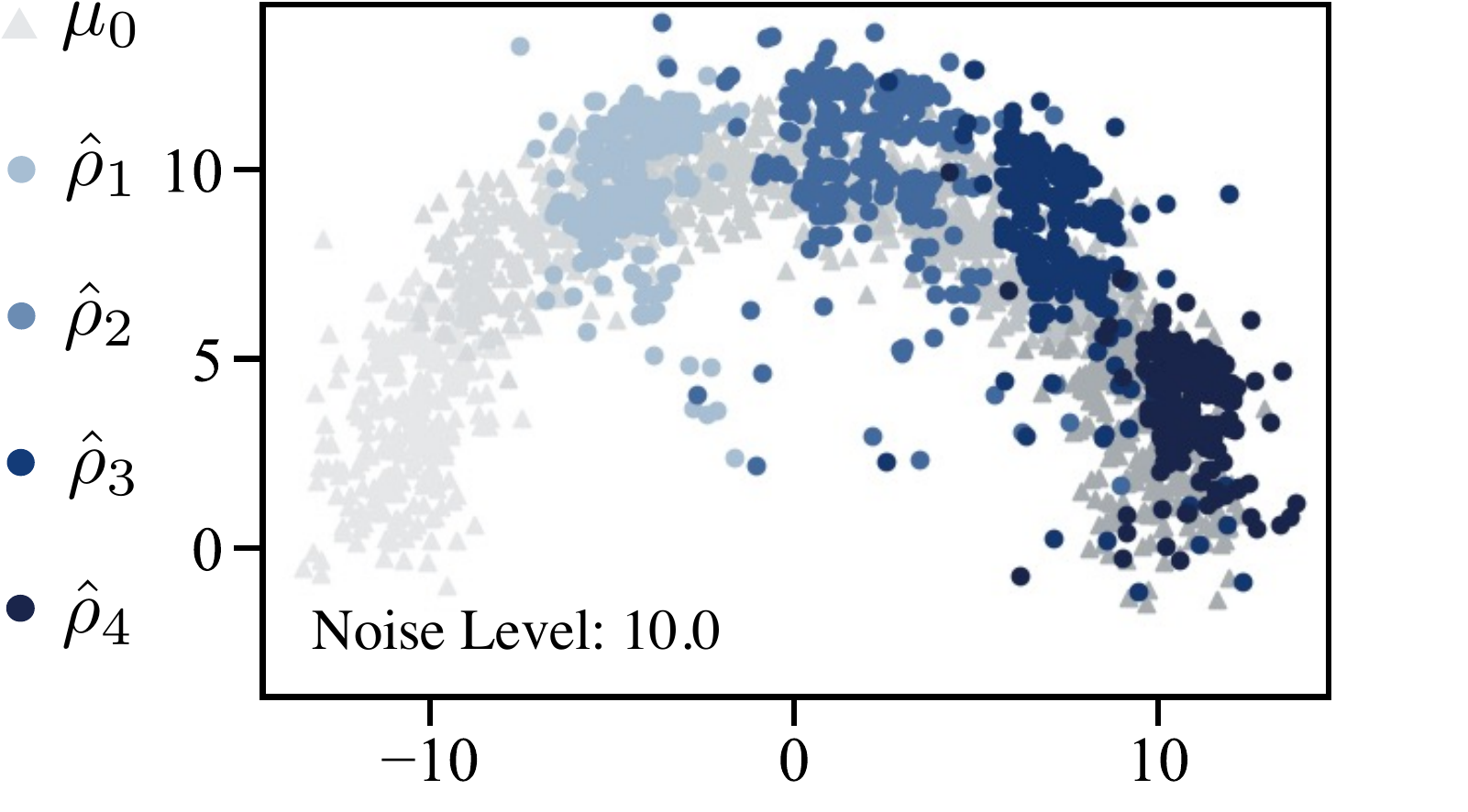}}
\subfloat[\centering Forward method on 30\% corrupted data.]{\includegraphics[width=.26\linewidth]{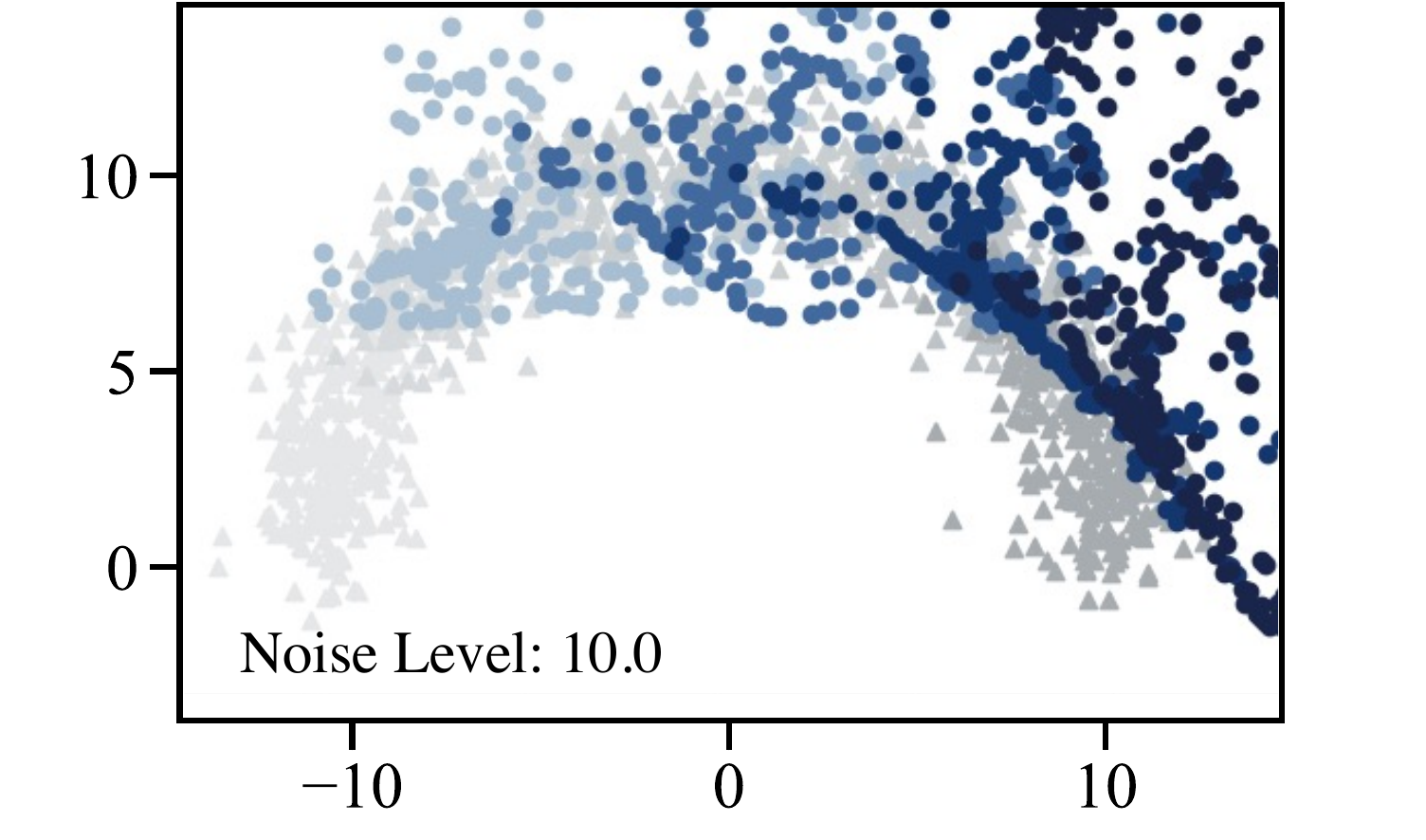}}
\subfloat[\centering $W_\epsilon$ \eqref{eq:reg-ot} vs. noise level on 20\% corrupted data.]{\includegraphics[width=.23\linewidth]{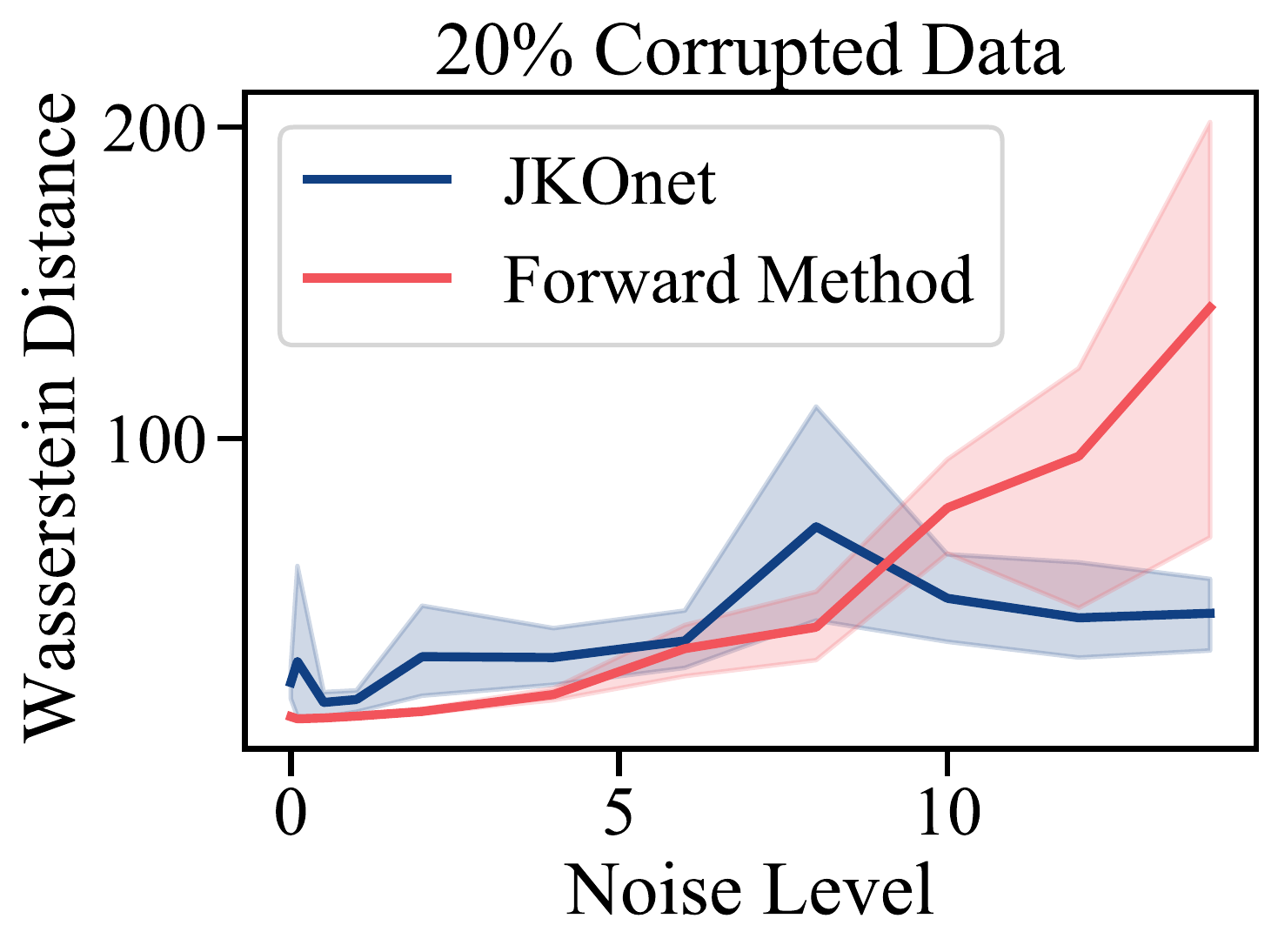}}
\subfloat[\centering $W_\epsilon$ \eqref{eq:reg-ot} vs. noise level on 30\% corrupted data.]{\includegraphics[width=.22\linewidth]{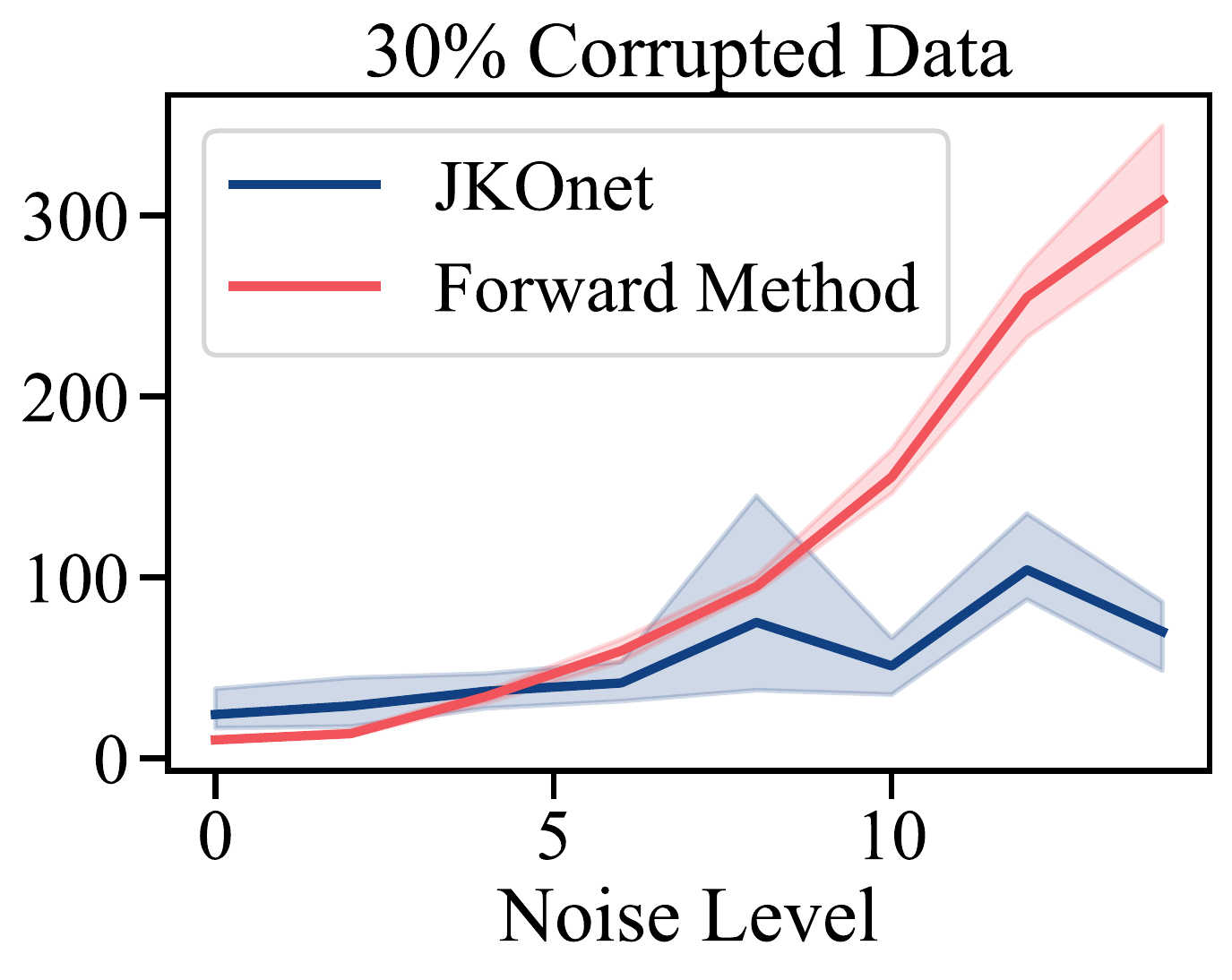}}
\caption{Comparison between \textsc{JKOnet} and the forward method in settings of increasing noise on corrupted data on the semicircle trajectory task.}
\label{fig:exp_comp_noise}
\end{figure*}

\begin{figure*}[ht]
\subfloat[\centering Forward method, \protect\newline \emph{with} teacher forcing.]{\includegraphics[width=.25\linewidth]{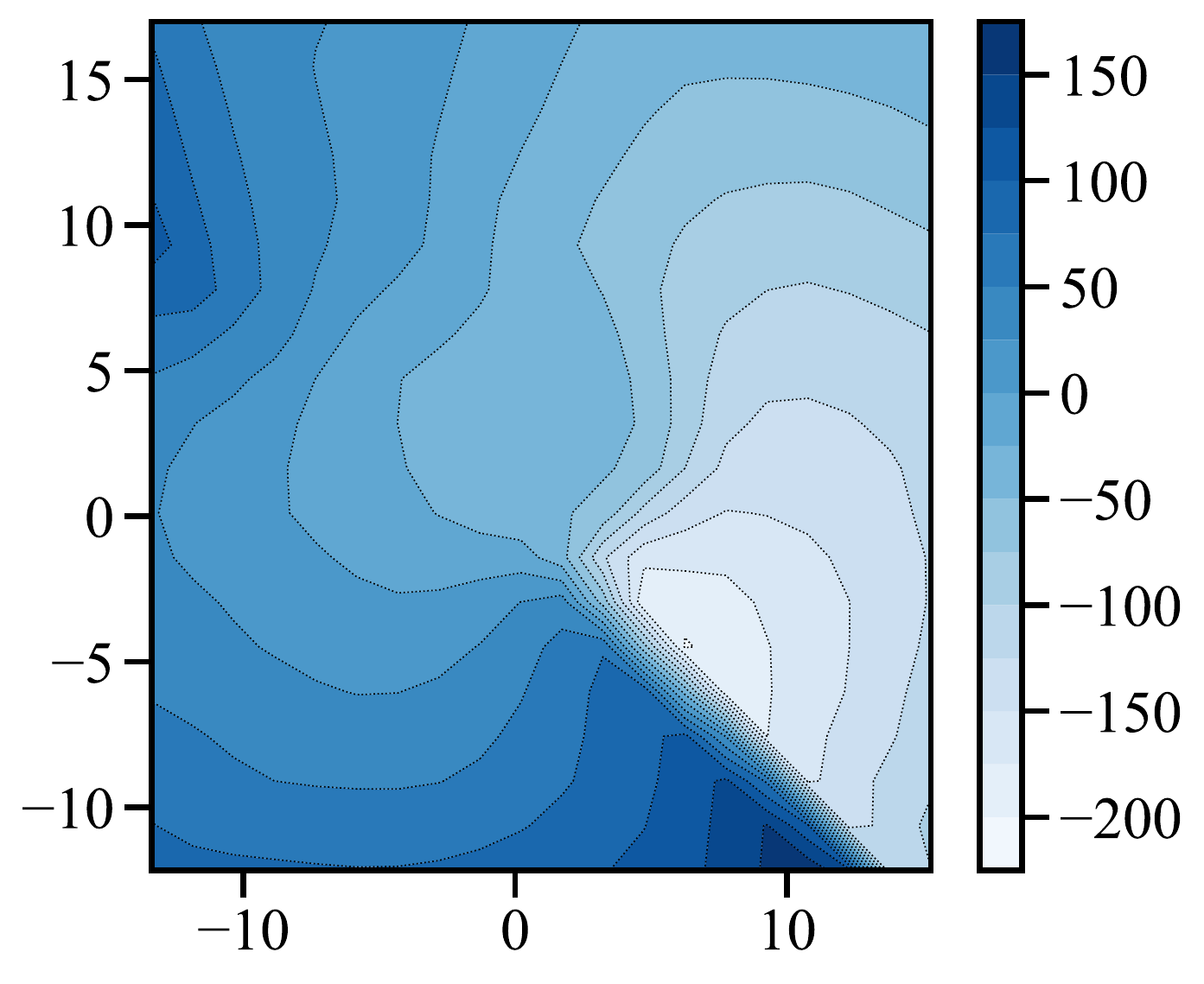}}
\subfloat[\centering Forward method, \protect\newline \emph{no} teacher forcing.]{\includegraphics[width=.25\linewidth]{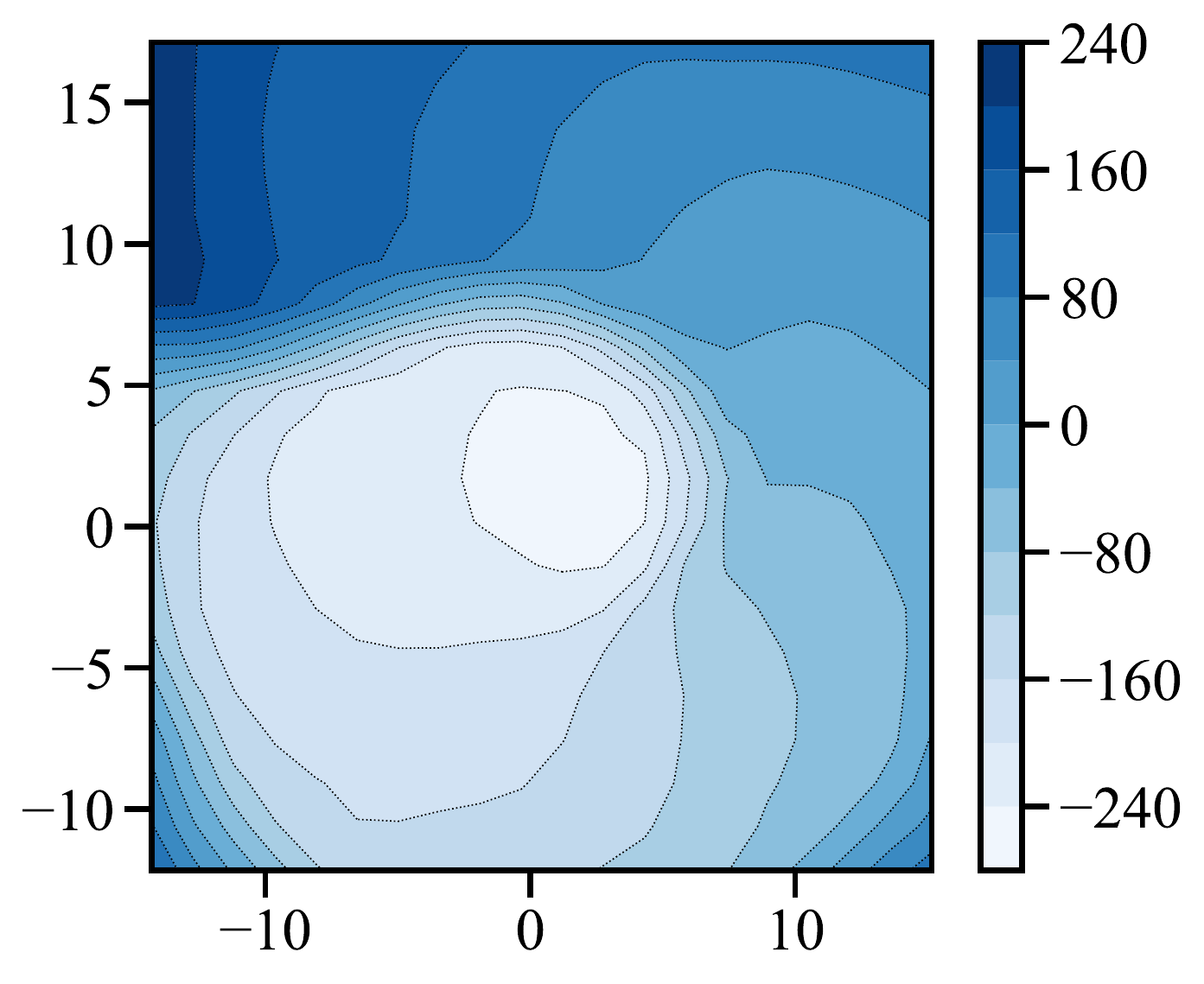}}
\subfloat[\centering \textsc{JKOnet}, \protect\newline \emph{with} teacher forcing.]{\includegraphics[width=.25\linewidth]{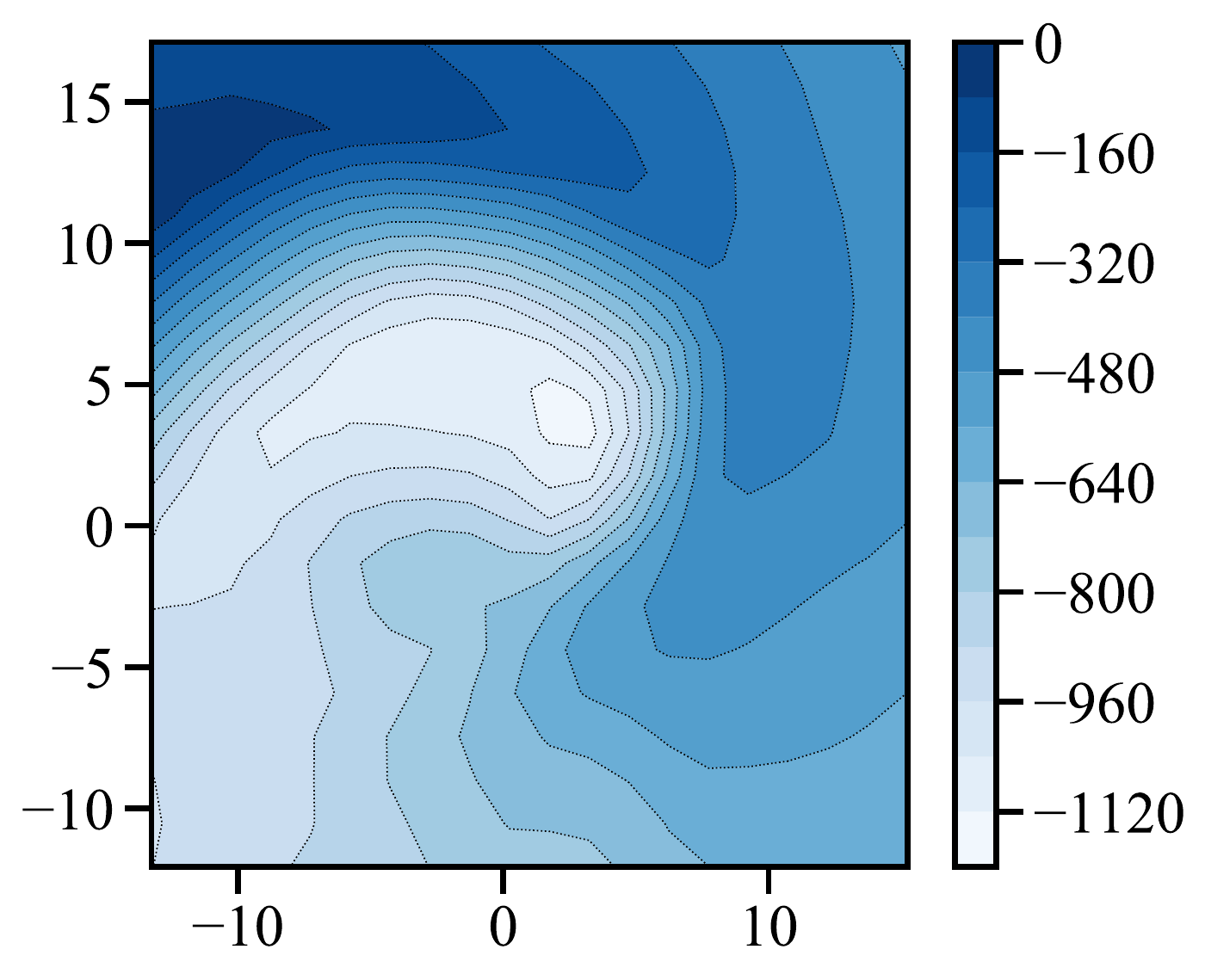}}
\subfloat[\centering \textsc{JKOnet}, \protect\newline \emph{no} teacher forcing.]{\includegraphics[width=.25\linewidth]{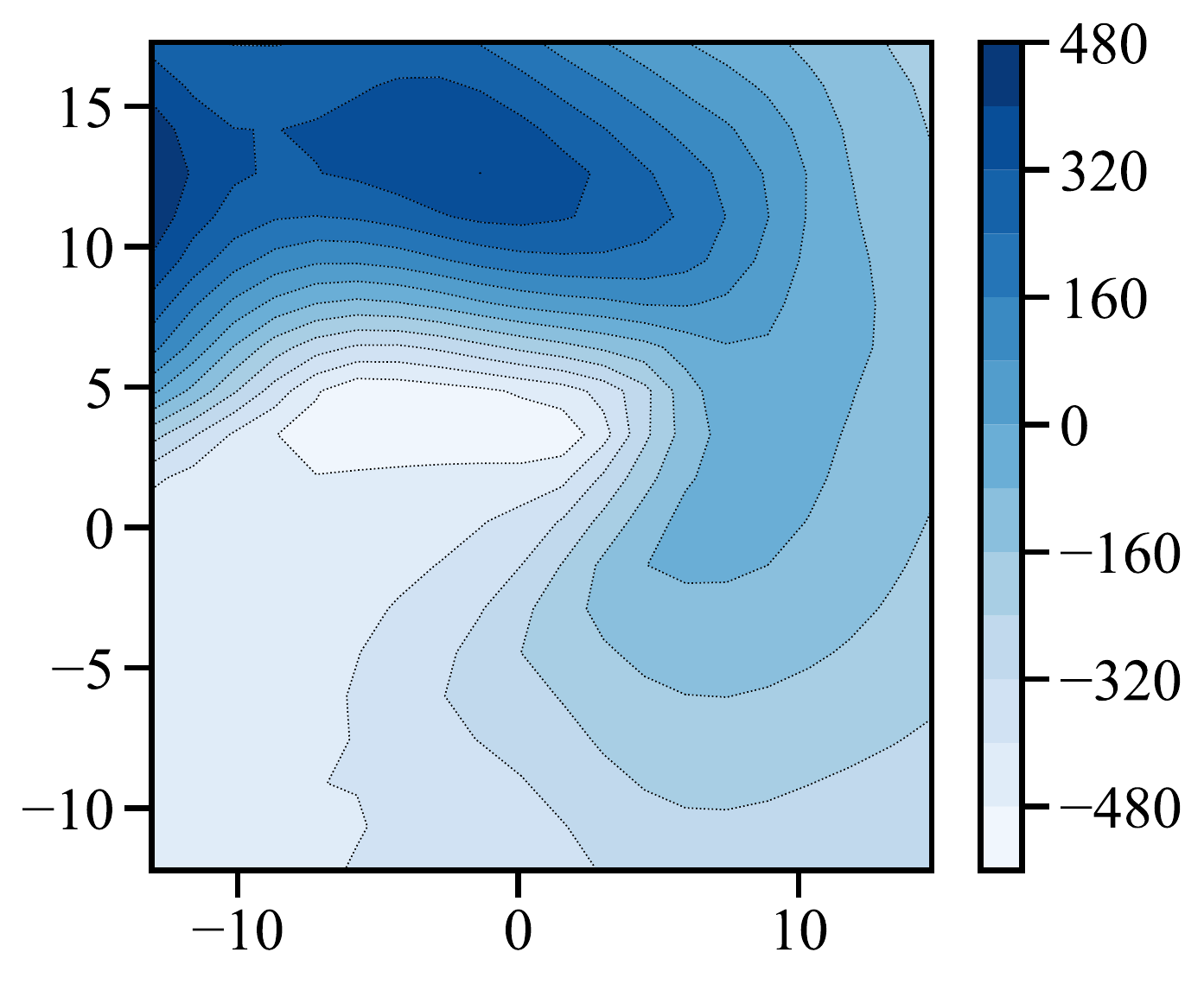}}
\caption{Comparison between energy functionals $J_\xi$ of the spiral trajectory task (see \ref{fig:task_overview}) between the forward method and \textsc{JKOnet}, trained with or without teacher forcing \S~\ref{sec:learn_energy}). When using teacher forcing, the forward method overfits a gap on the lower-right corner of the spiral, outputting a highly irregular energy. When taking into account the entire trajectory recursively, the Forward method does better overall, but is unable to recover an energy as precise as that returned by \textsc{JKOnet}.}
\label{fig:exp_comp_spiral}
\end{figure*}

\newpage
\textbf{Inner Loop Termination.} A question that arises when defining $\rho_{t+1}(\xi)$ lies in the budget of gradient steps needed or allowed to optimize the parameters $\theta$ of the ICNN, before taking a new gradient step on $\xi$ in the outer loss. A straightforward approach in \texttt{JAX} \citep{jax2018github} would be to use a preset number of iterations with a \texttt{for} loop (\texttt{jax.lax.scan}). 
We do observe, however, that the number of iterations needed to converge in relevant scenarios can vary significantly with the ICNN architecture and/or the hardness of the underlying task.
We propose to use instead a differentiable fixed-point loop to solve each JKO step up to a desired convergence threshold.
We measure convergence of the optimization of the ICNN via the average norm of the gradient of the JKO objective w.r.t. the ICNN parameters $\theta$, i.e., $\sum_i \norm{\nabla_{\theta_i}\mathcal{F}_{J_\xi}(\theta_i, \xi)}_2/\sum_i \text{count}(\theta_i)$.
We observe that this approach is robust across datasets and architectures of the ICNN. An exemplary training curve for the ICNNs updated successively along a time sequence is shown in Figure~\ref{fig:training_icnn}.

\textbf{Reverse-Mode Differentiation.} The Jacobian $\partial \rho_{t+1} / \partial\xi$ arising when computing the gradient $\nabla_\xi \cW(\rho_{t+1}(\xi), \mu_{t+1})$ is obtained by unrolling the while loop above. The gradient term of the Sinkhorn divergence w.r.t the first argument is given by the Danskin envelope theorem \citep{danskin2012theory}.

% The Jacobian $\partial \rho^\xi_{t+1} / \partial\xi$ that appears when computing $\nabla_\xi \cW(\text{data}_{t+1}, \rho_{t+1}(\xi))$ is computed by unrolling the iterations of the while loop above.

\textbf{Setting $\tau$ in \eqref{eq:JKO_psi}.} 
In usual JKO applications, $\tau$ needs to be tuned manually. In this work, the energy $J_\xi$ is not fixed, but trained to fit data. Since we put no constraints on the scaling of $J_\xi$, $\tau$ can be set to $1$ without loss of generality, as the parameter $\xi$ will automatically adjust so that the scale of $J_\xi$ induces steps of a relevant length to fit data. This only holds (as with a usual JKO step) if the trajectories are sampled regularly. For irregularly spaced time series, $\tau$ can be adapted at train and test time to the spacing of timestamps (shorter steps requiring larger $\tau$).

\begin{table*}[t]
    \caption{Evaluation of predictive performance w.r.t. the entropy-regularized Wasserstein distance $W_\varepsilon$ \eqref{eq:reg-ot} of \textsc{JKOnet} and the forward method on the embryoid body scRNA-seq data per time step (using 3 runs).}
    \label{tab:exp_jkonet_cell_pred}
    \centering
\adjustbox{max width=.75\linewidth}{%
    \begin{tabular}{lcccc}
    \toprule
         \textbf{Method} & \multicolumn{4}{c}{\textbf{Prediction Loss ($W_\varepsilon$})} \\
         \cmidrule{2-5}
         & Day 6 to 9 & Day 12 to 15 & Day 18 to 21 & Day 24 to 27 \\
    \midrule
    \textbf{One-Step Ahead} \\
        \tabindent Forward Method & $0.187 \pm 0.001$ & $0.162 \pm 0.010$ & $0.185 \pm 0.020$ & $0.203 \pm 0.004$ \\
        \tabindent \textsc{JKOnet} & $\bf{0.133 \pm 0.020}$ & $\bf{0.133 \pm 0.008}$ & $\bf{0.172 \pm 0.0130}$ & $\bf{0.169 \pm 0.004}$ \\
    \textbf{All-Steps Ahead} \\
        \tabindent Forward Method & $0.225 \pm 0.023$ & $0.160 \pm 0.001$ & $0.171 \pm 0.016$ & $0.183 \pm 0.007$ \\
        \tabindent \textsc{JKOnet} & $\bf{0.148 \pm 0.015}$ & $\bf{0.144 \pm 0.013}$ & $\bf{0.154 \pm 0.024}$ & $\bf{0.138 \pm 0.034}$ \\
    \bottomrule
    \end{tabular}
}
\end{table*}

\section{Evaluation} \label{sec:evaluation}
In the following, we evaluate our method empirically on a variety of tasks. This includes recovering synthetic potential- and trajectory-based population dynamics (see Fig.~\ref{fig:task_overview}), as well as the evolution of high-dimensional single-cell populations during a developmental process. 

\subsection{Synthetic Population Dynamics} \label{sec:eval_synt}
\paragraph{Energy-Driven Trajectories.} The first task involves evolutions of partial differential equations with known potential. We hereby consider both convex (e.g., the quadratic function $J(x) = \|x\|^2_2$) and nonconvex potentials (e.g., Styblinski function) (see Fig.~\ref{fig:task_overview}). These two-dimensional synthetic flows are generated using the Euler-Maruyama method~\citep{kloeden1992stochastic}. For details, see \S~\ref{app:potential_dataset}.
To recover the true potential via \textsc{JKOnet}, we parameterize both energy $J_\xi$ and ICNN $\psi_\theta$ with linear layers ($\epsilon = 1.0$, $\tau = 1.0$, \S~\ref{app:hyperparam}). More details on the architectures can be found in \S~\ref{app:architecure}.
Figure~\ref{fig:exp_jkonet_pot_traj}a-b demonstrate \textsc{JKOnet}'s ability to recover convex and nonconvex potentials via energy $J_\xi$.

\paragraph{Arbitrary Trajectories.} As a sanity check, we evaluate if \textsc{JKOnet} can recover an energy functional $J_\xi$ from trajectories that are not necessarily arising from the gradient of an energy. Here, a 2-dimensional Gaussian moves along a predefined trajectory with nonconstant speed. 
For details on the data generation, see \S~\ref{app:trajectory_dataset}.
We consider a line, a spiral, and movement along a semicircle (Fig.~\ref{fig:task_overview}). As visible in Figure~\ref{fig:exp_jkonet_pot_traj}c (5 snapshots), Figure~\ref{fig:exp_comp_line}b (2 snapshots), and Figure~\ref{fig:exp_comp_spiral}c-d (10 snapshots), \textsc{JKOnet} learns energy functionals $J_\xi$ that can then model the ground truth trajectories.
These trajectory-based dynamics are learned using the strong convexity regularizer ($\ell=0.8$, see \S~\ref{sec:jko_icnn}).

\textbf{Comparison to Forward Methods.} \label{sec:eval_comp_fb}
Instead of parameterizing the next iteration $\rho_{t+1}(\xi)$ as we do in the \textsc{JKOnet} formulation~\eqref{eq:jko}, the \emph{forward} scheme states that the prediction at time $t+1$, $\eta_{t+1}$, can be obtained as $(\nabla F_\xi)_{\#} \eta_t(\xi)$, where $F_\xi$ is any arbitrary neural network, as considered in \citet{hashimoto2016learning}, namely $\eta_0:=\mu_0$ and subsequently $\eta_{t+1}(\xi):=(\nabla F_\xi)_{\#} \eta_t(\xi)$. Although OT still plays an important role in that paper, since the potential $F$ is estimated by minimizing a Sinkhorn loss $\cW(\eta_{t+1},\mu_{t+1})$, as we do in \eqref{eq:fittingloss}, the forward displacement operator $(\nabla F_\xi)_{\#}$ has no spatial regularity. Because of that, we observe that the forward method can get more easily trapped in local minima, and, in particular, overfits the training data (see \S~\ref{app:overfitting}) as shown by a substantial decrease in performance in the presence of noise.
We demonstrate this in different scenarios: 
First, we compare the robustness of both \textsc{JKOnet} and the forward method to noise. For this, we corrupt $20\%$ or $30\%$ of the training data on the example of the semicircle trajectory with different levels of noise (see Fig.~\ref{fig:task_overview}). We insist that noise is only added at training time, as random shifts on both feature dimensions, while we test on the original semicircle trajectory.
In low noise regimes, where train and test data are similar, the forward method overfits and performs marginally better than \textsc{JKOnet} (see Fig.~\ref{fig:exp_comp_noise}c,d). As noise increases, the performance of the forward method deteriorates (Fig.~\ref{fig:exp_comp_noise}b), while \textsc{JKOnet}, constrained to move points with OT maps, is robust (Fig.~\ref{fig:exp_comp_noise}a).% This shows that the forward method is able to learn a network such that, on average, the ensemble of particles $(\nabla F_\xi)_{\#} \rho_t$ fits $\mu_{t+1}$, without, however

In a second experiment, we evaluate the capacity of \textsc{JKOnet} and the forward method to extrapolate and generalize the learned trajectories, e.g., when vertically translating a line during test time (Fig.~\ref{fig:task_overfitting}).
Due to the less constrained energy, the \emph{forward} method perfectly resembles the seen trajectory during training, but fails to extrapolate to shifted test data (Table~\ref{tab:comp_line} in \S~\ref{app:overfitting}).

Lastly, we compare the resulting energy functionals $F_\xi$ and $J_\xi$ of the forward method and \textsc{JKOnet}, respectively, on the spiral trajectory (see Fig.~\ref{fig:exp_comp_spiral}).
When learning long and complex population dynamics, teacher forcing improves training (see additional results in Fig.~\ref{fig:exp_forward_pot_traj}c-d as well as Fig.~\ref{fig:exp_jkonet_pot_traj}c-d).
While facilitating training of the forward method in some settings, it likewise results in wrong energy functionals $F_\xi$ (Fig.~\ref{fig:exp_comp_spiral}a).
\textsc{JKOnet}, on the other hand, is able to globally learn the energy functional $J_\xi$, despite being only exposed to a one-step history of snapshots during training with teacher forcing (see Fig.~\ref{fig:exp_comp_spiral}c).

\subsection{Single-Cell Population Dynamics} \label{sec:eval_cell}

\begin{table*}[t]
\begin{minipage}{0.58\linewidth}
    \subfloat[\centering PCA embedding of the embryoid body scRNA-seq data colored by the snapshot time.]{\includegraphics[width=.5\linewidth, valign=t]{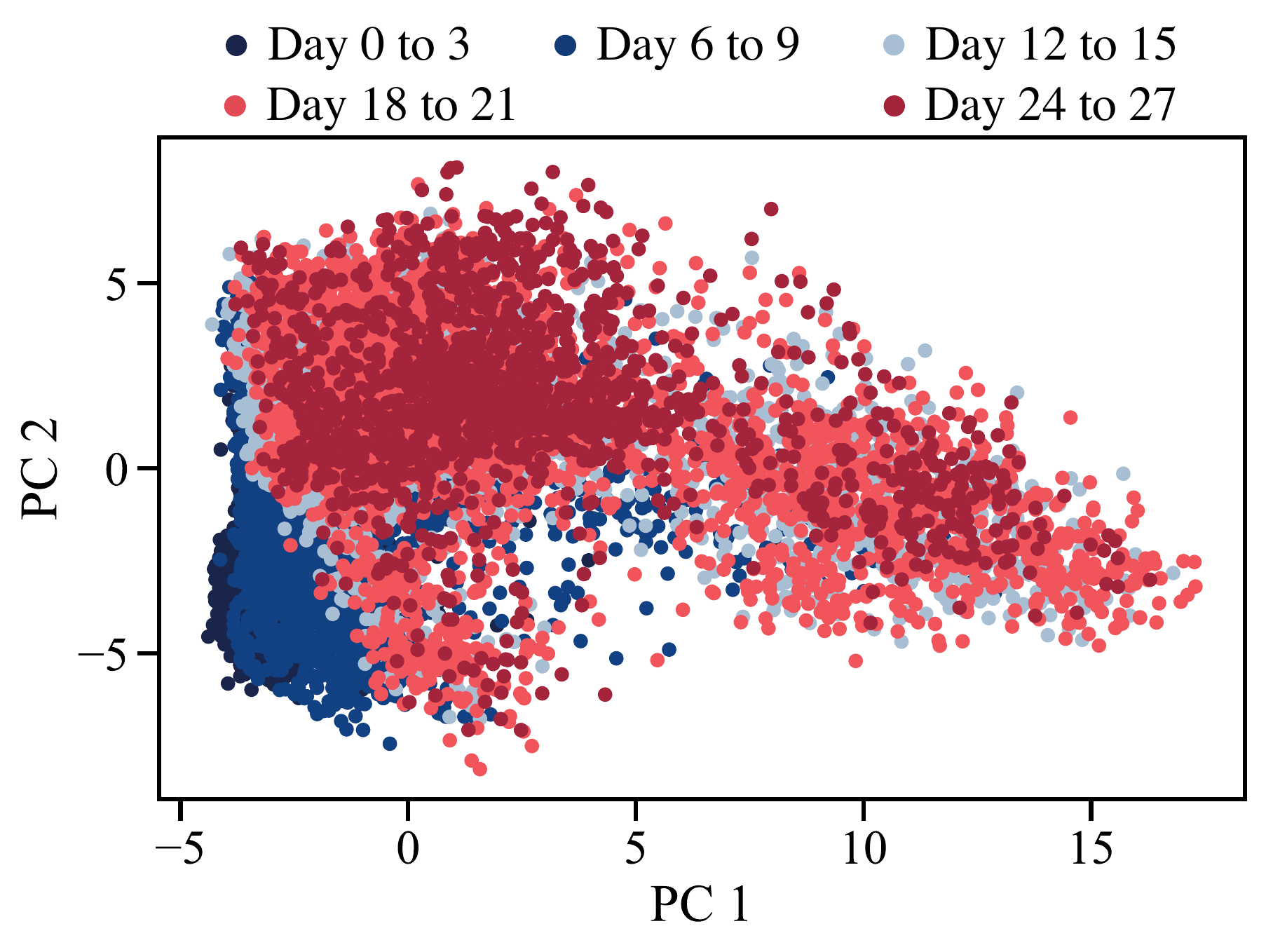}}
    \subfloat[\centering PCA embedding of the embryoid body scRNA-seq data colored by the lineage branch class.]{\includegraphics[width=.5\linewidth, valign=t]{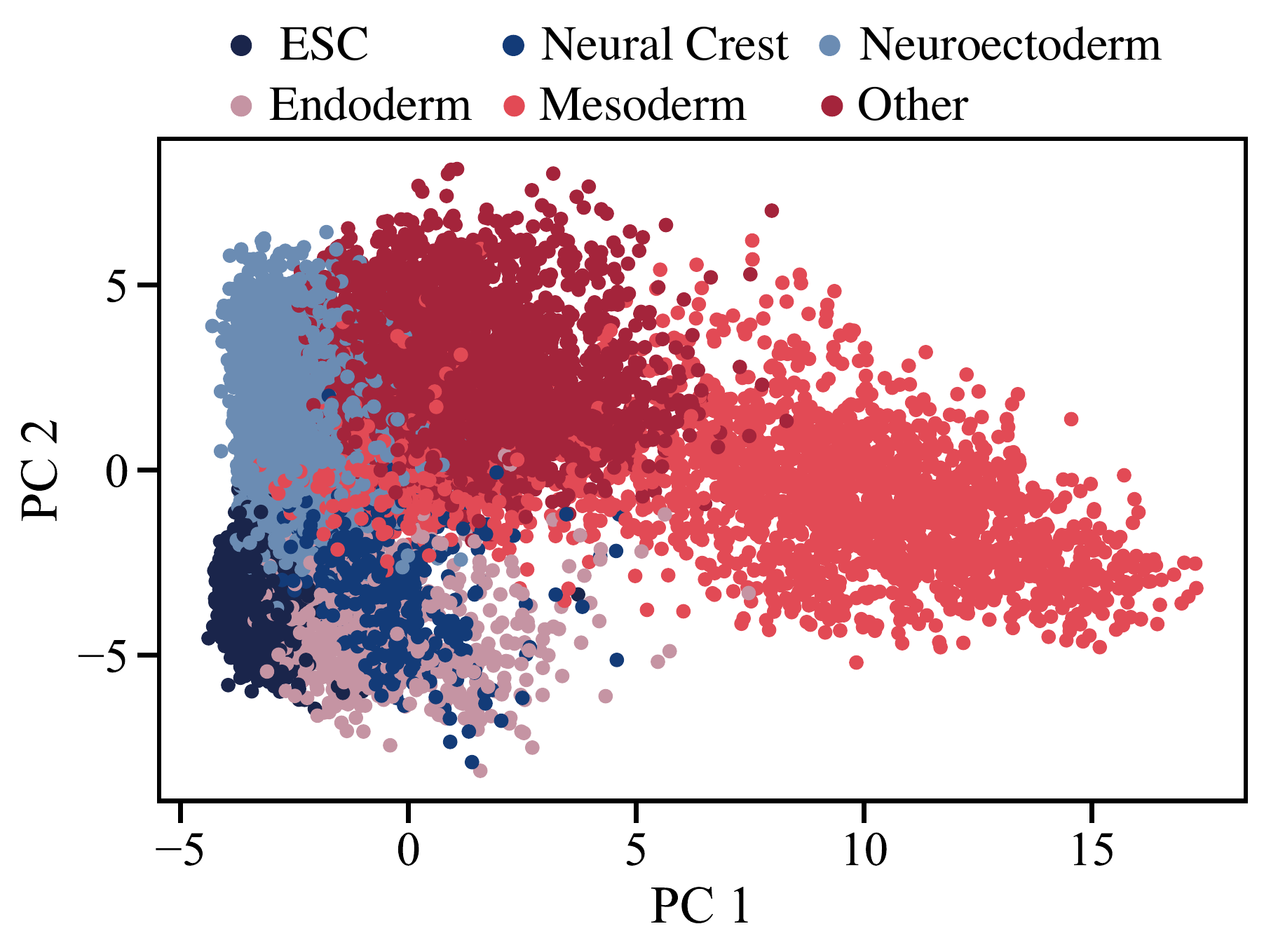}}
\end{minipage}\hfill
\begin{minipage}{0.39\linewidth}
    \subfloat[\centering Distribution of cell lineage branch classes in the data or  predicted by \textsc{JKOnet} or the forward method.]{\includegraphics[width=.9\linewidth, valign=t]{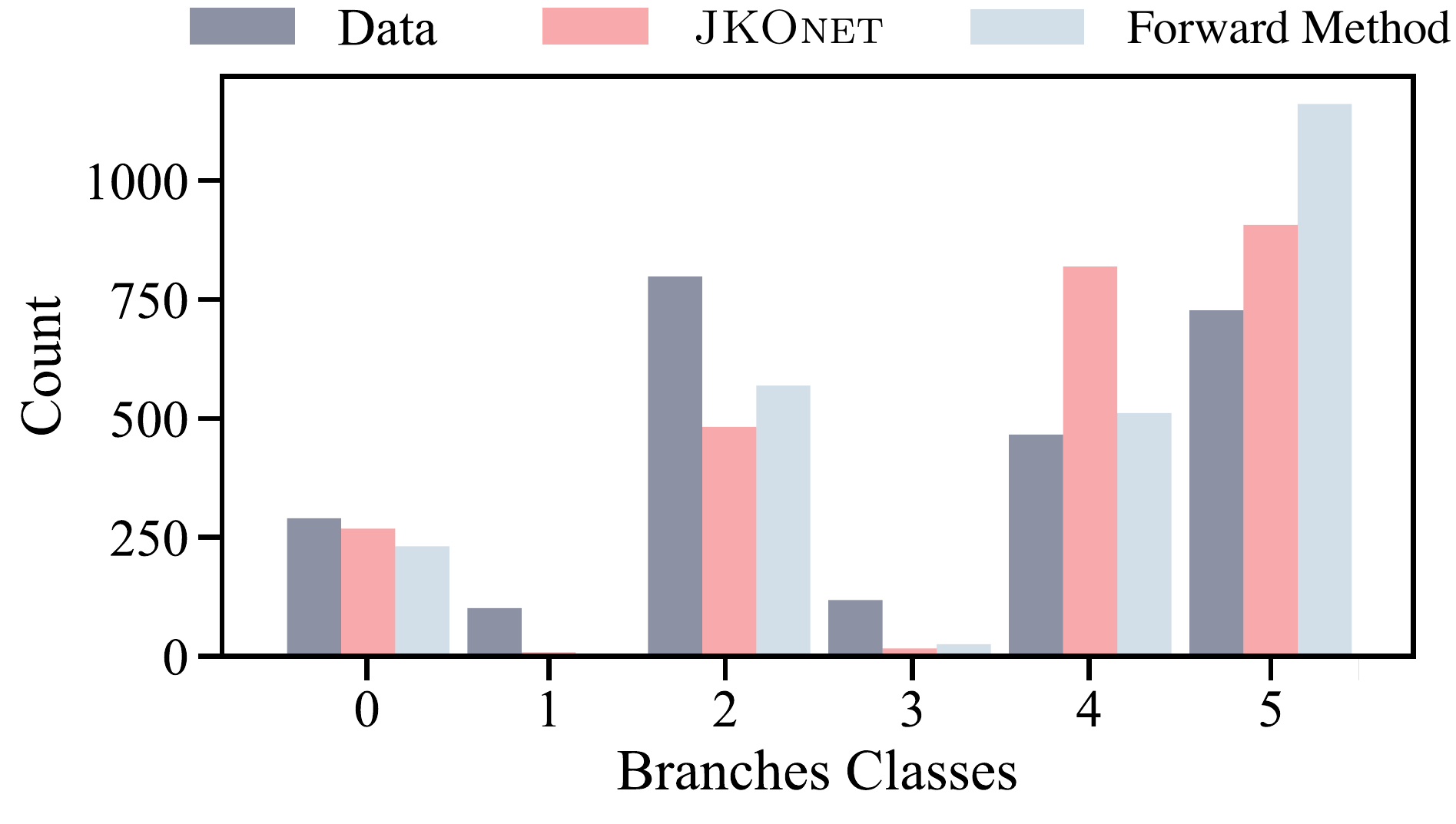}}
\end{minipage}\hfill

\begin{minipage}{0.58\linewidth}
    \subfloat[\centering PCA embedding of \textsc{JKOnet} predictions colored by the snapshot time.]{\includegraphics[width=.5\linewidth, valign=t]{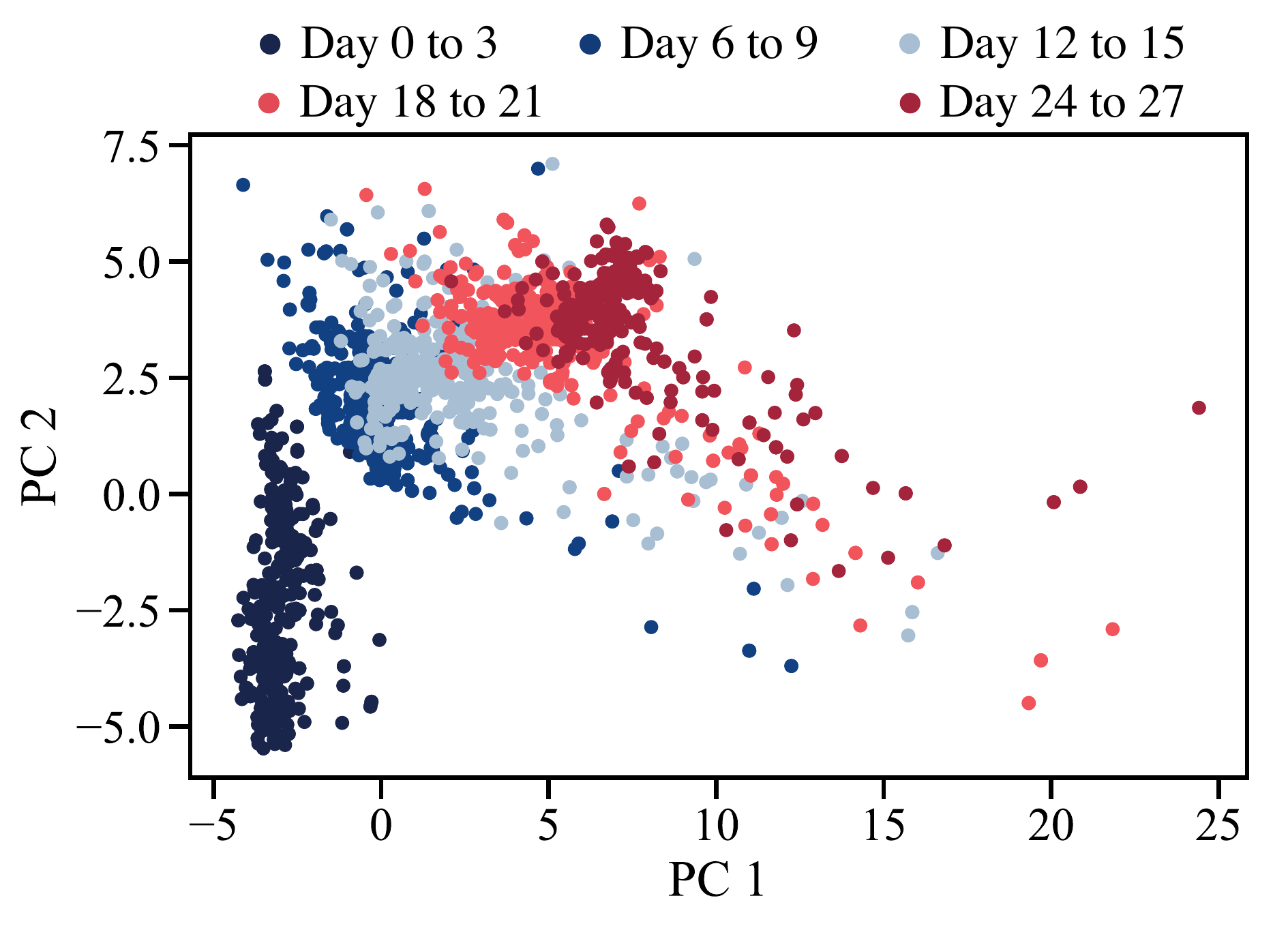}}
    \subfloat[\centering  PCA embedding of \textsc{JKOnet} predictions colored by the lineage branch class.]{\includegraphics[width=.5\linewidth, valign=t]{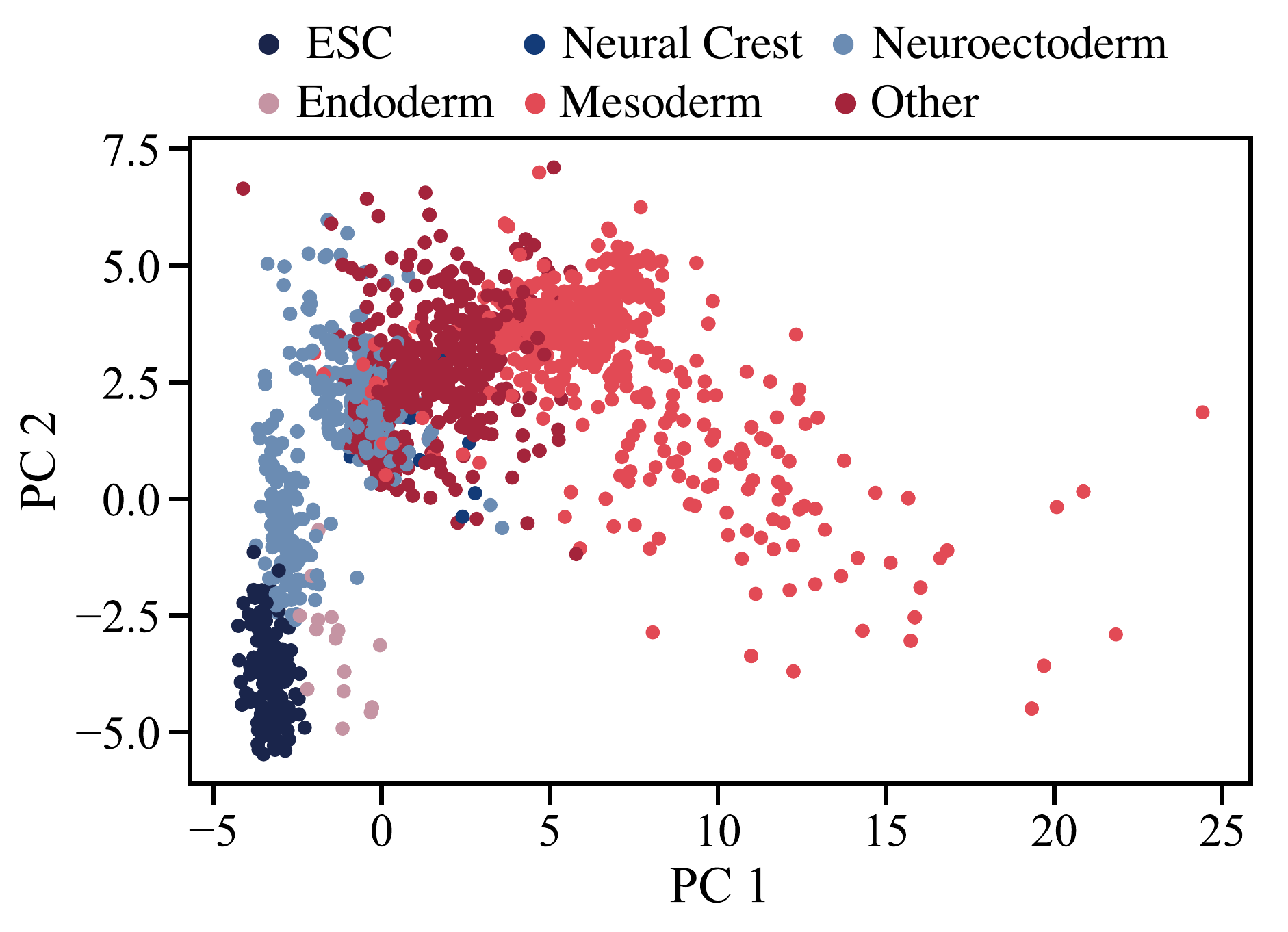}}
    \captionof{figure}{Analysis of population dynamics predictions of \textsc{JKOnet} on the embryoid body scRNA-seq data.}
    \label{fig:exp_jkonet_cell}
\end{minipage}\hfill
\begin{minipage}{0.39\linewidth}
% \vspace*{-\dimexpr\baselineskip+\heavyrulewidth+\abovetopsep\relax}

\captionof{table}{Evaluation of cell lineage branch classification performance of \textsc{JKOnet} and the forward method on the embryoid body scRNA-seq data based on the $\ell_1$-distance of the histograms and the Hellinger distance $H^2$ \eqref{eq:hellinger} of the predicted branch class distributions (using 3 runs).}
\label{tab:exp_jkonet_cell_class}
\centering
	\resizebox{\linewidth}{!}{%
    \begin{tabular}{lcc}
    \toprule
         \textbf{Method} & \multicolumn{2}{c}{\textbf{Cell Lineage Classification}} \\
         & $\ell_1$ & $H^2$ \\
    \midrule
    \textbf{One-Step Ahead} \\
        \tabindent Forward Method & $132.27 \pm 5.00$ & $0.026 \pm 0.002$ \\
        \tabindent \textsc{JKOnet} & $\bf{88.80 \pm 0.57}$ & $\bf{0.016 \pm 0.001}$ \\
    \textbf{All-Steps Ahead} \\
        \tabindent Forward Method & $\bf{185.47 \pm 12.18}$ & $0.033 \pm 0.002$ \\
        \tabindent \textsc{JKOnet} & $215.60 \pm 12.53$ & $0.034 \pm 0.004$ \\
    \bottomrule
    \end{tabular}
}
\end{minipage}
\end{table*}

We investigate the ability of \textsc{JKOnet} to predict the evolution of cellular and molecular processes through time.
The advent of single cell profiling technologies has enabled the generation of high-resolution single-cell data, making it possible to profile individual cells at different states in the development. 
A key difficulty in learning the evolution of cell populations is that a cell is (usually)  destroyed during a measurement. Thus, although one is able to collect features at the level of individual cells, the same cell cannot be measured twice. Instead, we collect independent samples at each snapshot, resulting in \emph{unaligned} distributions across snapshots, without access to ground-truth single-cell trajectories. 
The goal of learning individual dynamics is to identify ancestor and descendant cells, and get a better understanding of biological differentiation or reprogramming mechanisms. 

We apply \textsc{JKOnet} to embryoid body single-cell RNA sequencing (scRNA-seq) data \citep{moon2019}, describing the differentiation of human embryonic stem cells grown as embryoid bodies into diverse cell lineages over a period of 27 days. During this time, cells are collected at 5 different snapshots (day 1 to 3, day 6 to 9, day 12 to 15, day 18 to 21, day 24 to 27) and measured via scRNA-seq (resulting in 15,150 cells). For details on the dataset and data preprocessing see \S~\ref{app:cell_dataset}.
We run \textsc{JKOnet} as well as the baseline on the first 20 components of a principal component analysis (PCA) of the 4000 highly differentiable genes (see Fig.~\ref{fig:moon_expl_variance}).
We split the dataset into train and test data ($\sim 15 \%$) and parameterize both energy $J_\xi$ and ICNN $\psi_\theta$ with linear layers ($\epsilon = 1.0$, $\tau = 1.0$, \S~\ref{app:hyperparam}).

\paragraph{Capturing Spatio-Temporal Dynamics.}
Given the samples from the cell population at day 1 to 3 ($\mu_0$), \textsc{JKOnet} learns the underlying spatio-temporal dynamics giving rise to the developmental evolution of embryonic stem cells. 
As no ground truth trajectories are available in the data, we use distributional distances, i.e., the entropy-regularized Wasserstein distance $W_\varepsilon$ \eqref{eq:reg-ot} \citep{flamary2021pot}, to measure the correctness of the predictions at each time step.
We hereby measure the $W_\varepsilon$ discrepancy between data and predictions for one-step ahead as well as inference of the entire evolution (all-steps ahead) for each time step $t_i$, see results in Table~\ref{tab:exp_jkonet_cell_pred}. \textsc{JKOnet} outperforms the forward method in terms of $W_\varepsilon$ \eqref{eq:reg-ot} distance for both one-step ahead and all-steps ahead predictions for all time steps. 
The performance of both methods is relatively stable even until day 24 to 27, i.e., the $W_\varepsilon$ distance does not significantly grow for future snapshots.
We further visualize the first two principal components of the entire dataset (Fig.~\ref{fig:exp_jkonet_cell}a) and of \textsc{JKOnet}'s predictions on the test dataset ($\sim 500$ cells per snapshot, Fig.~\ref{fig:exp_jkonet_cell}d). Visualization of predictions of the forward method can be found in the Appendix (Fig.~\ref{fig:exp_forward_cell}a). 

\paragraph{Capturing Biological Heterogeneity.}
Besides measuring the ability of \textsc{JKOnet} to model and predict the spatio-temporal dynamics of embryonic stem cells, we would like to guarantee, at a more macroscopic level, that \textsc{JKOnet} is also able to learn the cell's differentiation into various cell lineages.
Embryoid bodies differentiation covers key aspects of early embryogenesis and thus captures the development of embryonic stem cells (ESC) into the mesoderm, endoderm, neuroectoderm, neural crest and others.

Following \citet[Fig. 6, Suppl. Note 4]{moon2019}, we compute lineage branch classes (Fig.~\ref{fig:moon_analysis}c) for all cells based on an initial $k$-means clustering ($k=30$) in a 10-dimensional embedding space using PHATE, a non-linear dimensionality reduction method capturing a denoised representation of both local and global structure of a dataset (Fig.~\ref{fig:moon_analysis}b).
For details, see \S~\ref{app:lineage_analysis}.
We then train a $k$-nearest neighbor ($k$-NN) classifier ($k=5$) to infer the lineage branch class based on a 20-dimensional PCA embedding of a cell (classes: ESC: 0, neural crest: 1, neuroectoderm: 2, endoderm: 3, mesoderm: 4, other: 5).

We analyze the captured lineage branch heterogeneity of the population predicted by \textsc{JKOnet} and the forward method by estimating the lineage branch class of each cell using the trained $k$-NN classifier. The predicted populations colored by the estimated lineage branch as well as the data with the true lineage branch labels are visualized in Figure~\ref{fig:exp_jkonet_cell}e and Figure~\ref{fig:exp_jkonet_cell}b, respectively.
The corresponding predicted and true distributions of lineage branch classes are shown in Figure~\ref{fig:exp_jkonet_cell}c.
To quantify how well \textsc{JKOnet} and the forward method capture  different cell lineage branches, we compute the $\ell_1$ distance between the  predicted and true histograms as well as the Hellinger distance 
\begin{equation} \label{eq:hellinger}
    H^2(a, b)=\frac{1}{2} \sum_{i=1}^{k}\left(\sqrt{a_{i}/\|a\|_1}-\sqrt{b_{i}/\|b\|_1}\right)^{2}
\end{equation}
between both true and predicted class discrete distributions $a$ and $b$.
Figure~\ref{fig:exp_jkonet_cell}c and Table~\ref{tab:exp_jkonet_cell_class} demonstrate that both, \textsc{JKOnet} and the forward method, capture most lineage branches during the differentiation of embryonic stem cells. Both methods, however, have difficulties recovering cells of the neural crest (class 1) and the endoderm (class 3), lineage branches which are scarcely represented in the original data. 
The analysis further suggests that both methods reduce in performance w.r.t. biological heterogeneity when predicting the entire trajectory (all-steps ahead), instead of inferring the next snapshot only (one-step ahead).

\vspace{-5pt}
\section{Conclusion}
\vspace{-5pt}
We proposed \textsc{JKOnet}, a model to infer and predict the evolution of population dynamics using a proximal optimal transport scheme, the JKO flow.
\textsc{JKOnet} solves local JKO steps using ICNNs and learns the energy that parameterizes these steps by fitting JKO flow predictions to observed trajectories using a fully differentiable bilevel optimization problem.
We validate its effectiveness through experiments on synthetic potential- and trajectory-based population dynamics, and observe that it is far more robust to noise than a more direct Forward approach. We use \textsc{JKOnet} to infer the developmental trajectories of human embryonic stem cells captured via high-dimensional and time-resolved single-cell RNAseq. 
Our analysis also shows that \textsc{JKOnet} captures diverse cell fates during the incremental differentiation of embryonic cells into multiple lineage branches.
Using proximal optimal transport to model real complex population dynamics thus makes for an exciting avenue of future work. Extensions could include modeling higher-order interactions among population particles in the energy function, e.g., cell-cell communication.

\vspace{-5pt}
\acknowledgments{
\vspace{-5pt}
This project received funding from the Swiss National Science Foundation under the National Center of Competence in Research (NCCR) Catalysis under grant agreement 51NF40 180544, and was supported by Google Cloud for Higher Education.
}

\bibliographystyle{abbrvnat}
\bibliography{main}

\begin{thebibliography}{68}
\providecommand{\natexlab}[1]{#1}
\providecommand{\url}[1]{\texttt{#1}}
\expandafter\ifx\csname urlstyle\endcsname\relax
  \providecommand{\doi}[1]{doi: #1}\else
  \providecommand{\doi}{doi: \begingroup \urlstyle{rm}\Url}\fi

\bibitem[Alvarez-Melis et~al.(2021)Alvarez-Melis, Schiff, and
  Mroueh]{alvarez2021optimizing}
D.~Alvarez-Melis, Y.~Schiff, and Y.~Mroueh.
\newblock {Optimizing Functionals on the Space of Probabilities with Input
  Convex Neural Networks}.
\newblock \emph{arXiv preprint arXiv:2106.00774}, 2021.

\bibitem[Ambrosio et~al.(2006)Ambrosio, Gigli, and
  Savar{\'e}]{ambrosio2006gradient}
L.~Ambrosio, N.~Gigli, and G.~Savar{\'e}.
\newblock \emph{{Gradient Flows in Metric Spaces and in the Space of
  Probability Measures}}.
\newblock Springer, 2006.

\bibitem[Amos et~al.(2017)Amos, Xu, and Kolter]{amos2017input}
B.~Amos, L.~Xu, and J.~Z. Kolter.
\newblock {Input Convex Neural Networks}.
\newblock In \emph{International Conference on Machine Learning (ICML)},
  volume~34, 2017.

\bibitem[Benamou et~al.(2016{\natexlab{a}})Benamou, Carlier, and
  Laborde]{benamou2016augmented}
J.-D. Benamou, G.~Carlier, and M.~Laborde.
\newblock An augmented lagrangian approach to wasserstein gradient flows and
  applications.
\newblock \emph{ESAIM: Proceedings and surveys}, 54:\penalty0 1--17,
  2016{\natexlab{a}}.

\bibitem[Benamou et~al.(2016{\natexlab{b}})Benamou, Carlier, M{\'e}rigot, and
  Oudet]{benamou2016}
J.-D. Benamou, G.~Carlier, Q.~M{\'e}rigot, and E.~Oudet.
\newblock Discretization of functionals involving the {Monge--Amp{\'e}re}
  operator.
\newblock \emph{Numerische Mathematik}, 134\penalty0 (3), 2016{\natexlab{b}}.

\bibitem[Bradbury et~al.(2018)Bradbury, Frostig, Hawkins, Johnson, Leary,
  Maclaurin, Necula, Paszke, Vander{P}las, Wanderman-{M}ilne, and
  Zhang]{jax2018github}
J.~Bradbury, R.~Frostig, P.~Hawkins, M.~J. Johnson, C.~Leary, D.~Maclaurin,
  G.~Necula, A.~Paszke, J.~Vander{P}las, S.~Wanderman-{M}ilne, and Q.~Zhang.
\newblock {JAX}: composable transformations of {P}ython+{N}um{P}y programs,
  2018.
\newblock URL \url{http://github.com/google/jax}.

\bibitem[Brenier(1987)]{Brenier1987}
Y.~Brenier.
\newblock D{\'e}composition polaire et r{\'e}arrangement monotone des champs de
  vecteurs.
\newblock \emph{CR Acad. Sci. Paris S{\'e}r. I Math.}, 305, 1987.

\bibitem[Burger et~al.(2010)Burger, Carrillo, and Wolfram]{burger2010}
M.~Burger, J.~A. Carrillo, and M.-T. Wolfram.
\newblock {A mixed finite element method for nonlinear diffusion equations}.
\newblock \emph{Kinetic \& Related Models}, 3\penalty0 (1), 2010.

\bibitem[Caffarelli(2000)]{caffarelli2000monotonicity}
L.~A. Caffarelli.
\newblock {Monotonicity Properties of Optimal Transportation and the FKG and
  Related Inequalities}.
\newblock \emph{Communications in Mathematical Physics}, 214\penalty0 (3),
  2000.

\bibitem[Carrillo et~al.(2021)Carrillo, Craig, Wang, and
  Wei]{carrillo2021primal}
J.~A. Carrillo, K.~Craig, L.~Wang, and C.~Wei.
\newblock {Primal Dual Methods for Wasserstein Gradient Flows}.
\newblock \emph{Foundations of Computational Mathematics}, 2021.

\bibitem[Chen et~al.(2019)Chen, Shi, and Zhang]{chen2018optimal}
Y.~Chen, Y.~Shi, and B.~Zhang.
\newblock {Optimal Control Via Neural Networks: A Convex Approach}.
\newblock In \emph{International Conference on Learning Representations
  (ICLR)}, 2019.

\bibitem[Combettes and Pesquet(2011)]{combettes2011proximal}
P.~L. Combettes and J.-C. Pesquet.
\newblock Proximal splitting methods in signal processing.
\newblock In \emph{Fixed-point algorithms for inverse problems in science and
  engineering}, pages 185--212. Springer, 2011.

\bibitem[Cuturi(2013)]{cuturi2013sinkhorn}
M.~Cuturi.
\newblock {Sinkhorn Distances: Lightspeed Computation of Optimal Transport}.
\newblock In \emph{Advances in Neural Information Processing Systems
  (NeurIPS)}, volume~26, 2013.

\bibitem[Cuturi et~al.(2022)Cuturi, Meng-Papaxanthos, Tian, Bunne, Davis, and
  Teboul]{cuturi2022optimal}
M.~Cuturi, L.~Meng-Papaxanthos, Y.~Tian, C.~Bunne, G.~Davis, and O.~Teboul.
\newblock {Optimal Transport Tools (OTT): A JAX Toolbox for all things
  Wasserstein}.
\newblock \emph{arXiv preprint arXiv:2201.12324}, 2022.

\bibitem[Danskin(1967)]{danskin2012theory}
J.~M. Danskin.
\newblock \emph{{The Theory of Max-Min and its Applications to Weapons
  Allocation Problems}}, volume~5.
\newblock Springer, 1967.

\bibitem[De~Bie et~al.(2019)De~Bie, Peyr{\'e}, and Cuturi]{de2019stochastic}
G.~De~Bie, G.~Peyr{\'e}, and M.~Cuturi.
\newblock {Stochastic Deep Networks}.
\newblock In \emph{International Conference on Machine Learning (ICML)},
  volume~36, 2019.

\bibitem[Edwards and Storkey(2017)]{edwards2016}
H.~Edwards and A.~Storkey.
\newblock {Towards a Neural Statistician}.
\newblock In \emph{International Conference on Learning Representations
  (ICLR)}, volume~5, 2017.

\bibitem[Feydy et~al.(2019)Feydy, S{\'e}journ{\'e}, Vialard, Amari, Trouv{\'e},
  and Peyr{\'e}]{feydy2018interpolating}
J.~Feydy, T.~S{\'e}journ{\'e}, F.-X. Vialard, S.-I. Amari, A.~Trouv{\'e}, and
  G.~Peyr{\'e}.
\newblock {Interpolating between Optimal Transport and MMD using Sinkhorn
  Divergences}.
\newblock In \emph{International Conference on Artificial Intelligence and
  Statistics (AISTATS)}, volume~22, 2019.

\bibitem[Figalli(2010)]{figalli2010optimal}
A.~Figalli.
\newblock {The Optimal Partial Transport Problem}.
\newblock \emph{Archive for Rational Mechanics and Analysis}, 195\penalty0 (2),
  2010.

\bibitem[Fisher et~al.(2009)Fisher, Nocedal, Tr{\'e}molet, and
  Wright]{fisher2009data}
M.~Fisher, J.~Nocedal, Y.~Tr{\'e}molet, and S.~J. Wright.
\newblock Data assimilation in weather forecasting: a case study in
  pde-constrained optimization.
\newblock \emph{Optimization and Engineering}, 10\penalty0 (3):\penalty0
  409--426, 2009.

\bibitem[Flamary et~al.(2021)Flamary, Courty, Gramfort, Alaya, Boisbunon,
  Chambon, Chapel, Corenflos, Fatras, Fournier, Gautheron, Gayraud, Janati,
  Rakotomamonjy, Redko, Rolet, Schutz, Seguy, Sutherland, Tavenard, Tong, and
  Vayer]{flamary2021pot}
R.~Flamary, N.~Courty, A.~Gramfort, M.~Z. Alaya, A.~Boisbunon, S.~Chambon,
  L.~Chapel, A.~Corenflos, K.~Fatras, N.~Fournier, L.~Gautheron, N.~T. Gayraud,
  H.~Janati, A.~Rakotomamonjy, I.~Redko, A.~Rolet, A.~Schutz, V.~Seguy, D.~J.
  Sutherland, R.~Tavenard, A.~Tong, and T.~Vayer.
\newblock Pot: Python optimal transport.
\newblock \emph{Journal of Machine Learning Research}, 22, 2021.

\bibitem[Genevay et~al.(2019)Genevay, Chizat, Bach, Cuturi, and
  Peyr{\'e}]{genevay2018}
A.~Genevay, L.~Chizat, F.~Bach, M.~Cuturi, and G.~Peyr{\'e}.
\newblock {Sample Complexity of Sinkhorn Divergences}.
\newblock In \emph{International Conference on Artificial Intelligence and
  Statistics (AISTATS)}, volume~22, 2019.

\bibitem[Grathwohl et~al.(2019)Grathwohl, Chen, Bettencourt, Sutskever, and
  Duvenaud]{grathwohl2018ffjord}
W.~Grathwohl, R.~T. Chen, J.~Bettencourt, I.~Sutskever, and D.~Duvenaud.
\newblock {FFJORD: Free-Form Continuous Dynamics for Scalable Reversible
  Generative Models}.
\newblock In \emph{International Conference on Learning Representations
  (ICLR)}, 2019.

\bibitem[Haasler et~al.(2019)Haasler, Ringh, Chen, and
  Karlsson]{haasler2019estimating}
I.~Haasler, A.~Ringh, Y.~Chen, and J.~Karlsson.
\newblock {Estimating ensemble flows on a hidden Markov chain}.
\newblock In \emph{2019 IEEE 58th Conference on Decision and Control (CDC)}.
  IEEE, 2019.

\bibitem[Haasler et~al.(2021{\natexlab{a}})Haasler, Ringh, Chen, and
  Karlsson]{haasler2021multimarginal}
I.~Haasler, A.~Ringh, Y.~Chen, and J.~Karlsson.
\newblock {Multimarginal Optimal Transport with a Tree-Structured Cost and the
  Schr\"odinger Bridge Problem}.
\newblock \emph{SIAM Journal on Control and Optimization}, 59\penalty0 (4),
  2021{\natexlab{a}}.

\bibitem[Haasler et~al.(2021{\natexlab{b}})Haasler, Singh, Zhang, Karlsson, and
  Chen]{haasler2021multi}
I.~Haasler, R.~Singh, Q.~Zhang, J.~Karlsson, and Y.~Chen.
\newblock Multi-marginal optimal transport and probabilistic graphical models.
\newblock \emph{IEEE Transactions on Information Theory}, 2021{\natexlab{b}}.

\bibitem[Hashimoto et~al.(2016)Hashimoto, Gifford, and
  Jaakkola]{hashimoto2016learning}
T.~Hashimoto, D.~Gifford, and T.~Jaakkola.
\newblock {Learning Population-Level Diffusions with Generative Recurrent
  Networks}.
\newblock In \emph{International Conference on Machine Learning (ICML)},
  volume~33, 2016.

\bibitem[He et~al.(2016)He, Zhang, Ren, and Sun]{he2016deep}
K.~He, X.~Zhang, S.~Ren, and J.~Sun.
\newblock {Deep Residual Learning for Image Recognition}.
\newblock In \emph{IEEE Conference on Computer Vision and Pattern Recognition
  (CVPR)}, 2016.

\bibitem[Huang et~al.(2021)Huang, Chen, Tsirigotis, and
  Courville]{huang2021convex}
C.-W. Huang, R.~T.~Q. Chen, C.~Tsirigotis, and A.~Courville.
\newblock {Convex Potential Flows: Universal Probability Distributions with
  Optimal Transport and Convex Optimization}.
\newblock In \emph{International Conference on Learning Representations
  (ICLR)}, 2021.

\bibitem[Jordan et~al.(1998)Jordan, Kinderlehrer, and
  Otto]{jordan1998variational}
R.~Jordan, D.~Kinderlehrer, and F.~Otto.
\newblock {The Variational Formulation of the Fokker--Planck Equation}.
\newblock \emph{SIAM Journal on Mathematical Analysis}, 29\penalty0 (1), 1998.

\bibitem[Kingma and Ba(2014)]{kingma2014adam}
D.~P. Kingma and J.~Ba.
\newblock {Adam: A Method for Stochastic Optimization}.
\newblock In \emph{International Conference on Learning Representations
  (ICLR)}, 2014.

\bibitem[Kloeden and Platen(1992)]{kloeden1992stochastic}
P.~E. Kloeden and E.~Platen.
\newblock {Stochastic Differential Equations}.
\newblock In \emph{{Numerical Solution of Stochastic Differential Equations}}.
  Springer, 1992.

\bibitem[Krishnan et~al.(2017)Krishnan, Shalit, and
  Sontag]{krishnan2017structured}
R.~Krishnan, U.~Shalit, and D.~Sontag.
\newblock {Structured Inference Networks for Nonlinear State Space Models}.
\newblock In \emph{AAAI Conference on Artificial Intelligence}, volume~31,
  2017.

\bibitem[LeCun et~al.(2012)LeCun, Bottou, Orr, and
  M{\"u}ller]{lecun2012efficient}
Y.~A. LeCun, L.~Bottou, G.~B. Orr, and K.-R. M{\"u}ller.
\newblock {Efficient Backprop}.
\newblock In \emph{Neural Networks: Tricks of the Trade}. Springer, 2012.

\bibitem[Lee et~al.(2019)Lee, Lee, Kim, Kosiorek, Choi, and Teh]{lee2019}
J.~Lee, Y.~Lee, J.~Kim, A.~Kosiorek, S.~Choi, and Y.~W. Teh.
\newblock {Set Transformer: A Framework for Attention-based
  Permutation-Invariant Neural Networks}.
\newblock In \emph{International Conference on Machine Learning (ICML)}, 2019.

\bibitem[Li et~al.(2020)Li, Wong, Chen, and Duvenaud]{li2020scalable}
X.~Li, T.-K.~L. Wong, R.~T. Chen, and D.~K. Duvenaud.
\newblock {Scalable Gradients and Variational Inference for Stochastic
  Differential Equations}.
\newblock In \emph{Symposium on Advances in Approximate Bayesian Inference}.
  PMLR, 2020.

\bibitem[Lorraine et~al.(2020)Lorraine, Vicol, and Duvenaud]{lorraine2020}
J.~Lorraine, P.~Vicol, and D.~Duvenaud.
\newblock {Optimizing Millions of Hyperparameters by Implicit Differentiation}.
\newblock In \emph{International Conference on Artificial Intelligence and
  Statistics (AISTATS)}, 2020.

\bibitem[Luecken and Theis(2019)]{luecken2019current}
M.~D. Luecken and F.~J. Theis.
\newblock {Current best practices in single-cell RNA-seqanalysis: a tutorial}.
\newblock \emph{Molecular Systems Biology}, 15\penalty0 (6), 2019.

\bibitem[Luo et~al.(2020)Luo, Xing, Milan, Zhang, Liu, and
  Kim]{luo2020multiple}
W.~Luo, J.~Xing, A.~Milan, X.~Zhang, W.~Liu, and T.-K. Kim.
\newblock Multiple object tracking: A literature review.
\newblock \emph{Artificial Intelligence}, page 103448, 2020.

\bibitem[Makkuva et~al.(2020)Makkuva, Taghvaei, Oh, and
  Lee]{pmlr-v119-makkuva20a}
A.~Makkuva, A.~Taghvaei, S.~Oh, and J.~Lee.
\newblock Optimal transport mapping via input convex neural networks.
\newblock In \emph{International Conference on Machine Learning (ICML)},
  volume~37, 2020.

\bibitem[Martin and Evans(1975)]{martin1975}
G.~R. Martin and M.~J. Evans.
\newblock {Differentiation of Clonal Lines of Teratocarcinoma Cells: Formation
  of Embryoid Bodies In Vitro}.
\newblock \emph{Proceedings of the National Academy of Sciences}, 72\penalty0
  (4), 1975.

\bibitem[Metz et~al.(2017)Metz, Poole, Pfau, and
  Sohl-Dickstein]{metz2016unrolled}
L.~Metz, B.~Poole, D.~Pfau, and J.~Sohl-Dickstein.
\newblock {Unrolled Generative Adversarial Networks}.
\newblock In \emph{International Conference on Learning Representations
  (ICLR)}, 2017.

\bibitem[Mokrov et~al.(2021)Mokrov, Korotin, Li, Genevay, Solomon, and
  Burnaev]{mokrov2021large}
P.~Mokrov, A.~Korotin, L.~Li, A.~Genevay, J.~Solomon, and E.~Burnaev.
\newblock {Large-Scale Wasserstein Gradient Flows}.
\newblock \emph{Advances in Neural Information Processing Systems (NeurIPS)},
  2021.

\bibitem[Monge(1781)]{Monge1781}
G.~Monge.
\newblock M{\'e}moire sur la th{\'e}orie des d{\'e}blais et des remblais.
\newblock \emph{Histoire de l'Acad{\'e}mie Royale des Sciences}, pages
  666--704, 1781.

\bibitem[Moon et~al.(2019)Moon, van Dijk, Wang, Gigante, Burkhardt, Chen, Yim,
  van~den Elzen, Hirn, Coifman, et~al.]{moon2019}
K.~R. Moon, D.~van Dijk, Z.~Wang, S.~Gigante, D.~B. Burkhardt, W.~S. Chen,
  K.~Yim, A.~van~den Elzen, M.~J. Hirn, R.~R. Coifman, et~al.
\newblock {Visualizing structure and transitions in high-dimensional biological
  data}.
\newblock \emph{Nature Biotechnology}, 37\penalty0 (12), 2019.

\bibitem[Pascanu et~al.(2013)Pascanu, Mikolov, and
  Bengio]{pascanu2013difficulty}
R.~Pascanu, T.~Mikolov, and Y.~Bengio.
\newblock {On the difficulty of training Recurrent Neural Networks}.
\newblock In \emph{International Conference on Machine Learning (ICML)},
  volume~28, 2013.

\bibitem[Paty et~al.(2020)Paty, d’Aspremont, and Cuturi]{paty2020regularity}
F.-P. Paty, A.~d’Aspremont, and M.~Cuturi.
\newblock {Regularity as Regularization: Smooth and Strongly Convex Brenier
  Potentials in Optimal Transport}.
\newblock In \emph{International Conference on Artificial Intelligence and
  Statistics (AISTATS)}, 2020.

\bibitem[Peyr{\'e}(2015)]{2015-Peyre-siims}
G.~Peyr{\'e}.
\newblock {Entropic Approximation of Wasserstein Gradient Flows}.
\newblock \emph{SIAM Journal on Imaging Sciences}, 8\penalty0 (4), 2015.

\bibitem[Peyré and Cuturi(2019)]{Peyre2019computational}
G.~Peyré and M.~Cuturi.
\newblock Computational optimal transport.
\newblock \emph{Foundations and Trends in Machine Learning}, 11\penalty0 (5-6),
  2019.
\newblock ISSN 1935-8245.

\bibitem[Ramdas et~al.(2017)Ramdas, Trillos, and Cuturi]{ramdas2017wasserstein}
A.~Ramdas, N.~G. Trillos, and M.~Cuturi.
\newblock {On Wasserstein Two Sample Testing and Related Families of
  Nonparametric Tests}.
\newblock \emph{Entropy}, 19\penalty0 (2):\penalty0 47, 2017.

\bibitem[Rezende and Mohamed(2015)]{rezende2015variational}
D.~Rezende and S.~Mohamed.
\newblock {Variational Inference with Normalizing Flows}.
\newblock In \emph{International Conference on Machine Learning (ICML)}, 2015.

\bibitem[Salimans et~al.(2018)Salimans, Zhang, Radford, and
  Metaxas]{salimans2018improving}
T.~Salimans, H.~Zhang, A.~Radford, and D.~Metaxas.
\newblock {Improving GANs Using Optimal Transport}.
\newblock In \emph{International Conference on Learning Representations
  (ICLR)}, 2018.

\bibitem[Santambrogio(2015)]{santambrogio2015optimal}
F.~Santambrogio.
\newblock {Optimal Transport for Applied Mathematicians}.
\newblock \emph{Birk{\"a}user, NY}, 55\penalty0 (58-63):\penalty0 94, 2015.

\bibitem[Santambrogio(2017)]{santambrogio2017euclidean}
F.~Santambrogio.
\newblock $\{$Euclidean, metric, and {Wasserstein}$\}$ gradient flows: an
  overview.
\newblock \emph{Bulletin of Mathematical Sciences}, 7\penalty0 (1), 2017.

\bibitem[Scaman and Virmaux(2018)]{scaman2018lipschitz}
K.~Scaman and A.~Virmaux.
\newblock Lipschitz regularity of deep neural networks: analysis and efficient
  estimation.
\newblock In \emph{Advances in Neural Information Processing Systems
  (NeurIPS)}, 2018.

\bibitem[Schiebinger et~al.(2019)Schiebinger, Shu, Tabaka, Cleary, Subramanian,
  Solomon, Gould, Liu, Lin, Berube, et~al.]{schiebinger2019}
G.~Schiebinger, J.~Shu, M.~Tabaka, B.~Cleary, V.~Subramanian, A.~Solomon,
  J.~Gould, S.~Liu, S.~Lin, P.~Berube, et~al.
\newblock {Optimal-Transport Analysis of Single-Cell Gene Expression Identifies
  Developmental Trajectories in Reprogramming}.
\newblock \emph{Cell}, 176\penalty0 (4), 2019.

\bibitem[Shamblott et~al.(2009)Shamblott, Kerr, Axelman, Littlefield, Clark,
  Patterson, Addis, Kraszewski, Kent, and Gearhart]{shamblott2009derivation}
M.~J. Shamblott, C.~L. Kerr, J.~Axelman, J.~W. Littlefield, G.~O. Clark, E.~S.
  Patterson, R.~C. Addis, J.~N. Kraszewski, K.~C. Kent, and J.~D. Gearhart.
\newblock {Derivation and Differentiation of Human Embryonic Germ Cells}.
\newblock In \emph{Essentials of Stem Cell Biology}. Elsevier, 2009.

\bibitem[Sheldon et~al.(2007)Sheldon, Elmohamed, and
  Kozen]{sheldon2007collective}
D.~Sheldon, M.~Elmohamed, and D.~Kozen.
\newblock {Collective Inference on Markov Models for Modeling Bird Migration}.
\newblock In \emph{Advances in Neural Information Processing Systems
  (NeurIPS)}, volume~20, 2007.

\bibitem[Sheldon and Dietterich(2011)]{sheldon2011collective}
D.~R. Sheldon and T.~G. Dietterich.
\newblock {Collective Graphical Models}.
\newblock In \emph{Advances in Neural Information Processing Systems
  (NeurIPS)}, 2011.

\bibitem[Sigrist et~al.(2015)Sigrist, K{\"u}nsch, and
  Stahel]{sigrist2015stochastic}
F.~Sigrist, H.~R. K{\"u}nsch, and W.~A. Stahel.
\newblock Stochastic partial differential equation based modelling of large
  space--time data sets.
\newblock \emph{Journal of the Royal Statistical Society: Series B: Statistical
  Methodology}, pages 3--33, 2015.

\bibitem[Tong et~al.(2020)Tong, Huang, Wolf, Van~Dijk, and
  Krishnaswamy]{tong2020trajectorynet}
A.~Tong, J.~Huang, G.~Wolf, D.~Van~Dijk, and S.~Krishnaswamy.
\newblock Trajectorynet: A dynamic optimal transport network for modeling
  cellular dynamics.
\newblock In \emph{International Conference on Machine Learning (ICML)}, 2020.

\bibitem[Vaswani et~al.(2017)Vaswani, Shazeer, Parmar, Uszkoreit, Jones, Gomez,
  Kaiser, and Polosukhin]{vaswani2017attention}
A.~Vaswani, N.~Shazeer, N.~Parmar, J.~Uszkoreit, L.~Jones, A.~N. Gomez,
  {\L}.~Kaiser, and I.~Polosukhin.
\newblock {Attention is All you Need}.
\newblock In \emph{Advances in Neural Information Processing Systems
  (NeurIPS)}, 2017.

\bibitem[Williams and Zipser(1989)]{williams1989learning}
R.~J. Williams and D.~Zipser.
\newblock {A Learning Algorithm for Continually Running Fully Recurrent Neural
  Networks}.
\newblock \emph{Neural Computation}, 1\penalty0 (2), 1989.

\bibitem[Wolf et~al.(2018)Wolf, Angerer, and Theis]{wolf2018scanpy}
F.~A. Wolf, P.~Angerer, and F.~J. Theis.
\newblock {SCANPY: large-scale single-cell gene expression data analysis}.
\newblock \emph{Genome biology}, 19\penalty0 (1), 2018.

\bibitem[Yang and Uhler(2019)]{yang2018scalable}
K.~D. Yang and C.~Uhler.
\newblock {Scalable Unbalanced Optimal Transport using Generative Adversarial
  Networks}.
\newblock \emph{International Conference on Learning Representations (ICLR)},
  2019.

\bibitem[Yang et~al.(2020)Yang, Damodaran, Venkatachalapathy, Soylemezoglu,
  Shivashankar, and Uhler]{yang2020predicting}
K.~D. Yang, K.~Damodaran, S.~Venkatachalapathy, A.~C. Soylemezoglu,
  G.~Shivashankar, and C.~Uhler.
\newblock Predicting cell lineages using autoencoders and optimal transport.
\newblock \emph{PLoS Computational Biology}, 16\penalty0 (4), 2020.

\bibitem[Zaheer et~al.(2017)Zaheer, Kottur, Ravanbakhsh, Poczos, Salakhutdinov,
  and Smola]{zaheer2017}
M.~Zaheer, S.~Kottur, S.~Ravanbakhsh, B.~Poczos, R.~R. Salakhutdinov, and A.~J.
  Smola.
\newblock {Deep Sets}.
\newblock In \emph{Advances in Neural Information Processing Systems
  (NeurIPS)}, volume~30, 2017.

\bibitem[Zheng et~al.(2017)Zheng, Terry, Belgrader, Ryvkin, Bent, Wilson,
  Ziraldo, Wheeler, McDermott, Zhu, et~al.]{zheng2017massively}
G.~X. Zheng, J.~M. Terry, P.~Belgrader, P.~Ryvkin, Z.~W. Bent, R.~Wilson, S.~B.
  Ziraldo, T.~D. Wheeler, G.~P. McDermott, J.~Zhu, et~al.
\newblock Massively parallel digital transcriptional profiling of single cells.
\newblock \emph{Nature communications}, 8\penalty0 (1), 2017.

\end{thebibliography}

\clearpage
\newpage
\onecolumn

\section*{Appendix}
\appendix

\section{Additional Evaluation}

\subsection{Synthetic Population Dynamics}
\begin{figure*}[h]
\subfloat[\centering Quadratic Potential.]{\includegraphics[width=.25\linewidth]{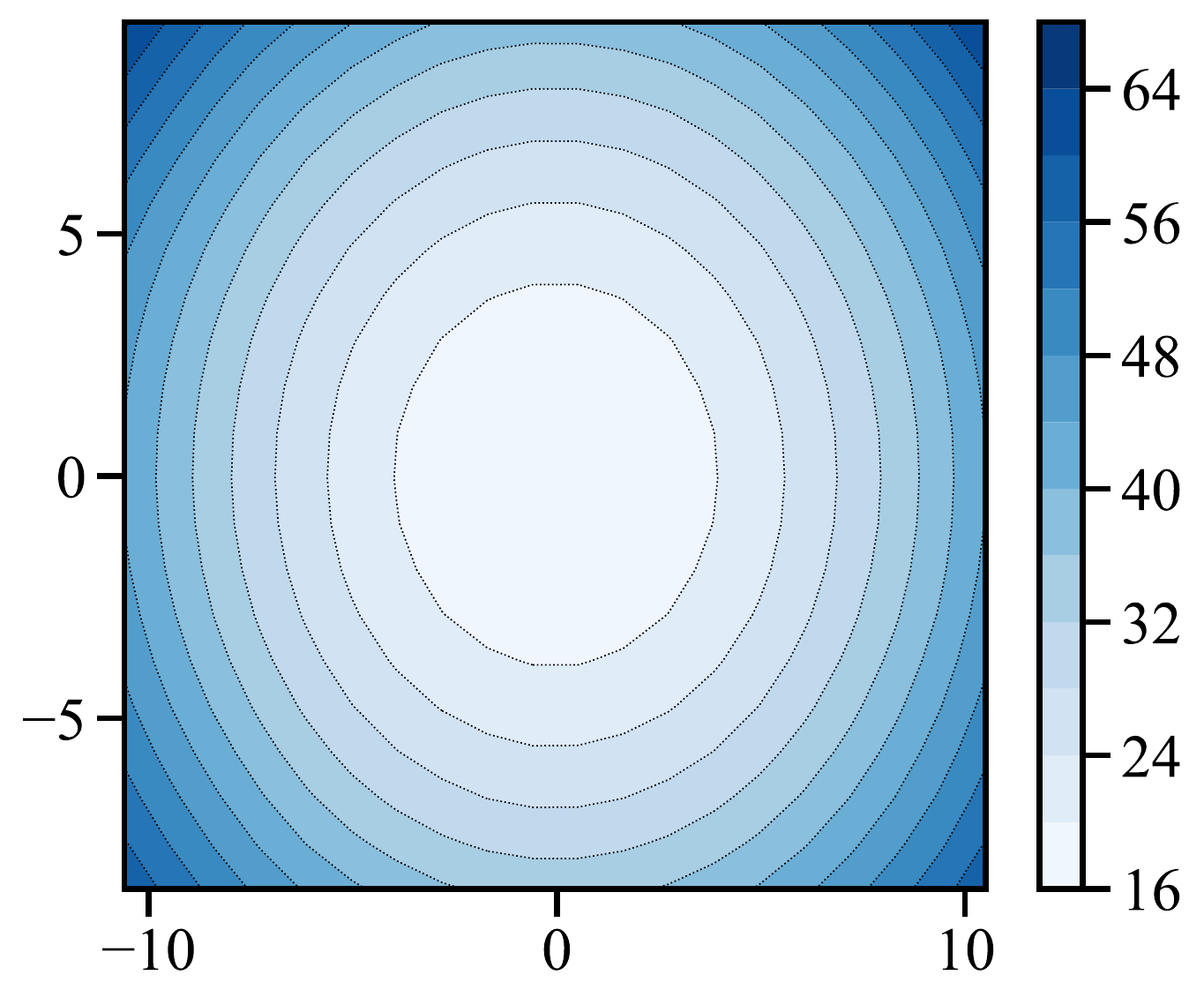}}
\subfloat[\centering Styblinski Potential.]{\includegraphics[width=.25\linewidth]{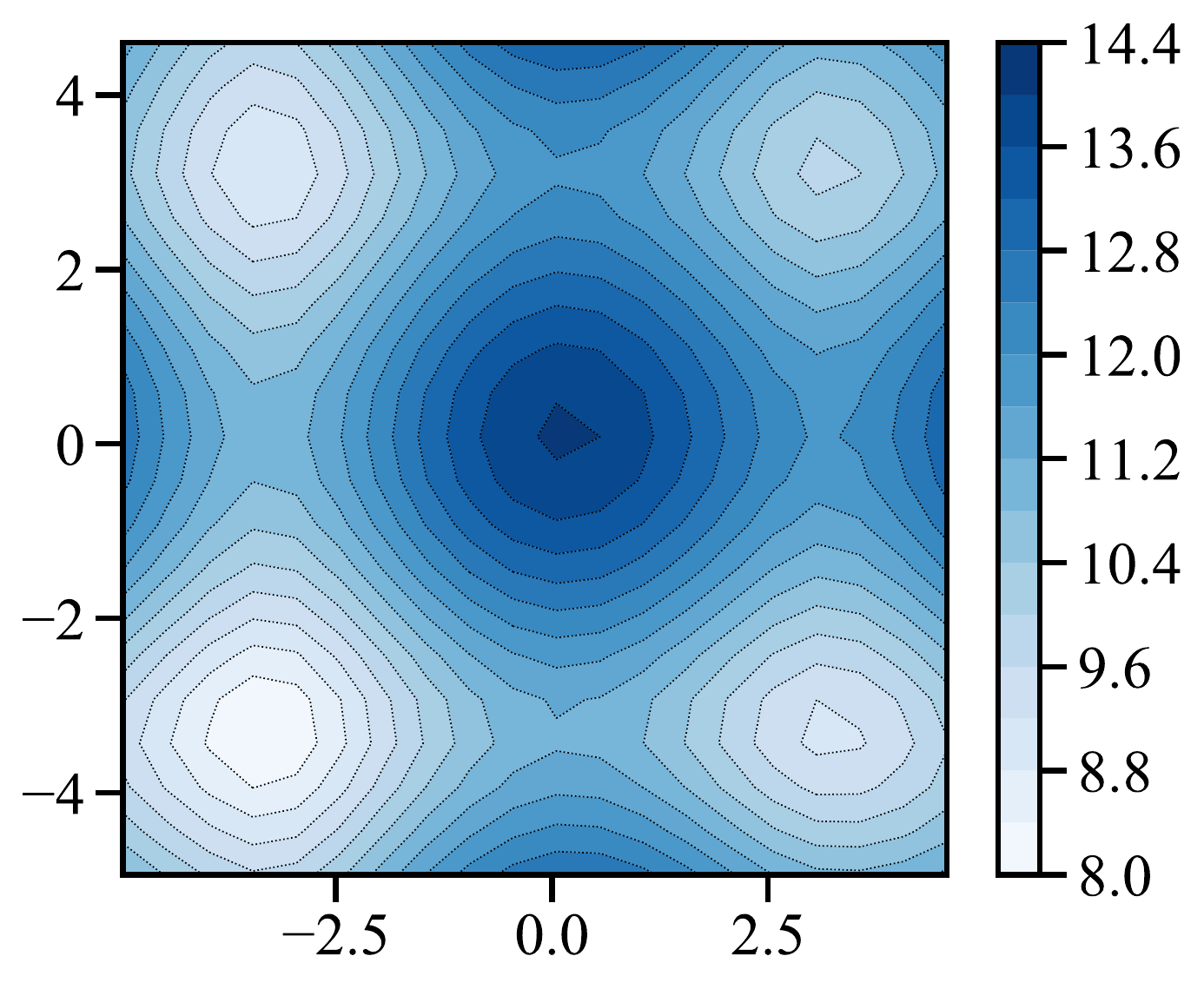}}
\subfloat[\centering Semicircle Trajectory \protect\linebreak \emph{with} teacher forcing.]{\includegraphics[width=.25\linewidth]{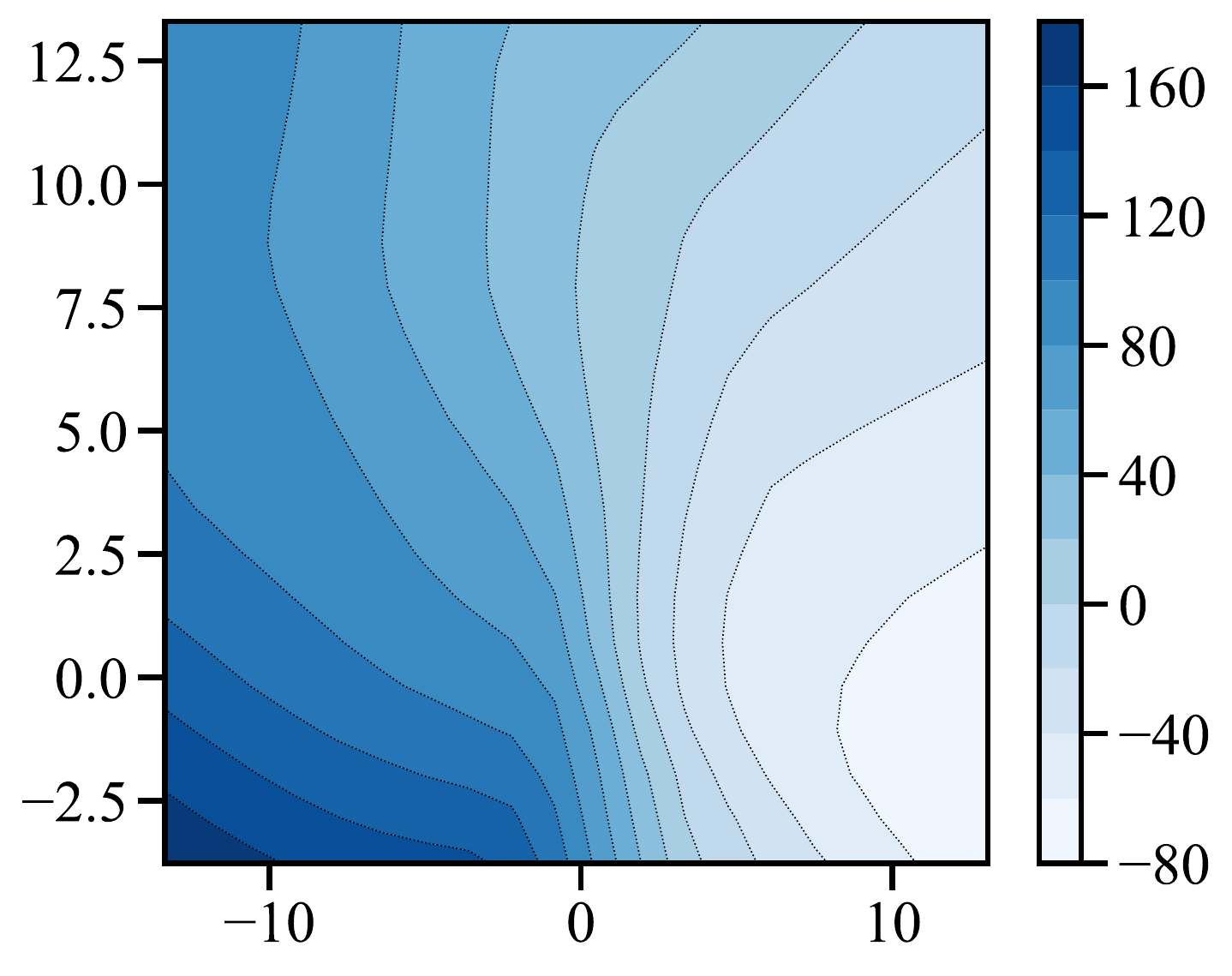}}
\subfloat[\centering Semicircle Trajectory.]{\includegraphics[width=.25\linewidth]{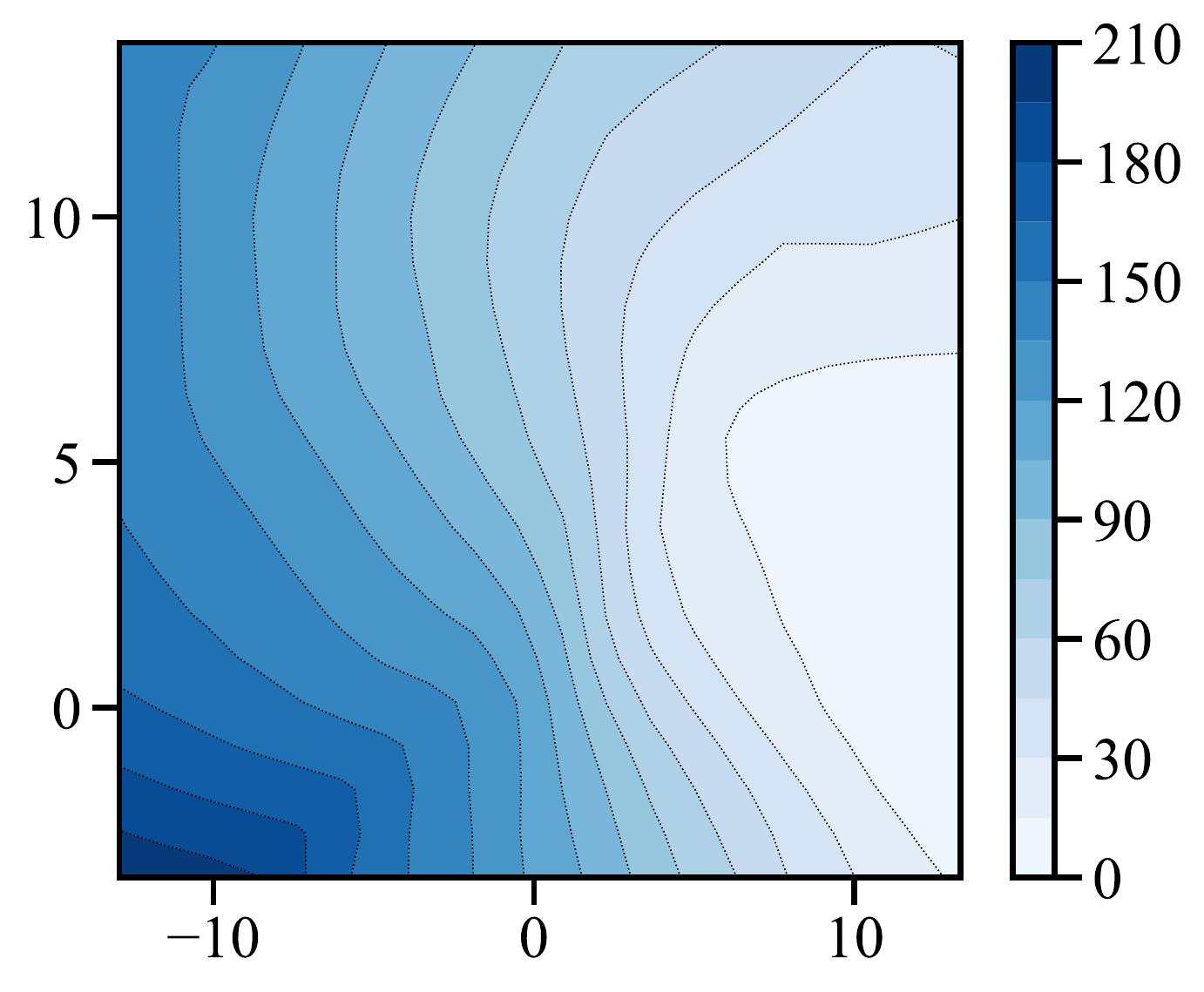}}
\caption{\textbf{Results of the Forward Method on Potential- and Trajectory-based Dyanamics.} (a)-(d) Contour plots of the energy functionals $F_\xi$ of the forward method on potential- and trajectory-based population dynamics in different training settings (i.e., trained with or without teacher forcing \S~ \ref{sec:learn_energy}), color gradients depict the magnitude of $F_\xi$.}
\label{fig:exp_forward_pot_traj}
\end{figure*}

\begin{figure*}[h]
\centering
\subfloat[\centering PCA embedding of predictions of the forward method colored by the snapshot time.]{\includegraphics[width=.45\linewidth]{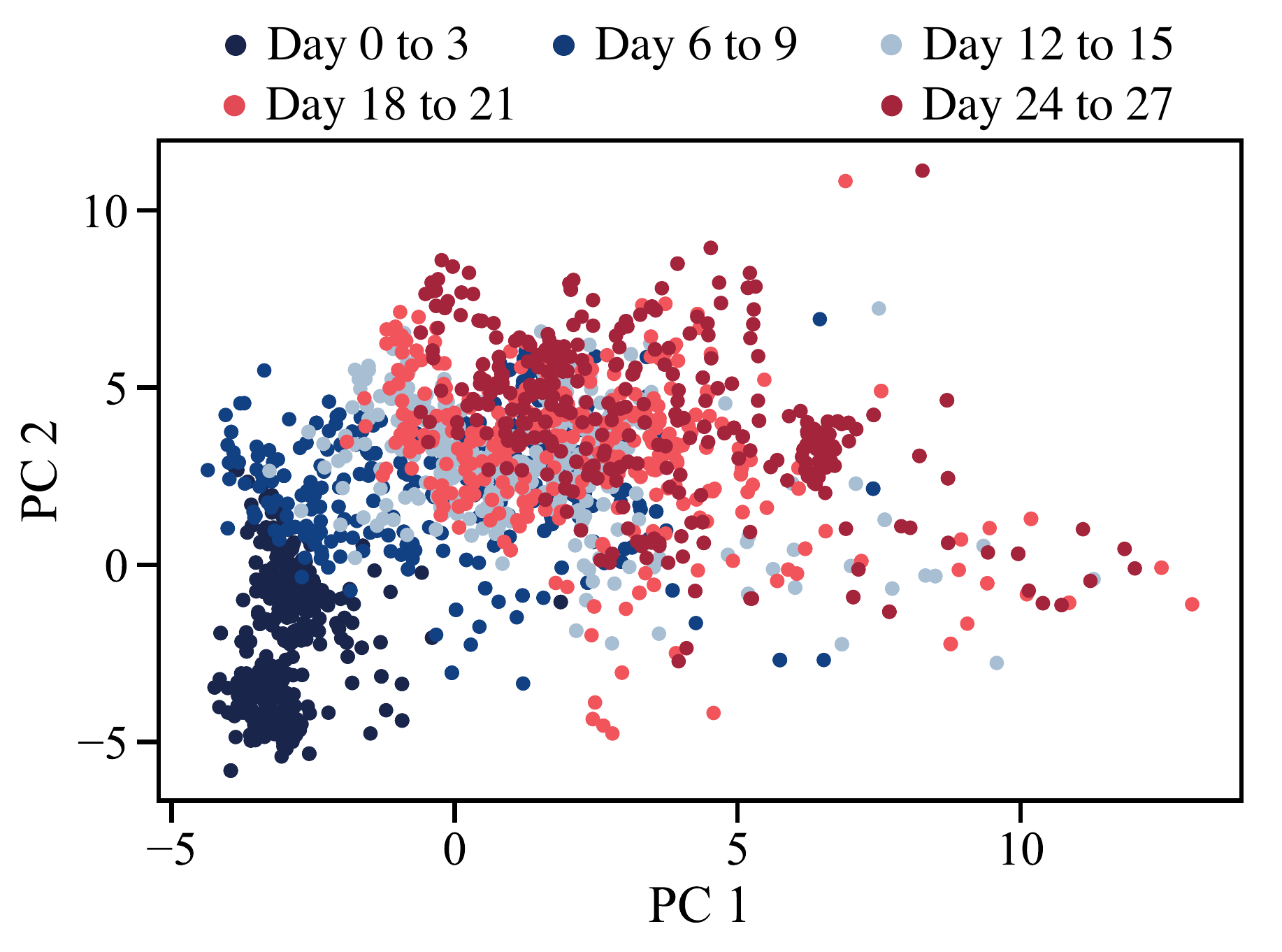}}
\subfloat[\centering PCA embedding of predictions of the forward method colored by the lineage branch class.]{\includegraphics[width=.45\linewidth]{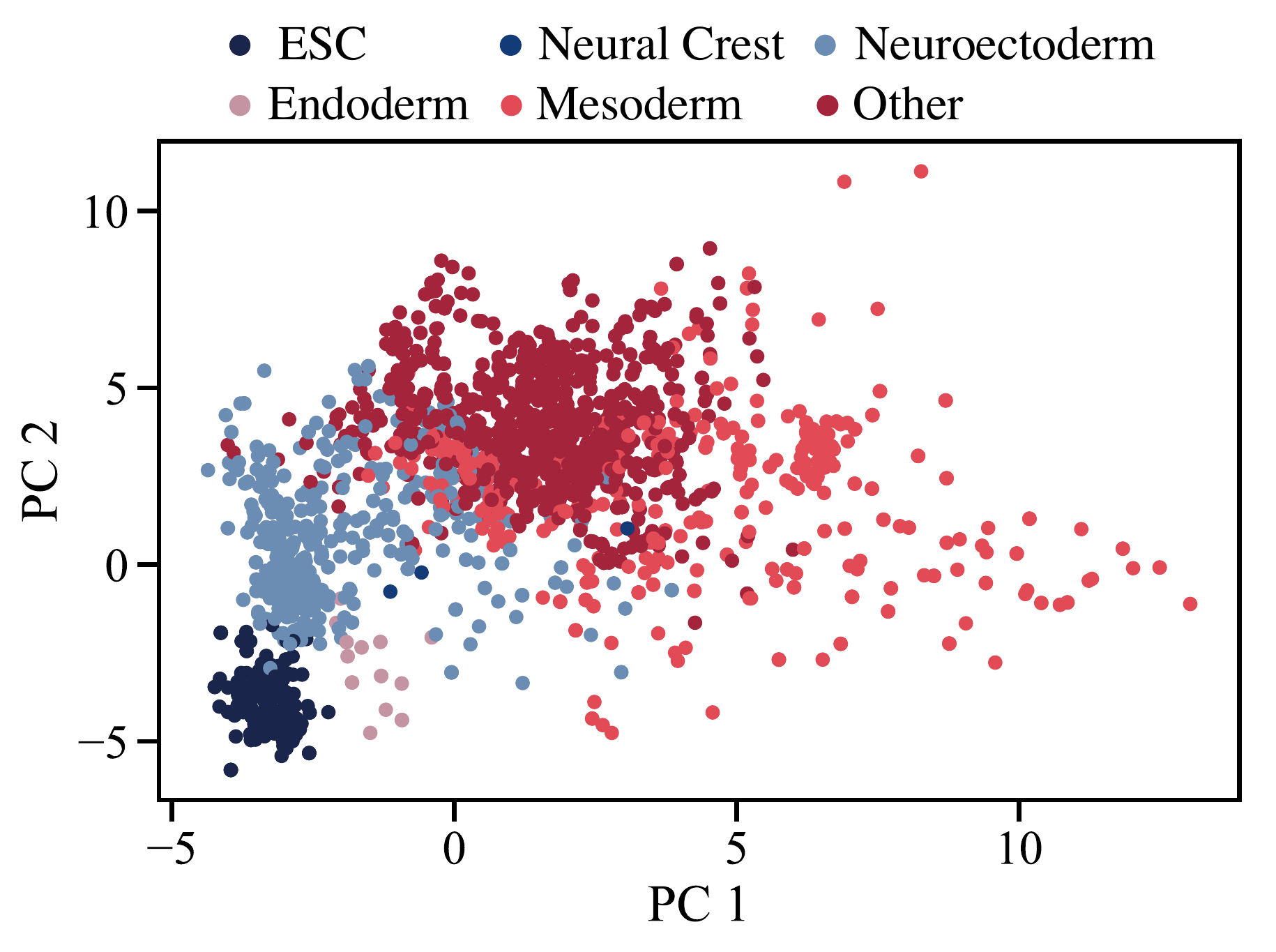}}
\caption{Predictions of the forward method on time-resolved embryoid body scRNA-seq data.}
\label{fig:exp_forward_cell}
\end{figure*}

\textsc{JKOnet} provides a model to understand complex population dynamics, by inferring the mechanism driving the population's time evolution.
This is achieved via solving a proximal gradient descent step in the Wasserstein space, which in our case is approximated using ICNNs.
\emph{Forward} methods, on the other hand, estimate the population at the next time step $t+1$ by directly moving along the gradient direction. Thus, $\eta_{t+1}$ is inferred via $(\nabla F_\xi)_{\#} \eta_t$, where $F_\xi$ is any arbitrary neural network \citep{hashimoto2016learning} and $\eta_t$ the predicted population at time point $t$.
In Figure~\ref{fig:exp_forward_pot_traj} we further evaluate the forward method on convex (\ref{fig:exp_forward_cell}a) and non-convex (\ref{fig:exp_forward_cell}b) potential-based dynamics, as well as trajectory-based dynamics (\ref{fig:exp_forward_cell}c and d).
Similarly as in the \textsc{JKOnet} setting, teacher forcing generally stabilizes and improves training of the energy functional $F_\xi$ (see Fig.~\ref{fig:exp_forward_pot_traj}c vs. ~\ref{fig:exp_forward_pot_traj}d).
Figure~\ref{fig:exp_forward_cell} further shows the performance of the forward method on predicting embryoid body developmental trajectories. For further discussion of the results, see \S~\ref{sec:eval_cell}.

\newpage
\subsection{Comparison to \emph{Forward} Methods} \label{app:overfitting}

\begin{table*}[h]
	\begin{minipage}{0.5\linewidth}
        \subfloat[\centering Forward method.]{\includegraphics[width=.49\linewidth]{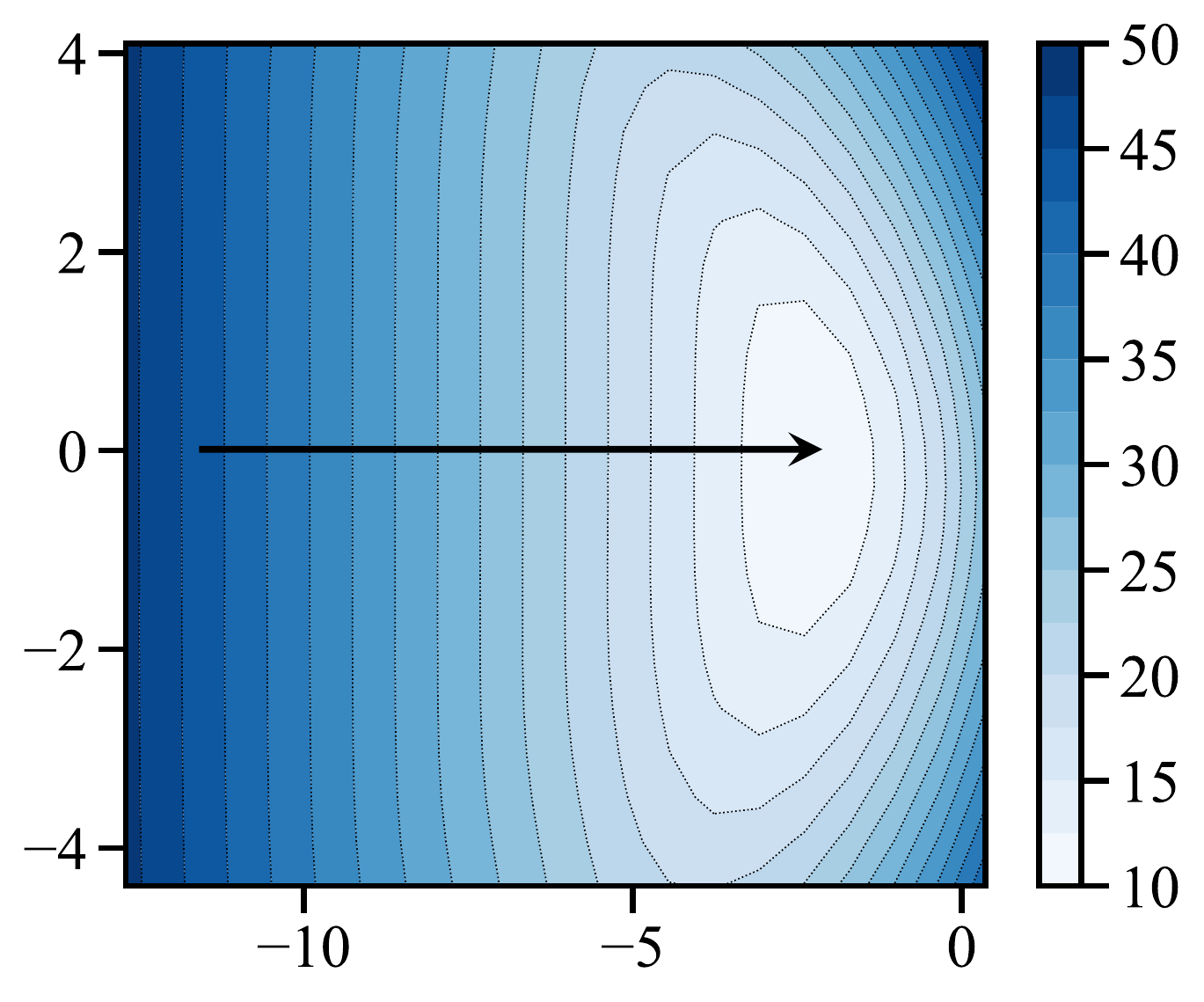}}
        \subfloat[\centering \textsc{JKOnet}.]{\includegraphics[width=.49\linewidth]{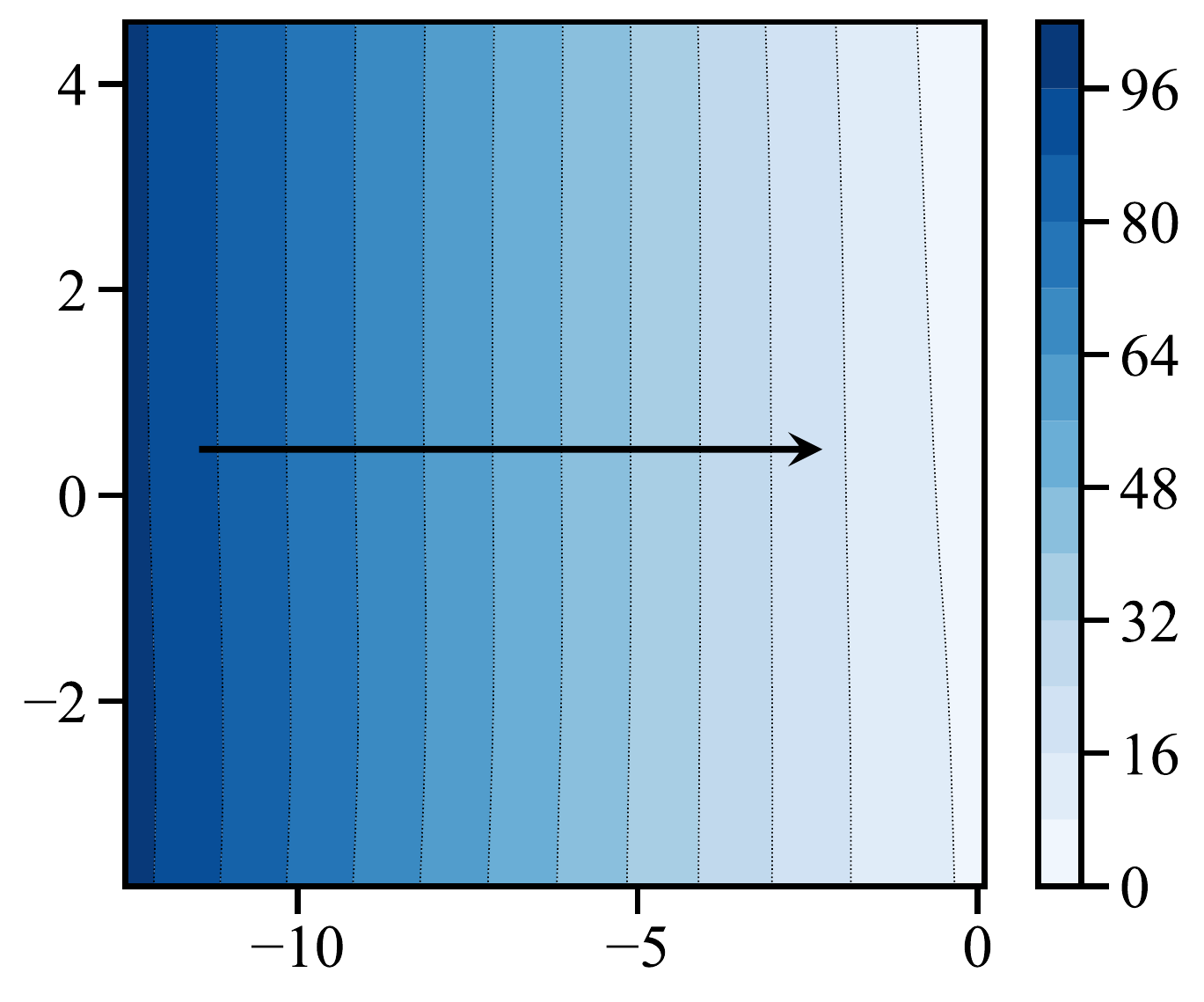}}
    	\captionof{figure}{Comparison between energy functionals $J_\xi$ of the line trajectory task between the forward method and \textsc{JKOnet}.}
        \label{fig:exp_comp_line}
	\end{minipage}\hfill
	\begin{minipage}{0.45\linewidth}
    	\centering
        \caption{Comparison of \textsc{JKOnet} to the forward method for predicting and extrapolating linear translations (see Figure~\ref{fig:exp_comp_line}) (using 3 runs).}
        \label{tab:comp_line}
		\resizebox{\textwidth}{!}{%
        \begin{tabular}{lcc}
            \toprule
            \textbf{Method} & \multicolumn{2}{c}{\textbf{Sinkhorn Distance}} ($\cW$) \\
            & Validation & Test \\
            \midrule
            Forward Method &  \textbf{1.94 $\pm$ 0.06} & 26.10 $\pm$ 1.76 \\
            \textsc{JKOnet} & 2.90 $\pm$ 0.37 & \textbf{20.30 $\pm$ 0.65} \\
            \bottomrule
        \end{tabular}}
	\end{minipage}
\end{table*}

\begin{wrapfigure}{r}{0.23\textwidth}
  \begin{center}
    \includegraphics[width=0.15\textwidth]{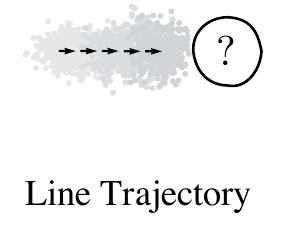}
  \end{center}
  \caption{Out-of-Sample Predictions along a Line.}
  \label{fig:task_overfitting}
\end{wrapfigure}

In the following, we extend the comparison of \textsc{JKOnet} to the forward method (see also \S~\ref{sec:eval_comp_fb}) and further demonstrates, that in the absence of any prior, we observe that the forward method can get more easily trapped in local minima, and overfit the training data.
Figure~\ref{fig:exp_comp_line} shows a simple experiment, in which we want to learn a population evolution along a line. During evaluation, we shift the line (see Fig.~\ref{fig:task_overfitting}) and evaluate the prediction performance w.r.t. the Sinkhorn distance \eqref{eq:sinkhorn}. Due to the less constrained energy, the \emph{forward} method perfectly resembles the seen trajectory during training, but fails to generalize and extrapolate on shifted test data (see Table~\ref{tab:comp_line}).

\section{Datasets} \label{app:datasets}
To evaluate \textsc{JKOnet}, we use multiple datasets comprising different examples of population dynamics. This includes synthetic population dynamics (potential- and trajectory-based dynamics), whose results are described in \S~\ref{sec:eval_synt}, as well as single-cell dynamics of a human developmental process, which we cover in \S~\ref{sec:eval_cell}.

\subsection{Potential-Based Dynamics} \label{app:potential_dataset}
In the following, we assume a random diffusion process evolving according to an {\^I}to stochastic difference equation (SDE) across time
\begin{equation*}
    d X_t=-\nabla \Phi(X_t) d t+\sqrt{2 \sigma^{2}} d B_t,
\end{equation*}
where $B(t)$ is the unit Brownian motion (standard Wiener process with magnitude $\sigma > 0$) and the drift is defined via a potential function $\Phi(x) : \mathbb{R}^d \rightarrow \mathbb{R}$.
The population-level inference problem on $X_t$ at each $t$ then satisfies the Fokker-Planck equation with fixed diffusion coefficient
\begin{equation*}
    \frac{\partial \rho_{t}}{\partial t}=\operatorname{div}\left(\nabla \Phi(x) \rho_{t}\right)+\sigma^{-1} \Delta \rho_{t}
\end{equation*}
with given initial condition $\rho_{0}=\rho^{0}$.
We generate the potential-based data by approximating trajectories $X_t$ via the Euler-Maruyama method \citep[\S~9.2]{kloeden1992stochastic}. 
Then given a \texttt{drift} (i.e., $\nabla \Phi$), one step of the Euler-Maruyama method is defined as
\begin{lstlisting}[language=Python]
    X = X + drift(X) * dt + np.random.normal(scale=sd, size=X.shape) * np.sqrt(dt).
\end{lstlisting}
In our experiments, we consider examples of convex, i.e., the quadratic potential $\Psi(x) = \|x\|^2_2$, and nonconvex potentials, i.e., Styblinski flow $\Psi(x)=\| 3 x^{3} -32 x+5 \|_{2}^{2}$.
For the convex potential, we simulate the trajectories using the Euler-Maruyama method with $\texttt{dt} = 0.25$ and $\texttt{sd}=0.2$ for $\texttt{n} = \texttt{t}/\texttt{dt}$ iterations, where $\texttt{t} = 1.0$.
Trajectories of the nonconvex potential are generated with $\texttt{dt} = 0.06$ and $\texttt{sd}=0.4$ for $\texttt{n} = \texttt{t}/\texttt{dt}$ iterations, where $\texttt{t} = 0.5$.

\subsection{Trajectory-Based Dynamics} \label{app:trajectory_dataset}
Besides population dynamics evolving according to a potential $\Psi$, we consider population dynamics following trajectories in space.
To achieve this, we generate data by moving a 2-dimensional Gaussian distribution along a pre-defined trajectory.
We compute 2-dimensional trajectories along the coordinates $x$ and $y$ via
\begin{lstlisting}[language=Python]
    x = r * np.cos(theta)
    y = r * np.sin(theta)
\end{lstlisting}
with radius \texttt{r} and angles \texttt{theta}.
The semicircle trajectory is computed using \texttt{r = 10} and \texttt{theta = np.linspace(2 * np.pi, 0, 100)}.
For the spiral trajectory, \texttt{r = np.linspace(10, 1, 100)} and \texttt{theta = np.linspace(2.75 * np.pi, 0, 100)} is used.
The line trajectory is generated using \texttt{x = np.linspace(-10, -2.5, 100)} and \texttt{y = np.zeros(100)}, where at test time, \texttt{x} is shifted to \texttt{x = np.linspace(-5, 7.5, 100)}.
Trajectory-based dynamics are then simulated by moving a 2-dimensional Gaussian distribution along these trajectories.
For the semicircle trajectories, this results in $T=5$ snapshots, the spiral-based population dynamics contain $T=10$ snapshots, and the line $T=2$ snapshots.

\subsection{Single-Cell Dynamics} \label{app:cell_dataset}

Developmental processes in biology involve tissue and organ development, body axis formation, cell division, and cell differentiation, e.g., the development of stem cells into functional cell types.
An example of such a process is the differentiation of embryonic stem cells (ESCs) into  hematopoietic, cardiac, neural, pancreatic, hepatocytic and germ lineages.
This development can be approximated \textit{in vitro} using embryoid bodies (EBs) \citep{martin1975}, three-dimensional aggregates of pluripotent stem cells, including ESCs \citep{shamblott2009derivation}.
Recently, \citet{moon2019} conducted a scRNA-seq analysis to unveil the developmental trajectories, as well as cellular and molecular identities through which early
lineage precursors emerge from human ESCs.
The dataset is available via \href{https://data.mendeley.com/datasets/v6n743h5ng}{Mendeley Data (V6N743H5NG)}\footnote{Dataset available via \url{https://data.mendeley.com/datasets/v6n743h5ng}.}.
In the following, we describe the preprocessing of the raw scRNA-seq data as well as the lineage branch analysis extracting the functional cell types emerging in this developmental process.

\subsubsection{Data Preprocessing}
To preprocess the data, we follow the analysis of \citet{moon2019} as well as \citet{luecken2019current}. For the analysis, we use the Python package \texttt{scanpy} \citep{wolf2018scanpy}.

\begin{figure}
    \centering
    \includegraphics[width=.3\linewidth]{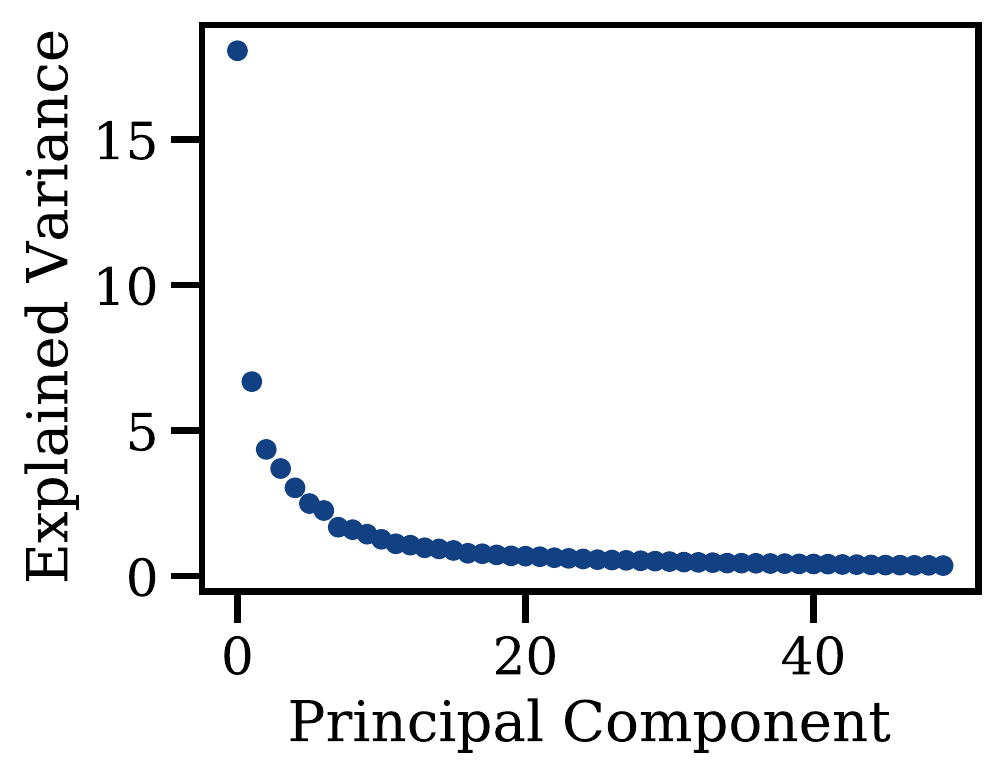}
    \caption{Proportion of explained variance per principal component of the embryoid body scRNA-seq data after preprocessing.}
    \label{fig:moon_expl_variance}
\end{figure}

\citet{moon2019} originally measure  approximately 31,000 cells over a 27 days differentiation time course, comprising gene expression matrices and barcodes, i.e., DNA tags used to identify reads originating from the same cell. The measured cells are then filtered in a quality control stage, their gene expression levels normalized and further processed in a feature selection step, where only highly-differentiated genes are selected.
The resulting data is then visualized using standard PCA as well as the dimensionality reduction method PHATE \citep{moon2019}, in order to extract biological labels.

The data quality control is based on the number of counts per barcode (count depth), the number of genes per barcode, and the fraction of counts from mitochondrial genes per barcode. We only keep cells with at least 4000 and at most 10000 counts, as well as more than 550 expressed genes and less than $20\%$ of mitochondrial counts, as a high fraction is indicative of cells whose cytoplasmic mRNA has leaked out through a broken membrane \citep{luecken2019current}.
For the subsequent analysis, we further only keep genes which are expressed in at least 10 genes.
After quality control, the dataset consists of 15150 cells and 17945 genes.
We normalize each cell by total counts over all genes and logarithmize the data matrix. We extract 4000 highly variable genes (HVG) using the 10X genomics preprocessing software \texttt{Cell Ranger} \citep{zheng2017massively} to further reduce the dimensionality of the dataset and include only the most informative genes.
Given the resulting data matrix with 15150 cells and 4000 genes across 5 different time points, we compute a corresponding low-dimensional embedding using PCA. Figure~\ref{fig:moon_expl_variance} shows the proportion of explained variance of each principal component (PC). We use the first 20 PCs for predicting population dynamics using \textsc{JKOnet} and the forward method.
This is in alignment with previous analysis of developmental trajectories, which use 5 \citep{tong2020trajectorynet} and 30 PCs \citep{schiebinger2019}, respectively.

\subsubsection{Lineage Branch Analysis of the Embryoid Body scRNA-Seq Data} \label{app:lineage_analysis}

\begin{figure*}[t]
\subfloat[\centering PHATE embedding hued by time of snapshot.]{\includegraphics[width=.8\linewidth]{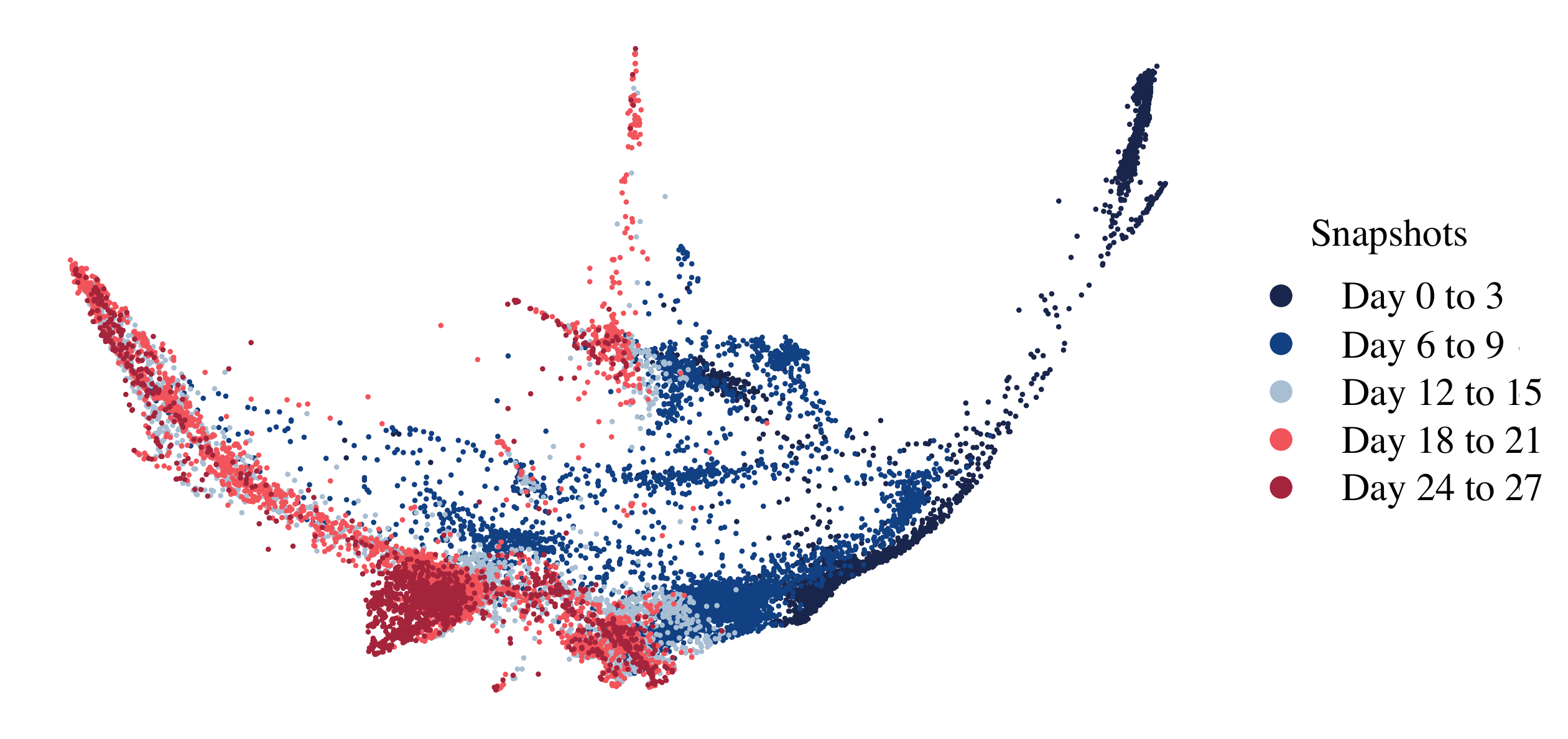}}

\subfloat[\centering PHATE embedding hued by k-Means clustering ($k=30$).]{\includegraphics[width=.8\linewidth]{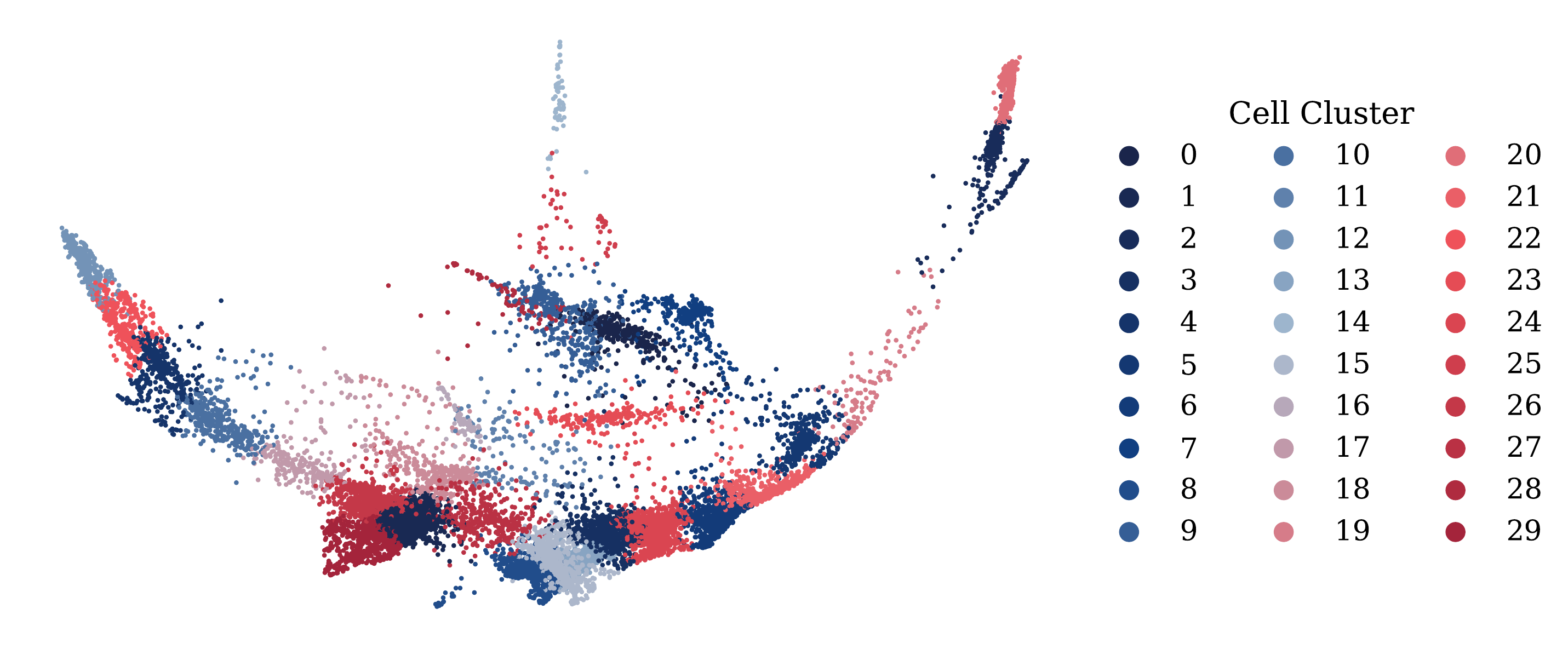}}

\subfloat[\centering PHATE embedding hued by predicted lineage branch.]{\includegraphics[width=.8\linewidth]{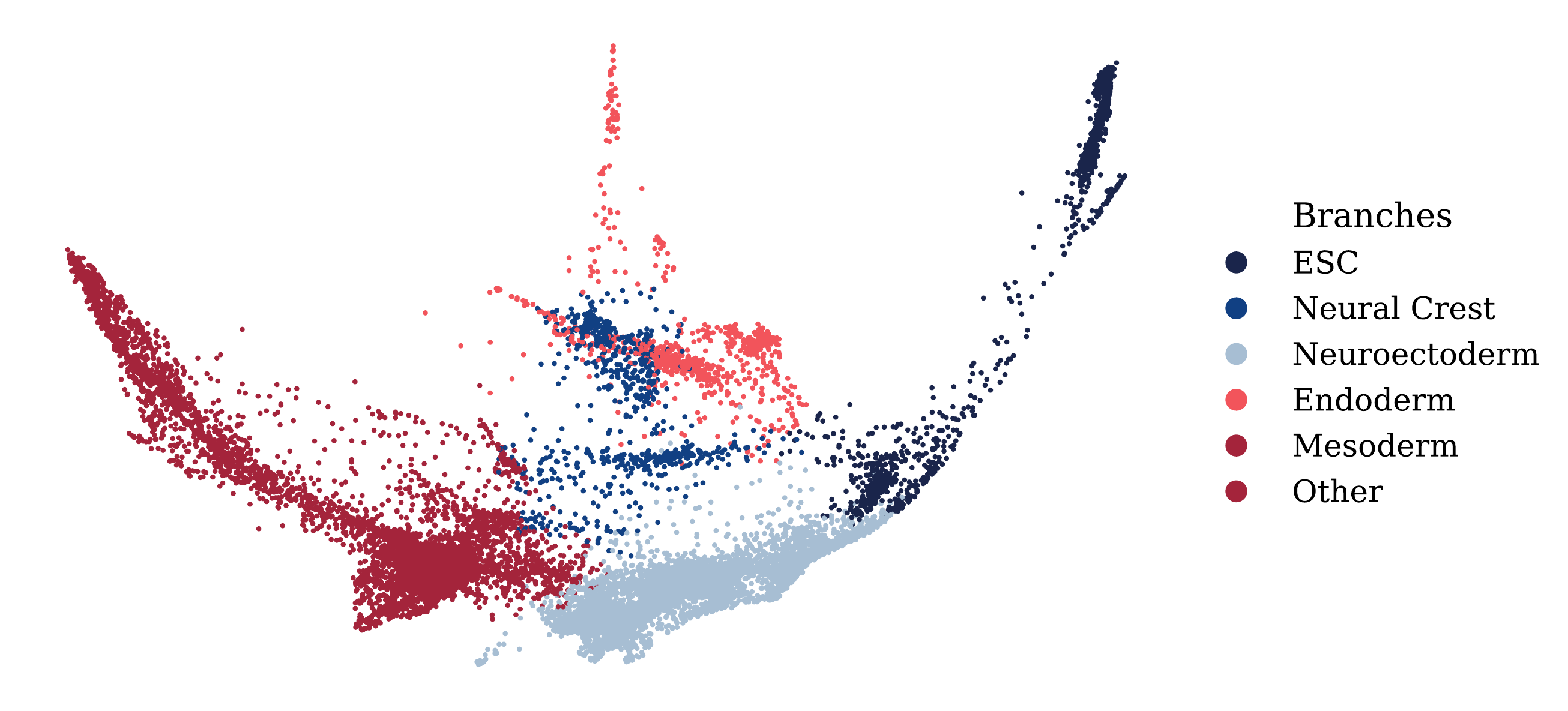}}

\caption{Analysis of embryoid body scRNA-seq data based on PHATE embedding \citep{moon2019}. Lineage branches are determined based on contiguous k-means clusters.}
\label{fig:moon_analysis}
\end{figure*}

To annotate the developmental process and detect lineage branches originating from the differentiation of embryonic stem cells, we follow the analysis of \citet{moon2019}.
Using a 10-dimensional PHATE embedding of the embryoid body scRNA-seq data (see the first two PHATE components in Fig.~\ref{fig:moon_analysis}a), we segment the dataset into 30 clusters using k-means.
PHATE is a non-linear dimensionality reduction method capturing a denoised representation of both local and global structure of a dataset \citep{moon2019}.
We then assign the resulting cluster to a lineage subbranch (\textit{i} - \textit{x}), using the following assignment of subbranch to cluster identification (see Fig.~\ref{fig:moon_analysis}b):
\begin{enumerate}[noitemsep, topsep=0pt, label=\roman*., , labelsep=*]
\begin{multicols}{4}
    \item 2, 20
    \item 5, 19
    \item 9, 11, 23
    \item 3, 6, 8, 13, 15, 21, 24
    \item 0, 7, 14, 25, 28
    \item 16, 18, 27
    \item 4, 10, 12, 17, 22
    \item 1
    \item 26
    \item 29.
\end{multicols}
\end{enumerate}
Then, subbranches are summarized to lineage branches using the assignment in \citet[Suppl. Note 4]{moon2019}:
\begin{itemize}[noitemsep, topsep=0pt, labelsep=*]
\begin{multicols}{3}
\centering
    \item[ESC.] i, ii
    \item[Neural Crest.] iii
    \item[Neuroectoderm.] iv
    \item[Endoderm.] v
    \item[Mesoderm.] vi, vii
    \item[Other.] viii, ix, x.
\end{multicols}
\end{itemize}
The resulting lineage branch annotation of the embryoid body scRNA-seq data can be found in Figure~\ref{fig:moon_analysis}c.

\section{Experimental Details} \label{app:experiments}
In the following, we describe the baselines considered, as well as provide details on network architectures and hyperparameters used.

\subsection{Baselines} \label{app:baselines}
We compare \textsc{JKOnet} with \emph{explicit} integration schemes (forward methods) such as \citet{hashimoto2016learning}.
In our proximal method, the prediction of the population $\rho_t$ at the next time step $t+1$ is parameterized via a separate function ($\psi_\theta$ \eqref{eq:next_pop}) and is thus decoupled from the free energy functional $J_\xi$ driving the underlying dynamics.
When learning \emph{forward} methods, however, the prediction is based on the gradient of an energy functional $F_\xi$. Given a distribution $\rho_t$ at time $t$ and energy $F_\xi$, the population particles at time $t+1$ are thus predicted via
\begin{equation*} 
\rho_{t+1} := (\nabla F_\xi)_{\#} \rho_t.
\end{equation*}
We parameterize $F_\xi(x)$ with a MLP similar as in \textsc{JKOnet} (see \ref{app:energy} for more details).
In this work we only consider linear functions \textit{in the space of measures}, i.e., expectations over $\rho$ of a vector-input neural network $E_\xi$ \eqref{eq:energy}.
In these cases, we can compare \textsc{JKOnet} to the forward methods described above.
Considering energies which take particle interactions into account, however, is not straightforward when using \emph{forward} methods.

\subsection{Network Architectures} \label{app:architecure}
In the following, we describe network architectures used in \textsc{JKOnet} to parameterize the \citeauthor{Brenier1987} map  $\psi_\theta$ (Section~\ref{app:icnn}) as well as the free energy functional $J_\xi$ (Section~\ref{app:energy}).

\subsubsection{Parameterization of Brenier Map} \label{app:icnn}

In the following, we describe the architectural details of the ICNN, parametrizing the \citeauthor{Brenier1987} map  $\psi_\theta$.
We set the hidden layer size of $W^x_l$ and $W^z_l$ \eqref{eq:icnn} to $64$ and use $3$ hidden layers before the final output layer ($L=4$ layer).
Similar to \citep{pmlr-v119-makkuva20a}, we use a squared leaky ReLU function with a small positive constant $\beta$ as \emph{convex} activation function for the first layer, i.e., $a_0(x) = \max(\beta x, x)^2$, and leaky ReLU $a_l(x) = \max(\beta x, x), \, l = 1, \dots, L-1$ as \emph{monotonically non-decreasing} and \emph{convex} activation functions the remaining layers. 
Crucial for the stability of training ICNNs is the choice of weight initialization.
We initialize  $W^x_l$ and $W^z_l$ \eqref{eq:icnn} from the standard normal distribution with standard deviation of $0.1$, significantly improving in performance over the initialization strategies for standard MLPs \citep{he2016deep, lecun2012efficient}.

We further tested the performance of the \emph{vanilla} ICNN to advanced formulations such as input-augmented ICNNs \citep{huang2021convex}, whereby no difference in performance is evident.
In addition, we evaluated the performance of \textsc{JKOnet} when relaxing the convexity constraints of $\psi_\theta$ by adding a penalty
\begin{equation*}
    R\left(\theta\right)=\lambda \sum_{W^z_{l} \in \theta}\left\|\max \left(-W^z_{l}, 0\right)\right\|_{F}^{2},
\end{equation*}
instead of enforcing its weights $W^z_{l}$ to only take values $>0$ as suggested in \citet{pmlr-v119-makkuva20a}.
This, however, did not increase performance of our method.

\subsubsection{Parameterization of Energy Functional} \label{app:energy}

The free energy functional $J_\xi$ can take various forms, accounting for diffusion as well as  potentials of interaction. In this work, we concentrate on linear functions in the space of measures \eqref{eq:energy}. 
We parametrize $E_\xi$ as a MLP with $2$ hidden layers of size $64$ with softplus activation functions, followed by a one-dimensional output layer. 
Future work will involve an extension of the framework to energy functionals covering higher-level interactions and population growth and decline, i.e., via deep sets \citep{zaheer2017} or set transformers \citep{lee2019}.

% \subsubsection{Set Transformer}

% Set Transformers .... \citep{lee2019}. 
% Given two sets represented as $n$ $d$-dimensional vectors $X, Y \in \mathbb{R}^{n \time d}$, \citet{lee2019} adapt the encoder block of the Transformer \citep{vaswani2017} to set by defining the \emph{Multi-Head Attention Block} (MAB) with parameters $\eta$ as
% \begin{align*}
%     \text{MAB} (X, Y) = \text{LayerNorm}(H + \text{rFF}(H))
% \end{align*}
% where $H = \text{LayerNorm}(X + \text{MultiHead}(X, Y, Y; \eta))$, $\text{rFF}$ is any row-wise feedforward layer and $\text{LayerNorm}$ is layer normalization \citep{ba2016}.
% The \emph{Set Attention Block} (SAB) is then defined as
% \begin{align*}
%     \text{SAB} \defeq \text{MAB}(X, X),
% \end{align*}
% thus performing self-attention between elements in the set. Stacking multiple SABs allows us to capture potential higher-order interactions between individuals of a population.
% % To bypass the quadratic time complexity $\mathcal{O}(n^2)$ of the SAB module, \citet{lee2019} further introduce the \emph{Induced Set Attention Block} (ISAB).
% The module $\text{SAB}(\cdot)$ is permutation equivariant.

\subsection{Hyperparameters and Training} \label{app:hyperparam}

For all experiments, we use a batch size of $250$.
For training the ICNN $\psi_\theta$, we use the Adam optimizer \citep{kingma2014adam} with learning rate $\text{lr}_\theta = 0.01$ ($\beta_1 = 0.5$, $\beta_2 = 0.9$). 
The fixed-point loop runs for minimally $50$ and maximally $100$ iterations with $\alpha = 1$.
When using a static number of iterations, we set the number of iterations to $100$.
We again use the Adam optimizer for learning the energy functional $J_\xi$ with learning rate ranging from $\text{lr}_\xi = 0.001$ to $0.0001$ ($\beta_1 = 0.5$, $\beta_2 = 0.9$).
In our experiments, we use a constant JKO step size $\tau = 1.0$. For all experiments, we use $\varepsilon = 1.0$ for the Sinkhorn loss \eqref{eq:fittingloss}. Trajectory-based dynamics are trained with an additional strong convexity regularizer using $\ell=0.8$.
Both, \textsc{JKOnet} and the forward method, are trained with gradient clipping with maximum global norm for an update of 10 \citep{pascanu2013difficulty}. 

\section{Reproducability}
An implementation of \textsc{JKOnet} can be found on \href{https://github.com/bunnech/jkonet}{github.com/bunnech/jkonet}.

\end{document}